\documentclass{article}
\usepackage{geometry}
\usepackage[american]{babel}
\usepackage{euler}
\usepackage{hyperref}
\usepackage{natbib} 
\usepackage{mathtools} 
\usepackage{booktabs} 
\usepackage{tikz} 
\usepackage{arxiv}

\title{Efficient Algorithms for Logistic Contextual Slate Bandits with Bandit Feedback}

\author{%
  \begin{tabular}[t]{@{}c@{\hspace{1cm}}c@{}}
    \begin{tabular}[t]{@{}c@{}}
      \textbf{Tanmay Goyal}\\
      \textnormal{Microsoft Research India}\\
      \texttt{\href{mailto:t-tangoyal@microsoft.com}{t-tangoyal@microsoft.com}}
    \end{tabular}
    &
    \begin{tabular}[t]{@{}c@{}}
      \textbf{Gaurav Sinha}\\
      \textnormal{Microsoft Research India}\\
      \texttt{\href{mailto:gauravsinha@microsoft.com}{gauravsinha@microsoft.com}}
    \end{tabular}
  \end{tabular}
}

\date{}

\renewcommand{\headeright}{}
\renewcommand{\undertitle}{}

\usepackage{amssymb, amsmath}
\usepackage{xcolor}
\usepackage{amsthm}
\usepackage{bm}
\usepackage{bbm}
\usepackage{multirow}
\usepackage{graphicx}
\usepackage{subcaption}

\usepackage{algorithmic, algorithm}

\theoremstyle{definition}
\newtheorem{definition}{Definition}[section]

\theoremstyle{plain}
\newtheorem{lemma}{Lemma}[section]
\newtheorem{theorem}{Theorem}[section]

\newtheorem{claim}{Claim}[section]

\newtheorem{proposition}{Proposition}[section]
\newtheorem{assumption}{Assumption}[section]

\DeclareMathOperator*{\argmax}{arg\,max}
\DeclareMathOperator*{\argmin}{arg\,min}

\newcommand{\slateglincb}{\texttt{Slate-GLM-OFU}}
\newcommand{\slateglincbts}{\texttt{Slate-GLM-TS}}
\newcommand{\slateglincbtsfixed}{\texttt{Slate-GLM-TS-Fixed}}
\newcommand{\adaofuecolog}{\texttt{ada-OFU-ECOLog}}
\newcommand{\tsecolog}{\texttt{TS-ECOLog}}

\def\DEBUG{true} 
\ifdefined\DEBUG
\newcommand{\gaurav}[1]{\textcolor{red}{[Gaurav:#1]}}
\else
 	    \newcommand{\gaurav}[1]{}

\fi

\newcommand{\R}{\mathbb{R}}

\renewcommand{\P}{\mathbb{P}}

\newcommand{\X}{\mathcal{X}}

\newcommand{\E}{\mathbb{E}}
\newcommand{\C}{\mathcal{C}}
\renewcommand{\H}{\mathcal{H}}

\newcommand{\filteration}[1]{\mathcal{F}_{#1}}

\newcommand{\inv}{^{-1}}
\newcommand*\diff{\mathop{}\!\mathrm{d}}

\newcommand{\eigmax}[1]{\lambda_{\max}\left(#1\right)}
\newcommand{\eigmin}[1]{\lambda_{\min}\left(#1\right)}

\newcommand{\singmax}[1]{\sigma_{\max}\left(#1\right)}
\newcommand{\singmin}[1]{\sigma_{\min}\left(#1\right)}

\newcommand{\mleq}{\preccurlyeq}
\newcommand{\mgeq}{\succcurlyeq}

\newcommand{\twonorm}[1]{\left\Vert#1\right\Vert_2}
\newcommand{\norm}[1]{\left\Vert#1\right\Vert}
\newcommand{\matnorm}[2]{\left\Vert#1\right\Vert_{#2}}

\newcommand{\sbrak}[1]{\left[#1\right]}
\newcommand{\pbrak}[1]{\left(#1\right)}
\newcommand{\cbrak}[1]{\left\{#1\right\}}
\newcommand{\modulus}[1]{\left|#1\right|}

\newcommand{\thetastar}{\bm{\theta}_\star}

\newcommand{\inner}[2]{#1^\intercal#2}

\newcommand{\summation}[2]{\sum\limits_{#1}^{#2}}

\newcommand{\sigmoid}[1]{\mu\pbrak{#1}}
\newcommand{\dsigmoid}[1]{\dot{\mu}\pbrak{#1}}
\newcommand{\sens}[2]{\dot{\mu}\pbrak{\inner{#1}{#2}}}

\newcommand{\diam}[2]{\text{diam}_{#1}\pbrak{#2}}
  
\begin{document}
\maketitle

\begin{abstract}
We study the Logistic Contextual Slate Bandit problem, where, at each round, an agent selects a slate of $N$ items from an exponentially large set (of size $2^{\Omega(N)}$) of candidate slates provided by the environment. A single binary reward, determined by a logistic model, is observed for the chosen slate. Our objective is to develop algorithms that maximize cumulative reward over $T$ rounds while maintaining low per-round computational costs. We propose two algorithms, \texttt{Slate-GLM-OFU}\ and \texttt{Slate-GLM-TS}, that accomplish this goal. These algorithms achieve $N^{O(1)}$ per-round time complexity via ``local planning'' (independent slot selections), and low regret through ``global learning'' (joint parameter estimation). We provide theoretical and empirical evidence supporting these claims. Under a well-studied diversity assumption, we prove that \texttt{Slate-GLM-OFU}\ incurs only $\tilde{O}(\sqrt{T})$ regret. Extensive experiments across a wide range of synthetic settings demonstrate that our algorithms consistently outperform state-of-the-art baselines, achieving both the lowest regret and the fastest runtime. Furthermore, we apply our algorithm to select in-context examples in prompts of Language Models for solving binary classification tasks such as sentiment analysis. Our approach achieves competitive test accuracy, making it a viable alternative in practical scenarios.
\end{abstract}

\section{Introduction}
\label{section:introduction}
Online slate bandit problems provide a popular framework for modeling decision-making scenarios where multiple items must be selected in each round.  A slate consists of multiple slots, each with its own pool of candidate items, which may change over time. In each round, the learner selects one item per slot, thereby forming a slate.  A single reward drawn from a logistic model with unknown parameters is then received for the entire slate. The learner's objective is to adaptively optimize their slate selection policy to maximize the cumulative reward (or equivalently, minimize the cumulative regret) over time.  Online slate bandits naturally model various real-world applications.  A prominent example is landing page optimization \citep{Hill_2017}, where the goal is to optimize the selection of components for each part of a landing page to maximize conversions.  Another important application is the automatic optimization of advertising creatives \citep{Chen2021}, which requires advertisers to automatically compose ads from multiple elements, such as product images, text descriptions, and titles. Beyond these practical applications, slate bandits have been extensively studied in the academic literature, leading to the development of many interesting algorithms in diverse settings \citep{Kale2010, Dimakopoulou2019, Rhuggenaath2020}.

Although good progress has been made on a variety of online slate bandit settings, some significant challenges remain that limit the applicability of these algorithms. In applications such as those mentioned above, at each round, the learner has access to some contextual information (such as user query, user history, or demographics), which influences the set of available items per slot. To the best of our knowledge, the current literature focuses heavily on the non-contextual (fixed-arms\footnote{We use the terms arms and actions interchangeably.}) setting, i.e., they do not assume access to such contexts and therefore keep the set of items unchanged over time. Another limitation is that most of the prior work assumes that the reward of a slate is a (known or unknown) function of rewards of the items in the slate, which are themselves either adversarially chosen or are stochastic but disjoint from each other (i.e., each item's reward comes from a different distribution). This assumption neglects the inherent similarities between items.  A more realistic approach is to assume a unified parametric reward model shared across all slates.  This model allows the learner to leverage shared information, significantly simplifying the learning process.  Specifically, for binary rewards, models based on the logistic or probit function can effectively capture the reward structure.

A third, and equally important, limitation is the prevalent focus on the semi-bandit feedback setting.  This setting provides separate reward feedback for each item within a selected slate.  However, many practical applications (e.g., the ad creatives problem \citep{Chen2021}) offer only a single, slate-level reward (i.e., bandit feedback).  Although there are some methods for converting bandit feedback to semi-bandit feedback \citep{Dimakopoulou2019}, these are often heuristic and lack theoretical guarantees.  The item-level feedback in the semi-bandit setting facilitates per-slot exploration and exploitation, enabling the development of algorithms with $N^{O(1)}$ per-round complexity (e.g., \citep{Kale2010, Rhuggenaath2020}) by avoiding explicit iteration over the entire slate space.  It remains unclear if we can achieve a similar efficiency in the more challenging bandit feedback setting.  For example, directly applying state-of-the-art bandit algorithms \citep{Lattimore2020} to the slate bandit problem (by treating the set of slates as the arm-set) and selecting a slate by iterating through the $2^{\Omega(N)}$ sized set of all possible slates, results in exponential per-round time complexity.

Motivated by these challenges, our work introduces efficient and optimal algorithms for the logistic contextual slate bandit problem under bandit feedback, assuming time-varying item features and rewards generated from a global logistic model.  We now list our contributions.   

\subsection{Our Contributions}
\label{subsection:our-contributions}

\begin{enumerate}
    \item We propose two new algorithms \slateglincb\ and \slateglincbts\ that solve the logistic contextual slate bandit problem under bandit feedback. While \slateglincb\ is based on the OFU (Optimization in the Face of Uncertainty) paradigm, \slateglincbts\ follows the Thompson Sampling (TS) paradigm. Under a well-studied diversity assumption (Assumption \ref{assumption: diversity}), we prove that \slateglincb\ incurs a regret of $\Tilde{O}(dN\sqrt{T})$ with high probability. Here, the slate comprises $N$ slots, and each item in the slate is $d$-dimensional. Further, $T$ is the total number of rounds the algorithm is run for. Both algorithms explore and exploit at the slot level and thus have a per-round time complexity of $poly(N , \log T)$, making them feasible in practice.

    \item We also propose a fixed arm version \slateglincbtsfixed\ for the \slateglincbts\ algorithm in the non-contextual (fixed-arm) setting. Using an assumption similar to Assumption \ref{assumption: diversity}, we prove an $\tilde{O}(d^{3/2}N^{3/2} \sqrt{T})$ regret guarantee for \slateglincbtsfixed. Similar to \slateglincbts, \slateglincbtsfixed\ also explores and exploits at the slot level and has a per-round time complexity of $poly(N , \log T)$.
    
    \item We perform extensive experiments to validate the performance of our algorithms in both settings: contextual and non-contextual. Under a wide range of randomly selected instances, we see that \slateglincb\ incurs the least regret compared to all baselines. Also, \slateglincbts\ and  \slateglincbtsfixed\ are competitive with other state-of-the-art algorithms. We also evaluate the maximum and average per-round time complexity of our algorithms and compare them to the time complexities of our baselines. Our algorithms are exponentially (most of the time) faster than all baselines. 

    \item Finally, we use our algorithm \slateglincb\ to select in-context examples for tuning prompts of language models, applied to binary classification tasks. We perform experiments on two datasets \emph{SST2} and \emph{Yelp Review} and achieve a competitive test accuracy of $\sim 80\%$, making it a viable alternative in practical prompt-tuning scenarios.
\end{enumerate}

\subsection{Related Work}
\label{subsection:related-work}
Online slate bandits have received significant attention due to their wide applicability in applications such as recommendations and advertising \citep{Hill_2017, Chen2021, Dimakopoulou2019}; however, only a few theoretical studies provide regret guarantees \citep{Kale2010, Rhuggenaath2020}. While these papers make progress on the slate bandit problem, neither do they address the contextual setting, nor do they accommodate bandit feedback, which are the main motivations of our work. While analyzing the Thompson Sampling approach in \cite{Dimakopoulou2019} might be feasible, their algorithm assigns equal rewards to each slot to maintain slot-level policies for efficiency, and hence, proving optimal guarantees may still be difficult. However, we would like to acknowledge that in our experiments (Section \ref{section:experiments}), in the fixed-arm instances that we considered, we found their algorithm to be quite competitive with ours.

One way of achieving optimal regret guarantees for the slate bandit problem is to reduce it to the canonical logistic bandit problem by treating each candidate slate as a separate arm, and subsequently, use state-of-the-art algorithms, such as those in \citep{Faury2020, Abeille2021, Faury2022}. While these algorithms do achieve optimal ($\kappa$-free) regret, they would be rendered infeasible in practice. This is because the selection of an arm would require iterating through all the arms (which is a $2^{\Omega(N)}$ sized set), thereby incurring exponential time per round. Even though these algorithms are inefficient for the slate bandit problem, we utilize some of their key ideas with an efficient planning approach to design our algorithms. In Section \ref{section:experiments}, we demonstrate that our algorithms perform better than these state-of-the-art logistic bandit algorithms, both in regret and time complexity, when applied to a wide variety of slate bandit instances.

Recently, a large number of works \citep{Swaminathan2017, Kiyohara2024, Vlassis2021} have studied the slate bandit problem in the off-policy setting, wherein they utilize a dataset collected using some base policy to find optimal slate bandit policies. While these works have made significant progress both from the theoretical and practical sides, we focus on the online setting, and hence, they are not relevant to our work.

\section{Preliminaries}
\label{section:preliminaries}
In this section, we define the notations used in the paper. Following this, we
formulate the Slate Bandits problem and present the assumptions that enable us to prove the regret guarantees stated in Theorem \ref{theorem: Regret OFUL} and Theorem \ref{theorem:TS}.

\paragraph{Notations:} We denote the set $\cbrak{1,2\ldots,N}$ as $\sbrak{N}$. Unless otherwise specified, all vectors, matrices, and sets are represented using bold lower-case, bold upper-case, and calligraphic upper-case or Greek letters, respectively. 
For any matrix $\mathbf{A}$, we denote its minimum and maximum eigenvalues as $\lambda_{\min}(\mathbf{A})$ and $\lambda_{\max}(\mathbf{A})$ respectively. 
We write $\mathbf{A}\mgeq 0$, if all the eigenvalues of the matrix $\mathbf{A}$ are non-negative (i.e, it is positive semi-definite), and $\mathbf{A}\mgeq \mathbf{B}$, if $\mathbf{A}-\mathbf{B}\mgeq 0$. For a positive semi-definite matrix $\mathbf{A}$, we define the norm of a vector $\mathbf{x}$ with respect to $\mathbf{A}$ as $\matnorm{\mathbf{x}}{\mathbf{A}} = \sqrt{\mathbf{x}^\top{\mathbf{A}}\mathbf{x}}$ and the spectral norm of $\mathbf{A}$ as $\twonorm{\mathbf{A}} = \sqrt{\eigmax{\mathbf{A}^\top\mathbf{A}}}$. We use $\mathbf{I}_{m}$ and $\mathbf{0}_{m}$ to denote the $m\times m$ identity and zero matrices respectively. When the dimension $m$ is clear from the context, we use $\mathbf{I}$ and $\mathbf{0}$ instead. The symbols $\P$ and $\E$ denote probability and expectation, respectively. For a set $\mathcal{X} \subseteq \R^m$, we define the diameter of $\mathcal{X}$ as $diam(\mathcal{X}) = \max\limits_{\mathbf{x}_1 ,\mathbf{x}_2 \in \mathcal{X}} \lVert \mathbf{x}_1 - \mathbf{x}_2 \rVert_2$. Also, for two sets $\mathcal{A}$ and $\mathcal{X}$, such that $\mathcal{A},\mathcal{X} \subseteq \R^m$, the diameter of $\mathcal{X}$ with respect to $\mathcal{A}$ is defined as $diam_{\mathcal{A}}(\mathcal{X}) = \max\limits_{\mathbf{a} \in \mathcal{A}} \max\limits_{\mathbf{x}_1 , \mathbf{x}_2 \in \mathcal{X}} \lvert \mathbf{a}^\top (\mathbf{x}_1 - \mathbf{x}_2) \rvert$. Finally, throughout the paper, we use $\tilde{O}(.)$ to suppress logarithmic factors.

\subsection{Slate Bandits}
In the Slate Bandit problem, a learner interacts with the environment over $T$ rounds. At each round $t\in [T]$, the learner is presented with $N$ finite sets $\mathcal{X}^i_t$ $(\subset \R^{d}), i\in [N]$, of \emph{items} and is expected to select one item (say $\mathbf{x}^i_t$) from each $\mathcal{X}^i_t$. Based on the selected $N$-tuple $\mathbf{x}_t = (\mathbf{x}^1_t,\ldots,  \mathbf{x}^N_t)$ (called a ``slate''), the learner receives a stochastic binary reward $y_t(\mathbf{x}_t)$. The goal of the learner is to select a sequence of slates $\{\mathbf{x}_t\}_{t \in [T]}$ such that her expected regret
\[
Regret(T) = \sum\limits_{t=1}^T \bigg\{\max_{\mathbf{x}\in \mathcal{X}_t}\E[y_t(\mathbf{x})] - \E[y_t(\mathbf{x}_t)] \bigg\}
 \]
is minimized\footnote{We also use $R(T)$ for shorthand.}. Here, $\mathcal{X}_t$ denotes the set $\mathcal{X}^1_t\times\ldots\times \mathcal{X}^N_t$ of all possible slates at round $t$. When the chosen slate $\mathbf{x}_t$ is clear from the context, for simplicity, we will denote $y_t(\mathbf{x}_t)$ as $y_t$. For simplicity, we say that the slate $\mathbf{x}_t$ comprises $N$ ``slots'', and the item $\mathbf{x}_t^i$ is placed in slot $i$ in the slate.


In this work, we consider two well-known settings:
\begin{enumerate}
    \item \textbf{Stochastic Contextual Setting}: In this setting, at each round $t \in [T]$ and for each slot $i \in [N]$, we assume that the set $\mathcal{X}^i_t$ is constructed by sampling from a distribution $\mathcal{D}_i$, in an $\mathrm{i.i.d}$ fashion. Moreover, $\mathcal{X}_t^i$ and $\mathcal{X}_s^j$ are sampled independently of one another, for all $s,t \in [T]$ and $i,j \in [N]$. 
    \item \textbf{Non-Contextual Setting}: This setting is also known as the Fixed-Arm Setting.  Here, the item-sets $\mathcal{X}_t^i$ for all slots $i \in [N]$ remain fixed over time. Thus, for simplicity, we denote $\mathcal{X}_t^i$ by $\mathcal{X}^i$ for all $t \in [T]$.
\end{enumerate}

\paragraph{Logistic rewards:}In this paper, we assume that the binary reward variable $y_t$ comes from a Logistic Model. Therefore, $
\P[y_t=1 \mid \mathbf{x}_t] = \mu(\mathbf{x}_t^\top\mathbf\theta^\star)$,
where $\mu:\R\rightarrow\R$ is the logistic function, i.e., $\mu(a) = 1/(1+\exp(-a))$, and
$\mathbf{\theta}^\star\in \R^{dN}$ is an unknown $dN$-dimensional parameter vector. Similar to prior works on Logistic bandits \citep{Faury2020, Abeille2021, Faury2022}, we assume that $\twonorm{\mathbf\theta^\star}\leq S$, where $S$ is known to the learner\footnote{Often, an upper bound on $S$ suffices.}, and $\twonorm{\mathbf{x}^i}\leq 1/\sqrt{N}$, for all $\mathbf{x}^i\in \mathcal{X}_t^i$, $i\in [N], t\in [T]$\footnote{This implies the usual assumption $\twonorm{\mathbf{x}}\leq 1$ for all $\mathbf{x}\in \mathcal{X}_t$.}. 

Recent logistic bandit literature \citep{Filippi2010, Faury2020, Abeille2021, Faury2022} also identifies a critical parameter $\kappa$ that captures the non-linearity of the reward for the given problem instance, defined as follows:

\begin{equation}
    \kappa = \max_{t\in [T]} \; \max_{\substack{\mathbf{x} \in \X_t \\ \mathbf\theta\in\Theta}} \frac{1}{\dot{\mu}(\mathbf{x}^\top \mathbf\theta)},
    \label{eqn:kappa}
\end{equation}

where $\Theta =\{\twonorm{\mathbf\theta}\leq S\} \subset \R^{dN}$. The parameter $\kappa$ can be intuitively seen as the mismatch between the true reward function and a linear approximation of the same. It is well known that $\kappa$ grows exponentially with the diameter of $\Theta$ \cite{Faury2020}, and hence, developing algorithms with regret independent of $\kappa$ is ana ctive area of research that has gained significant attention \citep{Faury2020, Abeille2021, Faury2022, Sawarni2024}.
We refer the reader to Section 2 of \cite{Faury2020} for a thorough discussion on $\kappa$ and its implications on regret analysis. 

We now describe a key assumption that enables us to design algorithms with low per-round computational complexity and strong regret guarantees (Theorem \ref{theorem: Regret OFUL} in Section \ref{section:main-algo} and Theorem \ref{theorem:TS} in Appendix \ref{appendix:ts-algos}). 

\begin{assumption}
    (\textbf{Diversity Assumption}) Let $\mathcal{F}_t$ be the sigma algebra generated by $\{\mathbf{x}_1, y_1, \ldots, \mathbf{x}_{t-1}, y_{t-1}\}$ and $\emptyset = \mathcal{F}_0\subset\mathcal{F}_1\subset \ldots \mathcal{F}_T$, be the associated filtration. For each $i\in [N]$ and $t\in [T]$, we assume that,
\[
\E[\mathbf{x}_t^i\mid \mathcal{F}_t] = \mathbf{0} \hspace{1em}\text{and} \hspace{1em} \E[\mathbf{x}_t^i{\mathbf{x}_t^i}^\top \mid \mathcal{F}_{t}] \mgeq \rho \kappa \mathbf{I},
\]

where $\rho >0$ is a fixed constant and $\kappa$ is the non-linearity parameter defined earlier in \eqref{eqn:kappa}.
\label{assumption: diversity}
\end{assumption}

\textbf{Remarks on Assumption \ref{assumption: diversity}: }
The diversity assumption intuitively ensures that for each slot $i\in [N]$ and round $t\in [T]$, the item features $\mathbf{x}_t^i$ selected by the algorithm are sufficiently ``diverse'', i.e., the expected matrix $\E[\mathbf{x}_t^i{\mathbf{x}_t^i}^\top \mid \mathcal{F}_{t}]$ is full-rank and has sufficiently large eigenvalues. In our proofs, we first use this assumption to prove that, with high probability, the minimum eigenvalue of certain design matrices used in our algorithms (Algorithms \ref{algo:batch_OFUL}, \ref{algo:TS}, \ref{algo:TS-Fixed}) grows (sufficiently) linearly with time. We denote these design matrices as $\mathbf{W}^i_t$ and define them as
\[
\mathbf{W}_t^i = \mathbf{I}_d + \sum_{s\in [t-1]\setminus \mathcal{T}}\dot{\mu}(\mathbf{x}_s^\top\mathbf{\theta}_{s+1})\mathbf{x}_s^i{\mathbf{x}_s^i}^\top \; \forall i \in [N],
\]
where $\mathcal{T} \subset [T]$.
In particular, in Lemma \ref{lemma: min_eig_design}, we show that
\[ 
\P\cbrak{\lambda_{min}(\mathbf{W}_t^i) \geq 1 + c\rho |[t-1]\setminus \mathcal{T}|} \geq 1 - \delta,
\]
for a fixed constant $c \in (0,1)$. Then, in Lemma \ref{lemma: ineq on W} and Lemma \ref{lemma:multiplicative-equivalence}, we critically utilize this fact 
to prove a multiplicative equivalence between a block diagonal matrix comprising $\mathbf{W}^i_t$ for all slots $i \in [N]$, and a similarly defined slate-level design matrix
$\mathbf{W}_t = \mathbf{I}_{Nd} + \sum_{s\in [t-1]\setminus \mathcal{T}}\dot{\mu}(\mathbf{x}_s^\top\mathbf{\theta}_{s+1})\mathbf{x}_s\mathbf{x}_s^\top$, i.e, we show that for some $c^\prime \in (0,1)$
\[
    (1-c^\prime) \textrm{diag} (\mathbf{W}^1_t , \ldots , \mathbf{W}^N_t) \preceq \mathbf{W}_t \preceq (1+c^\prime) \textrm{diag} (\mathbf{W}^1_t , \ldots , \mathbf{W}^N_t).
\]
This multiplicative equivalence allows us to replace the slate-level exploration bonuses with the corresponding slot-level exploration bonuses, which results in a low per-round time complexity in Algorithms \ref{algo:batch_OFUL}, \ref{algo:TS} and \ref{algo:TS-Fixed}, while still exhibiting optimal regret guarantees. Details of the algorithm and the regret proof can be found in Sections \ref{section:main-algo}, \ref{section:TS} and Appendix \ref{appendix:ts-algos}. 

We would also like to highlight that many similar diversity assumptions have been studied throughout the literature, and several connections between these assumptions have been established (Section $3$, \cite{Papini2021}). Depending on the strength of the assumption, novel and stronger regret guarantees for well-known algorithms have been established (e.g., Lemma $2$, \cite{Papini2021} and Corollary $4$, \cite{Das_2024}). Interestingly, a core component of these regret proofs was also to show a linear lower bound on the minimum eigenvalue of the design matrix. Note that since the assumption is instance/algorithm-dependent, there may exist instances where this linear lower bound may not hold. To study this, we empirically examine the growth of the minimum eigenvalues ($\lambda_{\min}(\mathbf{W}_t^i)$) for a large number of randomly chosen instances. In all of the instances, we could clearly see a linear trend, thus validating the assumption, at least for these randomly picked instances. We refer the reader to Appendix \ref{appendix:empirical-validation} for more details on this empirical evaluation.

\clearpage
\section{\slateglincb}
\label{section:main-algo}

\begin{algorithm}[!ht] 
\caption{\texttt{Slate-GLM-OFU}}
\label{algo:batch_OFUL} 
\begin{algorithmic}[1]
\STATE \textbf{Inputs:} Horizon $T$, probability of failure $\delta$, and upper bound on parameter norm $S$.

\STATE Initialize $\{\mathbf{W}_1^i\}_{i \in [N]} = \mathbf{I}_d$, $\mathbf{W}_1 = I_{dN}$, $\Theta_1 = \{\twonorm{\bm\theta} \leq S\}$, $\bm\theta_1 \in \Theta_1$, $\eta_t(\delta) = O(S^2Nd\log(t/\delta))$, and $ \H_1 = \emptyset$.
        
\FOR{each round $t \in [T]$}
\STATE For all slots $i \in [N]$, obtain the set of items $\mathcal{X}^i_t$ and find
$\mathbf{x}^i_t = \argmax_{\mathbf{x} \in \X^i_t} \mathbf{x}^\top \bm\theta_t^{i}+ \sqrt{\eta_t(\delta)} \matnorm{\mathbf{x}}{(\mathbf{W}_t^i)^{-1}}$.

\STATE Construct the slate $\mathbf{x}_t = (\mathbf{x}^1_t, \ldots, \mathbf{x}^N_t)$ and obtain the corresponding reward $y_t$.

\STATE Obtain $\bm\theta_{t+1}$, $\{\mathbf{W}_{t+1}^i\}_{i=1}^N$, $\Theta_{t+1}$, $\mathcal{H}_{t+1}$ by calling Algorithm \ref{algo:adaptive-updates} with inputs $\mathbf{x}_t$, $y_t$, $\bm\theta_t$, $\mathbf{W}_t$, $\{\mathbf{W}_{t}^i\}_{i=1}^N$, $\Theta_t$, $\mathcal{H}_t$.
\ENDFOR
\end{algorithmic}
\end{algorithm}

\begin{algorithm}[ht] 
\caption{\texttt{ada-OFU-ECOLog-Updates}}
\label{algo:adaptive-updates} 
\begin{algorithmic}[1]
\STATE \textbf{Inputs:} Slate $\mathbf{x}_t$, reward $y_t$, design matrices $\mathbf{W}_t, \{\mathbf{W}_{t}^i\}_{i=1}^N$, admissible set of parameters $\Theta_t$, set of indices $\mathcal{H}_t$.
\vspace{1mm}
\STATE Initialize $\gamma_t(\delta) = O(S^2Nd\log(t/\delta))$ and $\beta_t(\delta) = O(S^6Nd\log(t/\delta))$.
\vspace{1mm}
\STATE Compute $\bar{\bm\theta}_t$, $\bm\theta^0_t$, and $\bm\theta^1_t$ using 
\eqref{theta_bar} and \eqref{theta_u}.

\IF{$\dot{\mu}(\mathbf{x}_t^\top\bar{\bm\theta}_t) \leq 2\dot{\mu}(\mathbf{x}_t^\top\bm\theta^u_t)$ for $u \in \{0,1\}$}

\STATE Let $\bm\theta_{t+1}$ be solution of  \eqref{equation:optimization} up to precision $1/t$.

\STATE Update $\mathbf{W}_{t+1}^i \gets \mathbf{W}_t^i + \dot{\mu}(\mathbf{x}_t^\top\bm\theta_{t+1})\mathbf{x}_t^i{\mathbf{x}_t^i}^\top$, $\forall i\in [N]$.

\STATE Update $\mathbf{W}_{t+1} \gets \mathbf{W}_t + \dot{\mu}(\mathbf{x}_t^\top\bm\theta_{t+1})\mathbf{x}_t{\mathbf{x}_t}^\top$.

 \STATE Update $\H_{t+1} \gets \H_{t}$ and $\Theta_{t+1} \gets \Theta_t$.
 

\ELSE
\STATE Update $\H_{t+1} \gets \H_t \cup \{(\mathbf{x}_t, y_{t})\}$. 

\STATE Let $\bm\theta^\H_{t+1}$ be solution of \eqref{equation:optimization2} up to precision $1/t$.
                
\STATE Update $\mathbf{V}_{t}^\H \gets \kappa^{-1}\sum_{(\mathbf{x},y) \in \H_t}\mathbf{x}\mathbf{x}^\top + \gamma_t(\delta)\mathbf{I}_{Nd}$.
                
\STATE Update $\Theta_{t+1} \gets \cbrak{\matnorm{\bm\theta-\bm\theta^\H_{t+1}}{\mathbf{V}^\H_{t}}^2 \leq \beta_t(\delta)} \cap \Theta_1$.
\STATE Update $\bm\theta_{t+1} \gets \bm\theta_t$, $\mathbf{W}_{t+1} \gets \mathbf{W}_t$, and $\mathbf{W}_{t+1}^i \gets \mathbf{W}_t^i$, $\forall i\in [N]$.
\ENDIF
\STATE \textbf{return} $\bm\theta_{t+1}, \mathbf{W}_{t+1}, \{\mathbf{W}_{t+1}^i\}_{i=1}^N, \Theta_{t+1}, \mathcal{H}_{t+1}$.
\end{algorithmic}
\end{algorithm}

In this section, we present our first algorithm \slateglincb\ (Algorithm \ref{algo:batch_OFUL}) based on the OFU (Optimization in the Face of Uncertainty) paradigm \citep{Yadkori2011} used in bandit algorithms. At a high level, \slateglincb\ (along with sub-routine Algorithm \ref{algo:adaptive-updates}) builds upon the \adaofuecolog\ algorithm (Algorithm $2$ in \cite{Faury2022}) which achieves an optimal ($\kappa$-free) $O(\sqrt{T})$ regret guarantee for logistic reward models and incurs $O(K\log T)$ per round computational cost, where $K$ is the total number of actions to choose from. In the slate bandit setting, $K$ is exponential in $N$, the number of slots in the slate, making a direct application of \adaofuecolog\ infeasible when $N$ is large. To address this, \slateglincb\ selects an item for each slot independently, reducing the per-round computational cost to $N^{O(1)}$. Interestingly, despite choosing all the items comprising the slate independently, \slateglincb\ (via sub-routine Algorithm \ref{algo:adaptive-updates}) estimates only a single reward model using the slate-level reward feedback. This is a critical difference with respect to prior works on slate bandits with bandit feedback \citep{Dimakopoulou2019}, where the algorithm attributes the single slate-level reward feedback to individual items in the slate and estimates $N$ different reward models. 

\slateglincb\ takes as input $T$, the total number of rounds, $\delta$, the error probability, and $S$, a known upper bound for $\twonorm{\bm\theta^\star}$.
Similar to \adaofuecolog\ \citep{Faury2022}, \slateglincb\ maintains vectors $\bm\theta_t$, and sets $\Theta_t$ and $\mathcal{H}_t$. The vector $\bm\theta_t$ provides an estimate of $\bm\theta^\star$ during the $t^{th}$ round.
The set $\Theta_t \subseteq \Theta_1 =  \{\twonorm{\bm\theta}\leq S\}$ is the admissible set for the values of $\bm\theta_{t+1}$, and with high probability, contains the true reward parameter $\bm\theta^\star$ (See Proposition 7 in \cite{Faury2022}). In order to facilitate adaptivity, \adaofuecolog\  
introduced the sets $\mathcal{T}$ and $\mathcal{H}_t$, where $\mathcal{T}$ comprises all time rounds at which an inequality criterion (described in \emph{Step 4} of Algorithm \ref{algo:adaptive-updates}) fails, and $\mathcal{H}_t$ comprises the arm-reward pairs corresponding to these time rounds up to round $t$, i.e, $\mathcal{H}_t = \{(\mathbf{x}_s , y_s)\}_{s \in [t-1] \cap \mathcal{T}}$.

In addition to these, to enable efficient per-round computation of parameter estimates, \adaofuecolog\ also introduced the matrix $\mathbf{W}_t$ as an on-policy proxy for the concentration matrix $\mathbf{H}_t$, where $\mathbf{H}_t$ and $\mathbf{W}_t$ are defined as
\[
    \mathbf{H}_t =  \mathbf{I}_{Nd} + \sum_{s \in [t-1] \setminus \mathcal{T}}\dot{\mu}(\mathbf{x}_s^\top\bm\theta^\star)\mathbf{x}_s\mathbf{x}_s^\top, \quad \mathbf{W}_t= \mathbf{I}_{Nd} + \sum_{s \in [t-1] \setminus \mathcal{T}} \dot{\mu}(\mathbf{x}_{s}^\top \bm\theta_{s+1})\mathbf{x}_{s}\mathbf{x}_{s}^\top.
\]

 In \slateglincb, we additionally maintain $N$ other such matrices corresponding to each of the $N$ slots. The $i^{th}$ matrix, denoted by $\mathbf{W}^i_t$, helps in the explore-exploit trade-off while selecting an item for the $i^{th}$ slot. This matrix $\mathbf{W}^i_t$ is defined as
 \[
    \mathbf{W}_t^i =  \mathbf{I}_{d} + \sum_{s \in [t-1]\setminus \mathcal{T}} \dot{\mu}(\mathbf{x}_{s}^\top \bm\theta_{s+1})\mathbf{x}_{s}^i{\mathbf{x}_{s}^i}^\top.
 \]
 
Next, we go through the steps of \slateglincb\ (Algorithm \ref{algo:batch_OFUL}) and its sub-routine (Algorithm \ref{algo:adaptive-updates}) to provide a more detailed explanation.
\emph{Steps 3-7} (Algorithm \ref{algo:batch_OFUL}) is where \slateglincb\ significantly differs from \adaofuecolog. Instead of obtaining the set of arm features $\mathcal{X}_t$ (in our case, slates) directly from the environment (as in \adaofuecolog), \slateglincb\ receives $N$ different sets of items $\mathcal{X}_t^i,$ for each slot $i\in [N]$. Then, it picks an item $\mathbf{x}_t^i \in \mathcal{X}_t^i$, using the optimistic rule mentioned in \emph{Step 4} (Algorithm \ref{algo:batch_OFUL}). Note that the underlying optimization problem for slot $i$ only requires the candidate items in $\X_t^i$ and the components $\bm\theta_t^i$ of $\bm\theta_t$ corresponding to the $i^{th}$ slot, and hence, can be solved independently and in parallel for all slots. A core technical part of our regret guarantee (Theorem \ref{theorem: Regret OFUL}) is to demonstrate how the independent selection of items at the slot-level leads to optimal selection at the slate-level. Essentially, we can show that, under our diversity assumption (Assumption \ref{assumption: diversity}), with high probability, the positive definite matrices $\mathbf{W}_t$ and $\textrm{diag}(\mathbf{W}_t^1,\ldots, \mathbf{W}_t^N)$ are multiplicatively equivalent, i.e, for some $c^\prime \in (0,1)$,
\[
    (1-c^\prime) \textrm{diag}(\mathbf{W}_t^1,\ldots, \mathbf{W}_t^N) \preceq \mathbf{W}_t \preceq (1+c^\prime) \textrm{diag}(\mathbf{W}_t^1,\ldots, \mathbf{W}_t^N).
\]
This multiplicative equivalence allows us to show that for all slates $\mathbf{x}_t = (\mathbf{x}^1_t , \ldots , \mathbf{x}^N_t)$, 
\[
    \matnorm{\mathbf{x}_t}{\mathbf{W}_t\inv} \leq  \frac{1}{1-c^\prime} \sum_{i\in [N]}\matnorm{\mathbf{x}_t^i}{(\mathbf{W}_t^i)\inv},
\]
i.e, the quantities $\matnorm{\mathbf{x}_t}{\mathbf{W}_t\inv}$ and $\sum_{i\in [N]}\matnorm{\mathbf{x}_t^i}{(\mathbf{W}_t^i)\inv}$ are also multiplicatively equivalent. We then exploit this observation to convert the slate-level optimistic selection rule into an equivalent optimistic selection rule for each slot.
In \emph{Step 5} (Algorithm \ref{algo:batch_OFUL}), we select the slate $\mathbf{x}_t = (\mathbf{x}_t^1, \ldots, \mathbf{x}_t^N)$, yielding a reward $y_t$. At this point, \slateglincb\ calls a sub-routine described in Algorithm \ref{algo:adaptive-updates} which updates the parameters $\bm\theta_t$, $\mathbf{W}_t$, $(\mathbf{W}_t^1, \ldots, \mathbf{W}_{t}^N)$, $\Theta_t$, and $\mathcal{H}_t$. The update rules in Algorithm \ref{algo:adaptive-updates} largely follow the one in \adaofuecolog, which is based on the following inequality criterion:
\begin{equation}
\dot{\mu}(\mathbf{x}_t^\top \bar{\bm\theta}_t) \leq 2 \min\{\dot{\mu}(\mathbf{x}_t^\top \bm\theta^0_{t}), \dot{\mu}(\mathbf{x}_t^\top \bm\theta^1_{t})\}.
\label{equation:adaptivity-criterion}
\end{equation}
Here $\bar{\bm\theta}_t, \bm\theta_t^0, \bm\theta_t^1 \in \R^{dN}$, are $\mathcal{F}_t$-adapted parameters that enable adaptivity. They are obtained as follows:
\begin{equation}
    \bar{\bm\theta}_t = \argmin\limits_{\bm\theta \in \Theta_t} \sbrak{\eta \matnorm{\bm\theta - \bm\theta_t}{\mathbf{W}_t}^2 + \ell(\inner{\mathbf{x}_t}{\bm\theta} , 0) + \ell(\inner{\mathbf{x}_t}{\bm\theta} , 1)},
    \label{theta_bar}
\end{equation}

\begin{equation}
    {\bm\theta}^u_t = \argmin\limits_{\bm\theta \in \Theta_t} \sbrak{\eta \matnorm{\bm\theta - \bm\theta_t}{\mathbf{W}_t}^2 + \ell(\inner{\mathbf{x}_t}{\bm\theta} , u)},
    \label{theta_u}
\end{equation}
where $\ell(\mathbf{x},y) = -y\log \mu(\mathbf{x}) - (1-y) \log (1-\mu(\mathbf{x}))$ is the cross-entropy loss and $\eta = (2+diam(\Theta_t))^{-1}$. When the inequality in \eqref{equation:adaptivity-criterion} is true, the algorithm updates $\bm\theta_{t}, \mathbf{W}_t$ and $\mathbf{W}_t^i$ ($i\in [N]$) as per \emph{Steps 5-7} (Algorithm \ref{algo:adaptive-updates}). First, in \emph{Step 5}, $\bm\theta_{t+1}$ is computed by solving the following optimization problem up to a precision of $1/t$:
\begin{equation}
    \label{equation:optimization}
    \bm\theta_{t+1} = \arg \min\limits_{\Theta_t} \sbrak{\eta\matnorm{\bm\theta - \bm\theta_t}{\mathbf{W}_t}^2 + \ell(\inner{\mathbf{x}_t}{\bm\theta} , y_{t})}.
\end{equation}
Next, $\mathbf{W}_t^i$ ($i\in [N]$) and $\mathbf{W}_t$ are updated in \emph{Step 6} and \emph{Step 7}  as per their definitions provided earlier. When the inequality in \eqref{equation:adaptivity-criterion} does not hold, the algorithm instead updates $\mathcal{H}_t$ and $\Theta_t$ as described in \emph{Steps 10-13} (Algorithm \ref{algo:adaptive-updates}). In \emph{Step 10}, $\mathcal{H}_t$ is updated to $\mathcal{H}_{t+1}$ by adding the arm-reward pair $(\mathbf{x}_t, y_t)$ to it. Using $\mathcal{H}_{t+1}$, in \emph{Step 11}, the algorithm recomputes the estimate of $\bm\theta^\star$, given by $\bm\theta_{t+1}^\mathcal{H}$, by minimizing the regularized cross-entropy loss (up to a precision $1/t$) as follows:
\begin{equation}
\label{equation:optimization2}
    \bm\theta^\H_{t+1} = \argmin\sum\limits_{(\mathbf{x},y) \in \H_{t+1}} \ell(\inner{\mathbf{x}}{\bm\theta} , y) + \gamma_t(\delta)\twonorm{\bm\theta}^2.
\end{equation}
Using this estimate, and a design matrix $\mathbf{V}_t^{\mathcal{H}}$ (computed in \emph{Step 12}), in \emph{Step 13}, the set $\Theta_t$ is updated to $\Theta_{t+1}$ by taking an intersection between a confidence set of radius $\beta_t(\delta)= O(dN\log (t/\delta))$ around the new estimate $\bm\theta_{t+1}^{\mathcal{H}}$ (that contains the optimal parameter $\bm\theta^\star$ with high probability) and the initial set $\Theta_1 = \{\twonorm{\bm\theta}\leq S\}$. Updating $\Theta_t$ when the inequality in \eqref{equation:adaptivity-criterion} fails ensures that \eqref{equation:adaptivity-criterion} is more likely to succeed in the future.
In Lemma \ref{lemma: warmup}, we show that $|\mathcal{H}_T| = \tilde O(\kappa d^2 N^2 S^6)$. The rounds corresponding to $\mathcal{H}_T$, therefore, incur at most $\tilde O(\kappa d^2 N^2 S^6)$ regret. 

In Theorem \ref{theorem: Regret OFUL}, we provide a regret guarantee for \slateglincb\ and present its proof in Appendix \ref{appendix: proof_regret_oful}.

\begin{theorem}[Regret of \slateglincb]
\label{theorem: Regret OFUL}
Let $\mathcal{T} = \{s\in [T]: (\mathbf{x}_s, y_s)\in \mathcal{H}_T\}$, i.e, the set of rounds up till round $T$ where the inequality condition in \eqref{equation:adaptivity-criterion} fails. Also, let $\mathbf{x}_{\star, t} = \argmax_{\mathbf{x}\in \mathcal{X}_t} \mu(\mathbf{x}^\top\bm\theta^\star)$, be the optimal slate at round $t\in [T]$. Under the diversity assumption (Assumption \ref{assumption: diversity}), at the end of $T$ rounds, the regret incurred by \slateglincb, denoted by $R(T)$, can be bounded as
\[
R(T) = \tilde O\left(SdN\sqrt{\sum\limits_{t\notin \mathcal{T}} \dot{\mu}(\mathbf{x}_{\star, t}^\top \bm\theta^\star)} + S^6 d^2 N^2 \kappa \right).
\] 
\end{theorem}

\textbf{Remark: }Let $\mathcal{T}$ be as defined in Theorem \ref{theorem: Regret OFUL}. The per-round time complexity of \slateglincb\ is $O(d^2 N^2 \log^2 t)$ for rounds $t\in [T]\setminus \mathcal{T}$, and, $O(dNt)$ for rounds $t\in \mathcal{T}$. However, from Lemma \ref{lemma: warmup}, we have that the $O(dNt)$ per-round complexity is incurred only for $|\mathcal{T}| = \tilde{O}(\kappa d^2 N^2 S^6)$ rounds.

\section{\slateglincbts}
\label{section:TS}
In this section, we present our second algorithm, \slateglincbts\ (Algorithm \ref{algo:TS}) based on the Thompson Sampling paradigm \citep{Thompson1933, Russo2018} used in bandit algorithms. \slateglincbts\ builds upon the \tsecolog\ algorithm (Algorithm $3$ in Appendix $D.2$, \cite{Faury2022}) while adapting to the changing action sets using the update strategy in Algorithm \ref{algo:adaptive-updates}. \tsecolog\ adapts the Linear Thompson Sampling algorithm from \cite{Abeille2017} that perturbs the estimated parameter vector by adding an appropriately transformed noise vector sampled from a suitable multivariate distribution $\mathcal{D}^{TS}$ (refer Figure 1 in \cite{Abeille2017}) satisfying certain properties (Definition \ref{def: D_TS}). Following this, the optimal action (in our case, slate) with respect to the new perturbed parameter vector is chosen. While \tsecolog\ achieves an optimal regret guarantee of $O(\sqrt{T})$ for fixed action sets and a logistic reward model, similar to \adaofuecolog\ , it also incurs per-round computational cost proportional to the number of actions $K$ (recall $K = 2^{\Omega(N)}$ in our setting), since it selects arms at the slate-level. To circumvent this exponential time-complexity, \slateglincbts\ operates at the slot-level. For each slot $i\in [N]$, it perturbs the components of the estimated parameter vector (corresponding to the $i^{th}$ slot) using a noise vector sampled independently of all other slots. Then, it chooses the optimal item from each slot independent of the others, hence, choosing the slate with $N^{O(1)}$ per-round time complexity. While the items for each slot are determined independently, similar to \slateglincb, \slateglincbts\ also estimates a single reward model and updates the parameter vector for this model jointly using the slate-level reward $y_t$, by employing the update strategy in Algorithm \ref{algo:adaptive-updates}. 

\slateglincbts\ takes as inputs the time horizon $T$, the error probability $\delta$, $S$, a known upper bound for $\twonorm{\bm\theta^\star}$, and a multivariate distribution satisfying the properties stated in (Definition $1$, \cite{Abeille2017}), denoted by $\mathcal{D}^{TS}$.
During the course of the algorithm, \slateglincbts\ maintains vectors $\bm\theta_t$, matrices $\mathbf{W}_t$, $\mathbf{W}_t^i$ ($i\in [N]$) and sets $\Theta_t, \mathcal{H}_t$ with exactly the same definitions as in \slateglincb.

Next, we go through the steps of \slateglincbts\ (Algorithm \ref{algo:TS}).
\emph{Steps 3-10} is where \slateglincbts\ differs significantly from \tsecolog. Instead of getting the set of arm features $\mathcal{X}_t$ (in our case, slates) directly from the environment (as in \tsecolog), \slateglincbts\ receives $N$ different sets of items $\mathcal{X}_t^i, i\in [N]$ in \emph{Step 4}.
While \tsecolog\ samples a noise vector $\bm\eta\in \R^{dN}$ from $\mathcal{D}^{TS}$ and perturbs the estimated parameter vector $\bm\theta_t$ \
by adding (a scalar multiple of) $({\mathbf W_t})^{-1/2} \bm\eta$, \slateglincbts\ samples $N$ \textrm{i.i.d} vectors $\bm\eta_1,\ldots, \bm\eta_N$ and perturbs each component of $\bm\theta_t = (\bm\theta^1_t , \ldots , \bm\theta^N_t)$ as 
\[
    \tilde{\bm\theta}^i_t = \bm\theta^i_t + \eta_t(\delta) (\mathbf{W}^i_t)^{-\frac{1}{2}}\bm\eta^i,
\]
i.e, the $i^{th}$ component of $\bm\theta_t$ is perturbed by adding to it (a scalar multiple of) $({\mathbf W_t^i})^{-1/2} \bm\eta_i$ (\emph{Step 7} and \emph{8}).
Thus, by perturbing $\bm\theta_t$ , the algorithm performs rejection sampling till the perturbed vector $\tilde{\bm\theta}_t$ belongs to the set of admissible parameters $\Theta_t$.
Then, in \emph{Step 11}, it picks the item $\mathbf{x}_t^i \in \mathcal{X}_t^i$, that is optimal with respect to the perturbed parameter vector $\tilde{\bm\theta}_t^i$. Note that, the underlying optimization problem for slot $i$ only requires the candidate items in $\X_t^i$ and the perturbed vectors $\tilde{\bm\theta}_t^i$, and hence, can be solved independently and in parallel for all slots. In \emph{Step 12}, we select the slate $\mathbf{x}_t = (\mathbf{x}_t^1, \ldots, \mathbf{x}_t^N)$, yielding a reward $y_t$. At this point, \slateglincb\ calls a sub-routine described in Algorithm \ref{algo:adaptive-updates} which performs updates to $\bm\theta_t$, $\mathbf{W}_t$, $(\mathbf{W}_t^1, \ldots, \mathbf{W}_{t}^N)$, $\Theta_t$, $\mathcal{H}_t$. 

Now, we make a few additional remarks about \slateglincbts\ below.
\begin{algorithm}[!ht] 
\caption{\texttt{Slate-GLM-TS}}
\label{algo:TS} 
\begin{algorithmic}[1]
\STATE \textbf{Inputs:} Horizon $T$, probability of failure $\delta$, upper bound on parameter norm $S$, and a distribution $\mathcal{D}^{TS}$.

\STATE Initialize $\{\mathbf{W}^i_1\}_{i \in [N]} = \mathbf{I}_d$, $\mathbf{W}_1 = I_{dN}$, $\Theta_1 = \{\twonorm{\bm\theta}\leq S\}$, $\bm\theta_1 \in \Theta_1$, $\eta_t(\delta) = O(S^2Nd\log(t/\delta))$, and $ \H_1 = \emptyset$.
        
\FOR{each round $t \in [T]$}
\STATE Obtain the set of items $\mathcal{X}^i_t$, $\forall i \in [N]$.
\STATE Set reject = True.
\WHILE{reject}
    \STATE Sample $\mathbf{\eta}^{1}, \ldots, \mathbf{\eta}^N \overset{\mathrm{iid}}{\sim} \mathcal{D}^{TS}$. 
    \STATE Define $\tilde{\bm\theta}^{i}_t = \bm\theta^{i}_t + \eta_t(\delta)(\mathbf{W}_t^i)^{-1/2}\mathbf{\eta}^{i}$, $\forall i\in [N]$.
    \STATE If $\tilde{\bm\theta}_t = (\tilde{\bm\theta}^{1}_t , \ldots  \tilde{\bm\theta}^{N}_t) \in \Theta_t$, set reject = False.
\ENDWHILE
\STATE For each $i\in [N]$, find item $\mathbf{x}^i_t = \argmax_{\mathbf{x} \in \X^i_t} \inner{\mathbf{x}}{\tilde{\bm\theta}^{i}_t}$.

\STATE Select slate $\mathbf{x}_t = (\mathbf{x}^1_t, \ldots, \mathbf{x}^N_t)$ and obtain corresponding reward $y_t$.

\STATE Obtain $\bm\theta_{t+1}$, $\mathbf{W}_{t+1}$, $(\mathbf{W}_{t+1}^1, \ldots, \mathbf{W}_{t+1}^N)$, $\Theta_{t+1}$, $\mathcal{H}_{t+1}$ by calling Algorithm \ref{algo:adaptive-updates} with inputs $\mathbf{x}_t$, $y_t$, $\bm\theta_t$, $\mathbf{W}_t$, $(\mathbf{W}_t^1, \ldots, \mathbf{W}_{t}^N)$, $\Theta_t$, $\mathcal{H}_t$.
\ENDFOR
\end{algorithmic}
\end{algorithm}

\textbf{Remark: }It's easy to see that the per-round time complexity of \slateglincbts\ is $N(d\log T)^{O(1)}$. This is significantly lower than that of \texttt{TS-ECOLog}, which has time complexity exponential in $N$. The improvement comes as a result of the slot-level selection in \slateglincbts. This, along with the efficient estimation of $\bm\theta_t$ in Algorithm \ref{algo:adaptive-updates}, ensures that the algorithm has low per-round time complexity, making it useful in practical scenarios. This is validated by our Synthetic experiments in Section \ref{section:experiments}. Also, in several different instances, we observe that the regret incurred by \slateglincbts\ is competitive (and in most cases, better) than other baselines. 
Even though we do not provide a theoretical guarantee for the regret of \slateglincbts, in Appendix \ref{appendix: regret_proof_TS}, we provide a fixed-arm version for \slateglincbts\ called \slateglincbtsfixed. Like \tsecolog\ , \slateglincbtsfixed\ operates in the non-contextual setting, i.e., the action features (in our case, set of slates) remain fixed over time. It combines a short warm-up procedure from \tsecolog\ and the slot-level selection technique from \slateglincbts\, which results in a per-round time complexity that grows as $poly(N)$. Under the diversity assumptions (Assumption \ref{assumption: diversity}), we utilize a multiplicative equivalence between  $\mathbf W_t$ and $diag(\mathbf W_t^1, \ldots , \mathbf W_t^N)$ (similar to the one described in Section \ref{section:main-algo}) to prove an optimal $O(\sqrt{T})$ regret guarantee on the number of rounds $T$. For brevity, we discuss details of \slateglincbtsfixed\ (Algorithm \ref{algo:TS-Fixed}) and its regret guarantee (Theorem \ref{theorem:TS}) in Appendix \ref{appendix: regret_proof_TS}.

\section{Experiments}
\label{section:experiments}
In this section, we perform a wide range of synthetic (\textbf{Experiments 1,2,3})
and real-world experiments (\textbf{Experiment 4})
to demonstrate the empirical performance of our algorithms \slateglincb, \slateglincbts\ and \slateglincbtsfixed. The codes for Experiments 1-3 and Experiment 4 can be found \href{https://github.com/tanmaygoyal258/Logistic_Slate_Bandits.git}{here} and \href{https://github.com/tanmaygoyal258/Prompt_Optimization_Slate_Bandits.git}{here} respectively. We now detail the setup for each of the experiments:

\begin{figure*}
	\centering
	\begin{subfigure}[b]{0.33\columnwidth}  
		\centering 
		\includegraphics[width=56mm]{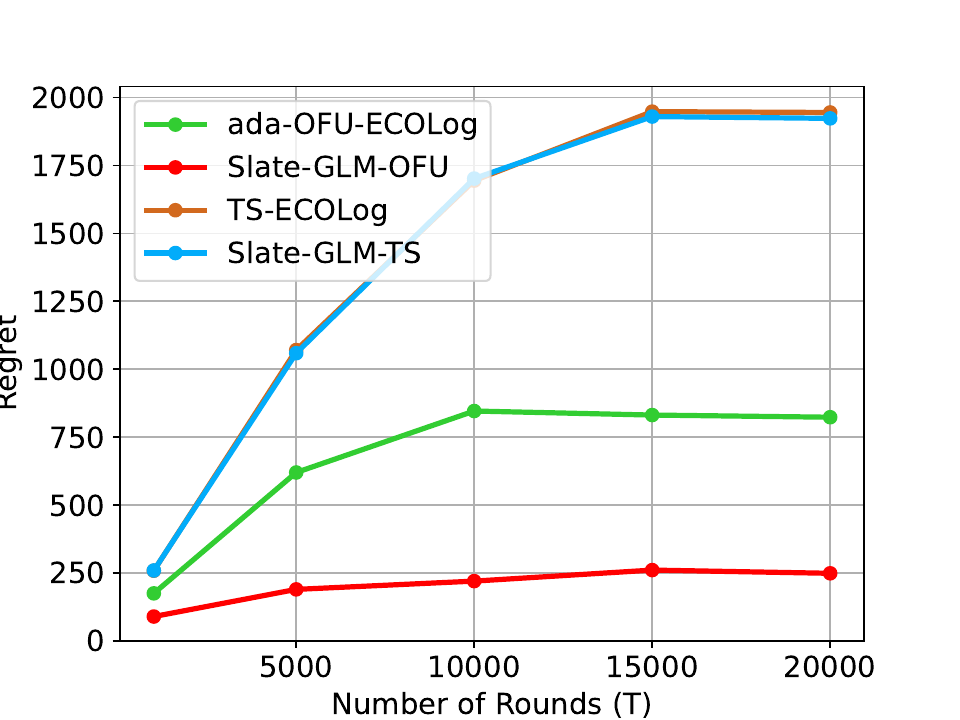}
		\caption{{\small Regret vs.\ $T$: Finite Context Setting}}   
		\label{fig:finite-context-logistic}
	\end{subfigure}
	\hfill
	\begin{subfigure}[b]{0.33\columnwidth}   
		\centering 
	\includegraphics[width=56mm]{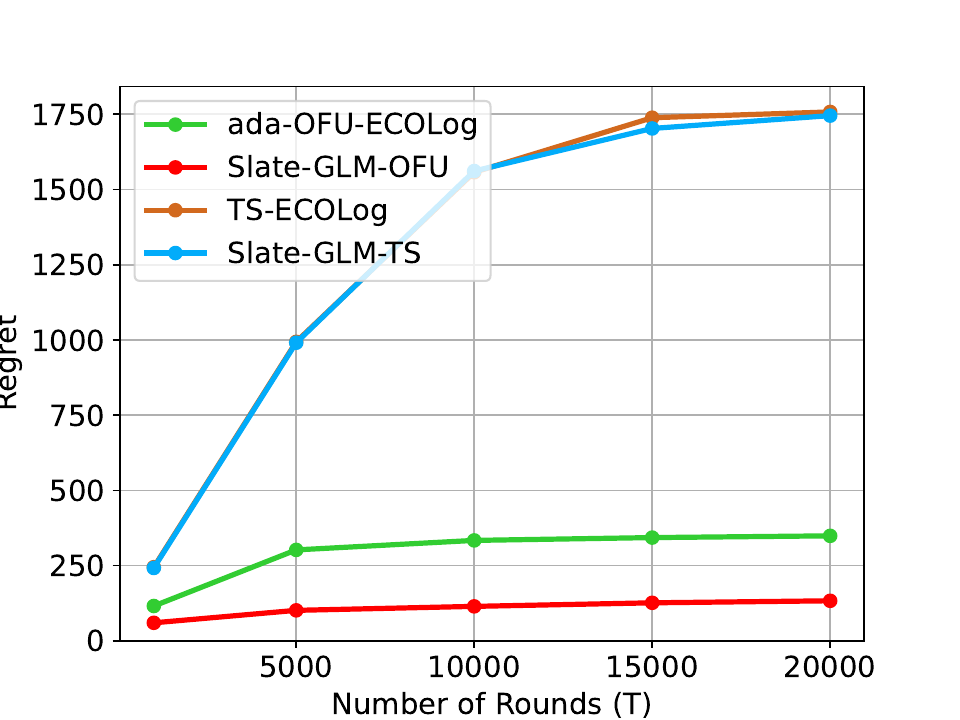}
		\caption{{\small Regret vs.\ $T$: Infinite Context Setting}}   
		\label{fig:infinite-context-logistic}
	\end{subfigure}
	\hfill
	\begin{subfigure}[b]{0.33\columnwidth}
		\centering
		\includegraphics[width=56mm]{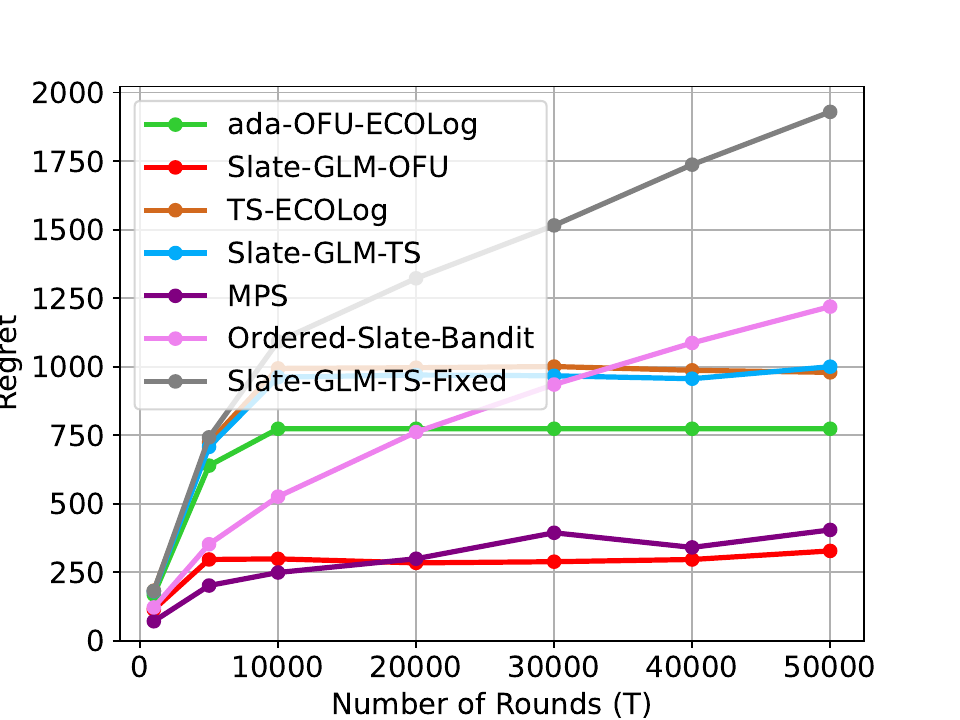}
		\caption{{\small Regret vs.\ $T$: Fixed-Arm Setting}}     
		\label{fig:non-contextual-logistic}
	\end{subfigure}
	\vskip\baselineskip
	\begin{subfigure}[b]{0.33\columnwidth}  
		\centering 
		\includegraphics[width=56mm]{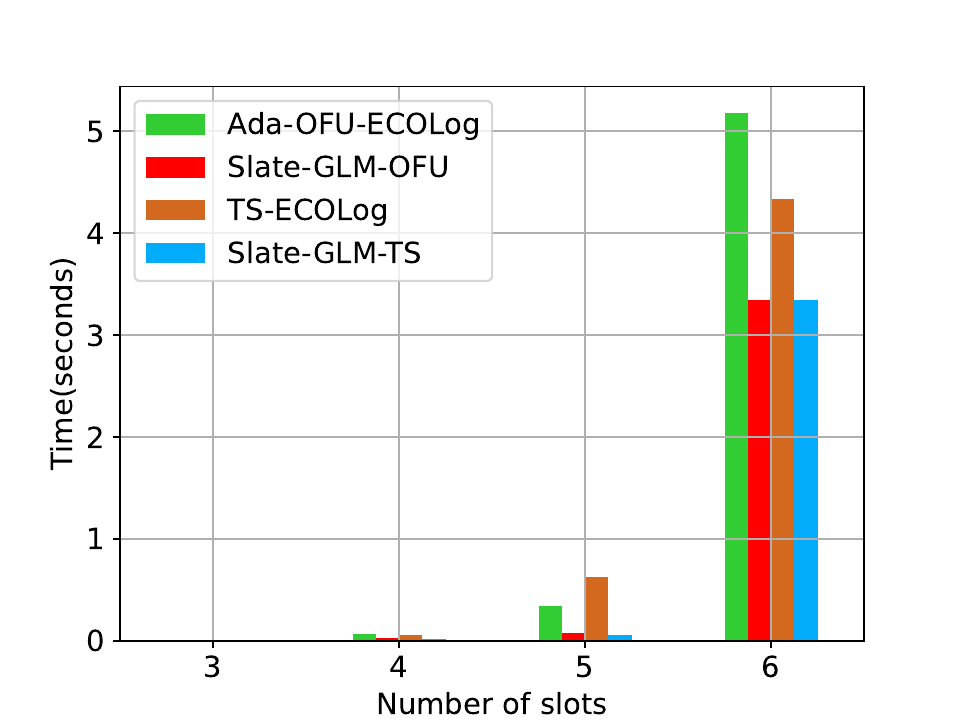}
		\caption{{\small Average running time (per-round)}}   
		\label{fig:average-time-per-round}
	\end{subfigure}
	\hfill
	\begin{subfigure}[b]{0.33\columnwidth}   
		\centering 
		\includegraphics[width=56mm]{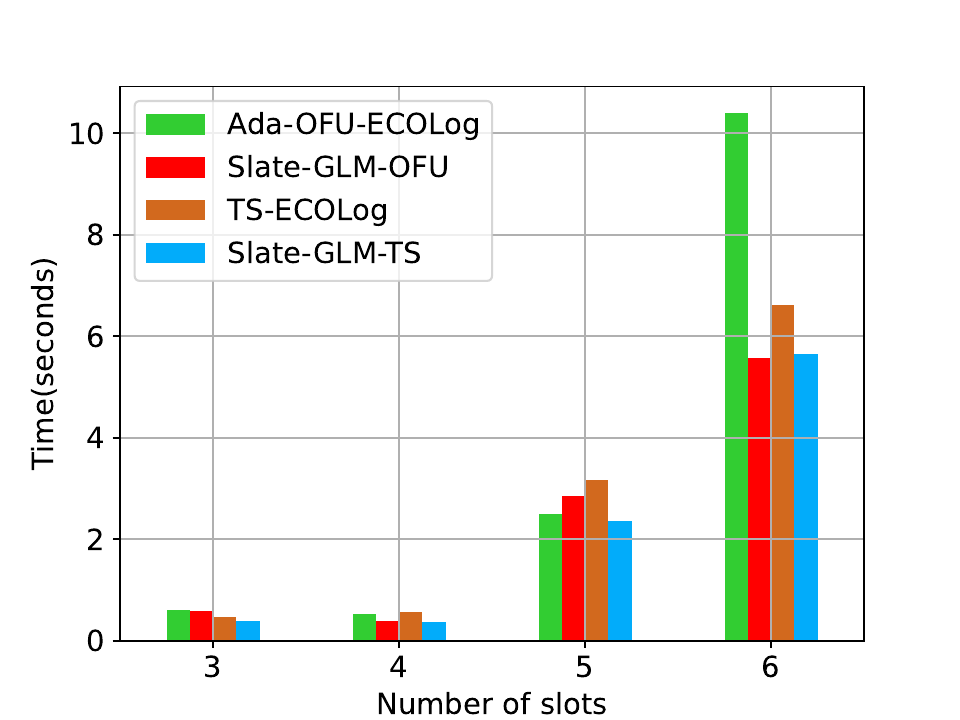}
				\caption{{\small Maximum running time (per-round)}}   
		\label{fig:maximum-time-per-round}
	\end{subfigure}
	\hfill
	\begin{subfigure}[b]{0.33\columnwidth}
		\centering
		\includegraphics[width=56mm]{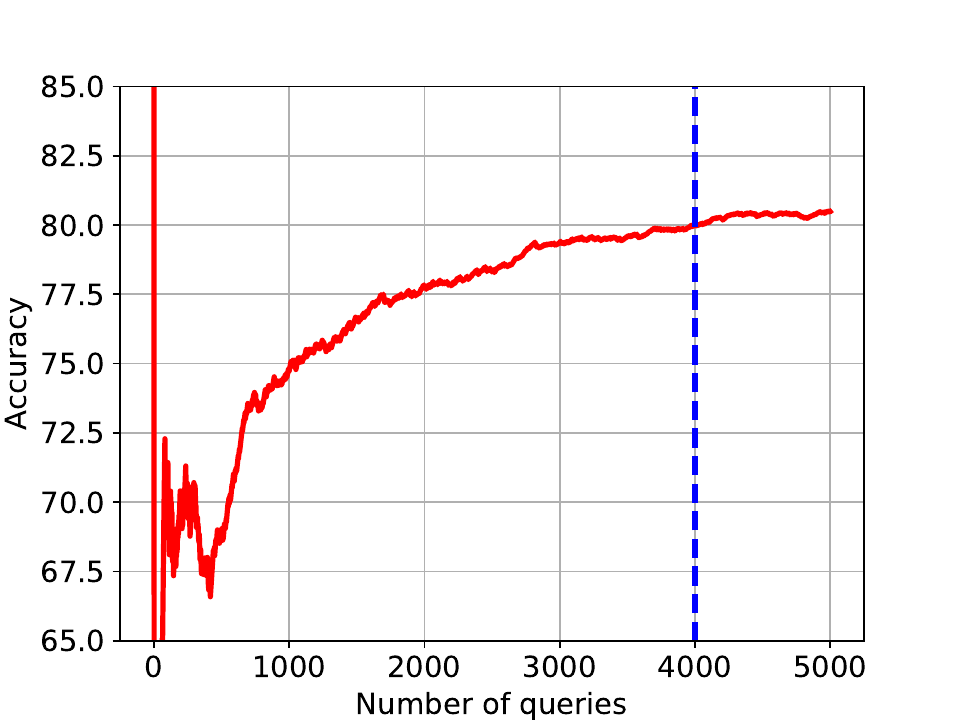}
		\caption[observationalAlgo]%
        {{\small Accuracy vs.\ $T$: Prompt Optimization}}   
		\label{fig:prompt_opt}
	\end{subfigure}
	\caption{}
    \label{fig:Plots}
\end{figure*}

\textbf{Experiment 1 ($R(T)$ vs.\ $T$, Contextual Setting):}  In this experiment, we compare our algorithms \texttt{Slate-GLM-OFU} and \texttt{Slate-GLM-TS} to their counterparts \texttt{ada-OFU-ECOLog} (Algorithm 2, \cite{Faury2022}) and \texttt{TS-ECOLog} (Section D.2, \cite{Faury2022}). To the best of our knowledge, these are the only logistic bandit algorithms that achieve optimal ($\kappa$-free) regret, while simultaneously being computationally efficient, with a per-round time complexity of $O(\log^2 T)$. We experiment in the following two settings:
\begin{itemize}
    \item \textbf{Finite Contexts: } We assume the contexts come from the finite set $\C = \{1,\ldots,C\}$. For each $c\in \C$ and $i\in [N]$, before beginning the algorithm, a set of items $\X^{i,c}$ is constructed by randomly sampling $K$ vectors from the $d$-dimensional ball with radius $1/\sqrt{N}$ . Then, during the course of the algorithm, at each round $t$, a context $c$ is sampled uniformly at random from $\C$, and the sets $\X^{1,c}, \ldots , \X^{N,c}$ are presented to the learner.
    
    \item \textbf{Infinite Contexts: } At each
        round $t\in [T]$, and for each slot $i\in [N]$, set $\X_t^i$ is constructed by sampling $K$ vectors randomly from the $d-$dimensional ball with radius $1/\sqrt{N}$. The learner is then presented with $\{\X_t^i\}_{i \in [N]}$.
\end{itemize}
For the finite context setting, we fix $C=5$. For both settings, we fix the number of slots $N=3$, the number of items per slot $K=5$, and the dimension of item features $d=5$. To simulate the reward, we randomly sample $\bm\theta^\star$ from $[-1,1]^{15}$. We run our algorithms by varying the time horizon $T$ in $\{1000,5000,10000,15000, 20000\}$. For each $T$, we average the regret obtained at the end of $T$ rounds over 20 different seeds used to sample the rewards. The results for the Finite and Infinite context settings are shown in Figures \ref{fig:finite-context-logistic} and \ref{fig:infinite-context-logistic} respectively. In both instances, \texttt{Slate-GLM-OFU} performs the best, while \texttt{Slate-GLM-TS} performs on par with \texttt{TS-ECOLog}. Further, in Appendix \ref{appendix:experiments}, we report the average results with two standard deviations.

\textbf{Experiment 2 (Per-Round Time vs.\ $N$): }
In this experiment, we compare the average and maximum (per-round) time taken by our algorithms \texttt{Slate-GLM-OFU} and \texttt{Slate-GLM-TS}, with respect to their counterparts \texttt{ada-OFU-ECOLog} and \texttt{TS-ECOLog} \citep{Faury2022} respectively\footnote{The per-round time is calculated as the sum of the per-round pull and per-round update times.}. We record the per-round time taken by all algorithms while varying the number of slots $N$ in the set $\{3,\ldots , 6\}$. The number of items $(K = |\mathcal{X}^i_t|)$ per slot is fixed to $7$ and the dimension $d$ of each item is fixed to $5$. The item features are selected by randomly sampling from $[-1,1]^5$ and normalized to have norm $1/\sqrt{N}$. For each $N\in \{3,4,5,6\}$, we select a different reward parameter vector $\bm\theta^\star$ by randomly sampling from $[-1,1]^{5N}$. Note that the number of possible slates is $K^N$ and thus, varying $N$ in $\{3,4,5,6\}$ results in $343$, \;$2401$, \;$16807$, and $117649$ slates respectively. We perform this experiment in the infinite context setting (See \textbf{Experiment 1} for details). We run all the algorithms for $T = 1000$ rounds and average the results over 10 different seeds for sampling rewards. We show the average per-round running time in Figure \ref{fig:average-time-per-round} and the maximum per-round running time in Figure \ref{fig:maximum-time-per-round}. As expected, we observe much lower running times for \slateglincb\ and \slateglincbts\ compared to their counterparts. Moreover, the plots also indicate near-exponential growth in the per-round running time for both \texttt{ada-OFU-ECOLog} and \texttt{TS-ECOLog}. Further, there is a significant gap between the maximum and average per-round time of \texttt{Slate-GLM-OFU} and \texttt{Slate-GLM-TS}, implying that the actual per-round time for these algorithms is generally much lower than their maximum values. In Appendix \ref{appendix:experiments}, we report these results with two standard deviations. We also report the per-round time required to choose an arm, as well as, update parameters, separately, for each of our algorithms.

\textbf{Experiment 3 ($R(T)$ vs.\ $T$, Non-Contextual Setting):} In this experiment, we compare our algorithms \texttt{Slate-GLM-OFU}, \texttt{Slate-GLM-TS}, and \texttt{Slate-GLM-TS-Fixed} (Algorithm \ref{algo:TS-Fixed}, Appendix \ref{appendix:ts-algos}) to several state-of-the-art baseline algorithms, in the non-contextual setting, i.e., the set of candidate slates remains fixed throughout the course of the algorithm. 
Like previous experiments, our baselines include \texttt{ada-OFU-ECOLog} and $\texttt{TS-ECOLog}$ from \cite{Faury2022}. However, for the non-contextual setting, we also include other state-of-the-art baselines such as the \texttt{MPS} algorithm (Algorithm 3, \cite{Dimakopoulou2019}) and the \texttt{Ordered Slate Bandit} algorithm (Figure 3, \cite{Kale2010}). The latter is designed for semi-bandit feedback, and hence, we adapt it to the bandit feedback setting as explained in Appendix \ref{appendix:experiments}. We fix the number of slots $N$ to $3$ and the number of items in each slot $|\mathcal{X}^i_t| = K$ to $5$. The dimension $d$ of items for each slot is fixed to $5$. The items for each slot are randomly sampled from $[-1,1]^5$ and normalized to have norm $1/\sqrt{3}$, while $\thetastar$ is randomly sampled from $[-1,1]^{15}$ and normalized. We run all the algorithms for $T \in \{1000,5000,10000,20000,30000,40000,50000\}$ rounds and average the results over 50 different seeds for sampling rewards. The rewards are shown in Figure \ref{fig:non-contextual-logistic}. We see that \texttt{Slate-GLM-OFU} has the best performance, with the only algorithm having comparable performance being \texttt{MPS}. Also, \texttt{Slate-GLM-TS} performs worse than \texttt{ada-OFU-ECOLog} and \texttt{MPS} while being on par with \texttt{TS-ECOLog}. In Section \ref{appendix:experiments}, we showcase the average results with two standard deviations, which show that \texttt{MPS} showcases high variance, and hence, is less reliable in practice.

\textbf{Experiment 4 (Prompt Tuning):} In this experiment, we apply our contextual slate bandit algorithm \slateglincb\ to select in-context examples for tuning prompts of Language Models, applied to binary classification tasks. Typically, for such applications, a labeled training set of (input query, output label) pairs is used to learn policies of editing different parts of the prompt (such as instruction, in-context examples, verbalizers) \citep{Tempera2022} depending on a provided test input query. To simplify our task, we fix the instruction and the verbalizer and only select $N$ in-context examples from an available pool of $K$ examples. There are $N$ available positions (slots) in the prompt. Given a test input query (context), we create context-dependent features for the $K$ pool examples and independently select one (with repetition) per slot. This matches the contextual slate bandit problem setting (See Section \ref{section:preliminaries}) and therefore \slateglincb\ can be applied. We experiment on a sampled subset of size $5000$ from two popular sentiment analysis datasets, \emph{SST2} and \emph{Yelp Review}.  We randomly order the set and use about $\sim80\%$ ($4128$ for \emph{SST2}, $4000$ for \emph{Yelp Review}) of them for ``warm-up'' training and the remaining ~$20\%$ for testing. Like most prompt tuning experiments \citep{Tempera2022}, we report our results only on the test set; however, our algorithm continues to learn throughout the $5000$ rounds. The warm-up rounds help in obtaining a good estimate of the hidden reward parameter vector. We fix $N = 4$ and vary $K$ in the set $\cbrak{8,16,32}$. All the slots choose an example from the same $K$-sized example pool. At each round, given an input query $\bm{q}$ that needs to be solved for, item features for each in-context example $\bm{e}=(\bm{x}, y)$, is constructed by embedding each of $\bm{q}$, $\bm{x}$, and $y$ into 64 dimensions \citep{nussbaum2024nomic} and concatenating them, thereby resulting in a $192$-dimensional item feature vector. After selecting the $4$ items (slate), the resulting prompt (also containing the input query $\bm{q}$) is passed through the RoBERTa \citep{Zhuang2021} model and a possible answer for $\bm{q}$ is generated. Hence, we are learning to choose the best in-context examples for RoBERTa. At each round, we use GPT-3.5-Turbo to provide feedback (binary, $0$ or $1$) for the generated answer. This is treated as the reward for the chosen slate and utilized by the rest of the \slateglincb\ algorithm.
Figure \ref{fig:prompt_opt} shows the increase in cumulative accuracy as we sequentially proceed through the $5000$ data points in the \emph{Yelp Review} dataset. The data points to the left of the dotted blue line are the warm-up points, and those to the right are the test points. We can see that the cumulative accuracy increases consistently as we sequentially proceed through the points. Also, on the test set, the accuracy stays well above $80\%$.
We vary $K$ in the set $\{8,16,32\}$ and report the test accuracy for both datasets in Table \ref{table:prompt_opt}. It can be seen that the cumulative test accuracies for \slateglincb\ are much higher compared to the Random Allocation baseline, where each in-context example is chosen randomly (and hence, there is no learning). Also, we see that the accuracy generally increases when the pool size increases, potentially because of the availability of better examples. We do see a small decrease in accuracy when $K$ increases from $16$ to $32$ for the \emph{Yelp Review} dataset, and potentially attribute this dip to higher exploration.

\begin{table}[H]
\centering
\def\arraystretch{1.0}%
\resizebox{0.6\columnwidth}{!}{
\begin{tabular}{ccccc}
\hline
\multirow{2}{*}{\begin{tabular}[c]{@{}c@{}}Pool \\Size\end{tabular}} & \multicolumn{2}{c}{\textbf{SST2}}                                                                                                   & \multicolumn{2}{c}{\textbf{Yelp Review}}                                         \\ \cline{2-5} 
& \multicolumn{1}{c}{Random}            & \multicolumn{1}{c}{\texttt{Slate-GLM-OFU}}           & \multicolumn{1}{c}{Random}            & \multicolumn{1}{c}{\texttt{Slate-GLM-OFU}} \\ \hline
\multicolumn{1}{c}{8}           & \multicolumn{1}{c}{54.22}               & \multicolumn{1}{c}{69.15} & \multicolumn{1}{c}{62.90}               & \multicolumn{1}{c}{74.00} \\ 
\multicolumn{1}{c}{16}           & \multicolumn{1}{c}{54.46}               & \multicolumn{1}{c}{80.96} & \multicolumn{1}{c}{63.30}               & \multicolumn{1}{c}{82.50} \\ 
\multicolumn{1}{c}{32}           & \multicolumn{1}{c}{53.82}               & \multicolumn{1}{c}{81.42} & \multicolumn{1}{c}{62.00}               & \multicolumn{1}{c}{79.50} \\ \hline
\end{tabular}
}
\caption{Prompt Tuning Test Accuracy}
\label{table:prompt_opt}
\end{table}
\section{Conclusions}
\label{section:conclusions}

In this paper, we proposed three algorithms: \slateglincb, \slateglincbts, \slateglincbtsfixed\ for the slate bandit problem with logistic rewards. While the first two algorithms are designed for both the contextual and non-contextual settings, the third is designed particularly for the non-contextual setting. All our algorithms perform explore-exploit at the slot level, making their average per-round time complexity logarithmic in the number of candidate slates. By building on algorithms from \cite{Faury2022}, the average per-round time is also logarithmic in the number of rounds $T$. As a result, our algorithms run much faster than state-of-the-art logistic bandit algorithms (having $2^{\Omega(N)}$ per round time complexity). We also show that under a popular diversity assumption (Assumption \ref{assumption: diversity}), the consequences of which we validate in practice (Section \ref{appendix:empirical-validation}), \slateglincb\ and \slateglincbtsfixed\ achieve $\kappa$-independent  $\tilde{O}(\sqrt T)$ regret, making them both optimal and computationally efficient.

\bibliography{main}

\newpage
\appendix
\appendixtitle

\section{General Results and Notations}
\label{appendix:notation}

In this section, we first present some general results about the logistic function that are integral to our regret analysis proofs. Subsequently, we discuss some notations that would be used throughout the appendix.

The logistic function satisfies an important \emph{self-concordance} property, which is critical to our regret bounds. Informally, this property claims that the gradients of the logistic function at two different points are multiplicatively equivalent, and this multiplicative factor grows exponentially with the absolute distance between the two points. We now present the formal claim below:

\begin{claim}
    Let $\mu:\R\rightarrow\R$ be the logistic function, defined as $\mu(x) = 1/(1+\exp(-x))$, and $\dot{\mu}, \ddot{\mu}$ be the first and second derivative of $\mu$. Then, the following are true:
    \begin{enumerate}
        \item $|\ddot{\mu}(x)| \leq \dot{\mu}(x)$, $\forall x\in \R$,
        \item $\dot{\mu}(x) \leq \dot{\mu}(y) \exp(|x-y|)$, $\forall x,y\in \R$.
    \end{enumerate}
    \label{claim:self-concordance}
\end{claim}

Now, we present the exact Taylor expansion of the logistic function:

\begin{definition}
    (Exact Taylor Expansion for the logistic function) The logistic function $\mu(x)$ can be expanded using an exact Taylor Expansion as follows:
    \[
    \mu(x) = \mu(y) + \dot{\mu}(y) (x-y) +  (x-y)^2 \int\limits_0^1 (1-v)\ddot{\mu}(x+v(y-x)) \diff v .
    \]
    \label{def:Exact_Taylor}
\end{definition}

Next, we define the functions: $\alpha : \R \times \R \mapsto \R$, and $\tilde\alpha : \R \times \R \mapsto \R$ as follows:
\[
    \alpha(x,y) = \int\limits_{0}^1 \dot{\mu}(x+v(y-x)) \diff v \quad \text{ and } \quad  \tilde\alpha(x,y) = \int\limits_0^1 (1-v)\dot{\mu}(x+v(y-x)) \diff v.
\]

Using the Mean Value Theorem, we can expand the logistic function as follows:
\[
    \mu(x) = \mu(y) + \alpha(x,y) (x-y).
\]

Similarly, combining Definition \ref{def:Exact_Taylor} with the definition of $\tilde\alpha$ and the self-concordance property (Claim \ref{claim:self-concordance}), we get
\[
    \mu(x) = \mu(y) + \dot\mu(y)(x-y) + \tilde\alpha(x,y)(x-y)^2.
\]
Now, we present some general notations that shall be used throughout the appendix. Throughout, we denote the maximum and minimum eigenvalues of a matrix $\mathbf{A}$ as $\lambda_{\max}(\mathbf{A})$ and $\lambda_{\min}(\mathbf{A})$ respectively. Similarly, we define the maximum and minimum singular values of a matrix $\mathbf{A}$ as $\sigma_{\max}(\mathbf{A})$ and $\sigma_{\min}(\mathbf{A})$ respectively.

Following \cite{Faury2022}, we define the following important scalar functions:
\[
    \gamma_t(\delta) = O(S^2Nd\log(t/\delta)),
\]
\[
     \beta_t(\delta) = O(S^6 Nd \log(t/\delta)),
\]
\[  
    \eta_t(\delta) = O(S^2Nd\log(t/\delta)).
\]

Note that $\gamma_t(\delta)$, $\beta_t(\delta)$, and $\eta_t(\delta)$ are all monotonically increasing functions of $t$. Hence, we shall always trivially bound $\gamma_t(\delta) \leq \gamma_T(\delta)$, $\beta_t(\delta) \leq \beta_T(\delta)$, and $\eta_t(\delta) \leq \eta_T(\delta)$, respectively, throughout the proofs.

Now, following the notation in Algorithm \ref{algo:adaptive-updates}, let $\mathcal{T}$ define the set of all indices where the data-dependent condition (\emph{Step 4}, Algorithm \ref{algo:adaptive-updates}) fails, and let $\mathcal{H}_t$ consist of  thearm-reward pairs corresponding to these time rounds upto time $t$. We define the slate-level and slot-level design matrices for the rounds $s \in \mathcal{T}$ and all slots $i \in [N]$ as follows:
\[
    \mathbf{V}_t^{\mathcal{H}} = \gamma_t(\delta)\mathbf{I}_{Nd} + \sum\limits_{(\mathbf{x},y) \in \mathcal{H}_t} \frac{\mathbf{x}\mathbf{x}^\top}{\kappa}  \quad \text{ and } \quad 
    \mathbf{V}_t^{\mathcal{H},i} = \gamma_t(\delta)\mathbf{I}_{d} + \sum\limits_{(\mathbf{x},y) \in \mathcal{H}_t} \frac{\mathbf{x}^i{\mathbf{x}^i}^\top}{\kappa}.
\]
For the sake of simplicity, we shall define the block diagonal matrix consisting of $\mathbf{V}^{\mathcal{H},i}_t$ for all slots as 
\[
    \mathbf{U}^{\mathcal{H}}_t = \textrm{diag}(\mathbf{V}_t^{\mathcal{H},1}, \ldots, \mathbf{V}_t^{\mathcal{H},N}).
\]

Also, another set of matrices that shall arise throughout the proofs involves the outer product of the items chosen in two different slots at time rounds $s \in \mathcal{T} \cap [t-1]$. We define for all $i,j \in [N]$
\[
    \mathbf{V}_t^{\mathcal{H},i,j} = \sum\limits_{(\mathbf{x},y) \in \mathcal{H}_t} \frac{\mathbf{x}^i{\mathbf{x}^j}^\top}{\kappa} .
\]

When the data-dependent condition (\emph{Step 4}, Algorithm \ref{algo:adaptive-updates}) succeeds, we maintain \emph{online} proxies of the concentration matrix defined in Section \ref{section:main-algo}. We define the slate-level and slot-level online matrices for all rounds $s \in [t-1]\setminus \mathcal{T}$ and all slots $i \in [N]$ as
\[
    \mathbf{W}_t = \mathbf I_{Nd} + \sum\limits_{s \in [t-1] \setminus \mathcal{T}} \dot{\mu}(\mathbf{x}_s^\top\bm\theta_{s+1}) \mathbf{x}_s \mathbf{x}_s^\top \quad \text{ and } \quad \mathbf{W}_t^i = \mathbf{I}_d + \sum\limits_{s \in [t-1] \setminus \mathcal{T}} \dot{\mu}(\mathbf{x}_s^\top\bm\theta_{s+1})\mathbf{x}_s^i {\mathbf{x}_s^i}^\top.
\]

We also define the block-diagonal matrix comprising $\mathbf{W}^i_t$ for all slots as 
\[
    \mathbf{U}_t = \textrm{diag}(\mathbf{W}^1_t , \ldots , \mathbf{W}^N_t).
\]

Finally, another matrix that arises throughout the proof involves outer products of the items chosen in two different slots, similar to $\mathbf{V}^{\mathcal{H},i,j}$, which we define, for all rounds $s \in [t-1]\setminus \mathcal{T}$ and all slots $i,j\in [N]$ as
\[
    \mathbf{W}^{i,j}_t = \sum_{s \in [t-1] \setminus \mathcal{T}} \mathbf{x}^i_s {\mathbf{x}^j_s}^\top.
\]

\clearpage
\section{Regret Analysis for \texttt{Slate-GLM-OFU}}
As defined in Section \ref{section:preliminaries}, let $\mathbf{x}^i\in \R^{d}$ be the item chosen in the $i^{th}$ slot, then, we define the ``\emph{lift}``
 of $\mathbf{x}^i$, denoted by $\tilde{\mathbf{x}}^i \in \R^{dN}$, as follows,
 
\[   
\tilde{\mathbf{x}}^i(j) = 
     \begin{cases}
       0 &\quad \text{if } j\notin [(i-1)d, id-1]\\
       \mathbf{x}(j - (i-1)d) &\quad\text{otherwise}\\ 
     \end{cases}.
\]

In other words, consider $\tilde{\mathbf{x}}^i$ to be a vector with $N$ slots of dimension $d$, such that the $i^{th}$ slot comprises the item $\mathbf{x}^i$ while the rest of the slots are set to the zero vector. Then, for any vector $\mathbf{z} = (\mathbf{z}^1, \ldots, \mathbf{z}^N) \in \R^{dN}$, such that $\mathbf{z}^i\in \R^d \; \forall i\in [N]$, we get that $\mathbf{z} = \tilde{\mathbf{z}}^1 + \ldots +\tilde{\mathbf{z}}^N$.


Also, define the constant $T_0 = \tilde{O}(N^3 \rho^{-1})$, where the exact expression for $T_0$ can be found in Lemma \ref{lemma : norm of slot design matrix}. We now restate the regret guarantees for \texttt{Slate-GLM-OFU}, before proving the same.







\begin{theorem}[Regret of \slateglincb] Let $T \geq \tilde{O}(T_0 + \kappa d^2 N^2 S^6)$. Then,  with probability at least $1 - 6\delta$, after $T$ rounds, the regret of \slateglincb\ can be bounded as
 \begin{align*}
   Regret(T) &\leq  \tilde{O}(\kappa d^2 N^2 S^6 + N^3 \rho^{-1}) + C \kappa (1 + \kappa) S^2N^2d\log \left(\frac{T}{\delta} \right) \pbrak{ d \log \left(\frac{T}{4N} \right) + \frac{1}{\rho}\log(T)}
   \\
   &+ CS N d^{1/2}\sqrt{ \left(d \log \left(\frac{T}{4N} \right) + \frac{1}{2\rho}\log T \right)\log \left(\frac{T}{\delta} \right)} \sqrt{\sum\limits_{t\notin\mathcal{T}}{}\sens{{\mathbf{x}_t^\star}}{\bm\theta^\star}}.
    \end{align*}
\label{appendix: proof_regret_oful}
\end{theorem}

\begin{proof}

Denote $\mathbf{x}_t^\star = \arg\max_{\mathbf{x} \in \mathcal{X}_t} \mu(\mathbf{x}^\top \bm\theta^\star)$. Also, recall from Section \ref{section:main-algo} the definition of $\mathcal{T} \subset [T]$; $\mathcal
{T}$ consists of the time rounds where the data-dependent condition (\emph{Step 4}, Algorithm \ref{algo:adaptive-updates}) fails. For each $t \in \mathcal{T}$, we can trivially upper bound the regret by $1$, resulting in

\begin{align*}
    Regret(T) &\leq |\mathcal{T}| +  
    \sum\limits_{t\notin\mathcal{T}}
    \mu({\mathbf{x}_t^\star}^\top \bm\theta^\star)
 - \mu(\mathbf{x}_t^\top \bm\theta^\star).
\end{align*}

Now, we can only invoke Lemma \ref{lemma : norm of slot design matrix} for rounds $t$ such that $|[t-1] \setminus \mathcal{T}| \geq T_0$. Let $t^\prime$ be a time round such that $t^\prime = T_0 + \tilde{O}(\kappa d^2N^2 S^6)$. Such a $t^\prime$ exists because $T \geq \tilde{O}(T_0 + \kappa d^2 N^2 S^6)$. Thus, for any round $s \geq t^\prime$, we have $|[s-1] \setminus \mathcal{T}| \geq T_0$ since $|\mathcal{T}| \leq \tilde{O}(\kappa d^2 N^2 S^6)$ (Lemma \ref{lemma: bounds on V for failing data dependent condition}).

A trivial upper bound of $1$ on the regret of rounds $t \in [t^\prime] \setminus \mathcal{T}$, along with the definition of $t^\prime$ results in 
\[
 Regret(T) \leq 2|\mathcal{T}| +  T_0 + 
    \underbrace{\sum\limits_{t \in [t^\prime,T] \setminus \mathcal{T}}
    \mu({\mathbf{x}_t^\star}^\top \bm\theta^\star)
 - \mu(\mathbf{x}_t^\top \bm\theta^\star)}_{R(T)}.
\]



 Using Definition \ref{def:Exact_Taylor} and Claim \ref{claim:self-concordance}, we can expand $R(T)$ as:

\begin{align*}
    R(T) &\leq \underbrace{\sum\limits_{t \in [t^\prime,T] \setminus \mathcal{T}} \dot{\mu}(\mathbf{x}_t^\top \bm\theta^\star)(\mathbf{x}_t^\star - \mathbf{x}_t)^\top \bm\theta^\star}_{R_1(T)}  +  \underbrace{\sum\limits_{t \in [t^\prime,T] \setminus \mathcal{T}} \tilde{\alpha}({\mathbf{x}_t^\star}^\top\bm\theta^\star, \mathbf{x}_t^\top \bm\theta^\star) ((\mathbf{x}_t^\star - \mathbf{x}_t)^\top \bm\theta^\star)^2}_{R_2(T)}.
\end{align*}

We now bound $R_1(T)$ and $R_2(T)$ separately.

\underline{\textbf{Bounding $R_1(T)$: }}

Define the following sets:
\[
    \mathcal{T}_1 = \{t \in [ t^\prime, T] \setminus \mathcal{T}: \sens{\mathbf{x}_t}{\bm\theta^\star} \geq \sens{\mathbf{x}_t}{\bm\theta_{t+1}}\} \quad \text{ and } \quad  \mathcal{T}_2 = \{t \in [ t^\prime, T] \setminus \mathcal{T}: \sens{\mathbf{x}_t}{\bm\theta^\star} < \sens{\mathbf{x}_t}{\bm\theta_{t+1}}\}.
\]
    
Then, it is easy to see that $\mathcal{T}_1$ and $\mathcal{T}_2$ are disjoint partitions of the time indices in $[t^\prime , T]\setminus \mathcal{T}$.

\clearpage
Let $z_t \in [\mathbf{x}_t^\top \bm\theta^\star, \mathbf{x}_t^\top \bm\theta_{t+1}]$. Then, summing $R_1(T)$ over all indices in $\mathcal{T}_1$, we get

\begin{align*}
   &\sum\limits_{t \in \mathcal{T}_1}{} \sens{\mathbf{x}_t}{\bm\theta^\star}\inner{\pbrak{\mathbf{x}_t^\star - \mathbf{x}_t}}{\bm\theta^\star}
   \overset{}{=} \underbrace{\sum\limits_{t \in \mathcal{T}_1}{} \sens{\mathbf{x}_t}{\bm\theta_{t+1}} \inner{\pbrak{\mathbf{x}_t^\star - \mathbf{x}_t}}{\bm\theta^\star}}_{R_1(T)_1} + 
   \underbrace{\sum\limits_{t \in \mathcal{T}_1}{} \ddot{\mu}\pbrak{z_t}\pbrak{\inner{\mathbf{x}_t}{\bm\theta^\star} - \inner{\mathbf{x}_t}{\bm\theta_{t+1}}}\inner{\pbrak{\mathbf{x}_t^\star - \mathbf{x}_t}}{\bm\theta^\star}}_{R_1(T)_2}.
\end{align*}

where the equality follows from the mean-value theorem.

We first bound $R_1(T)_1$ as follows:

\begin{align*}
    R_1(T)_1 & = \sum\limits_{t \in \mathcal{T}_1} \dot{\mu}(\mathbf{x}_t^\top \bm\theta_{t+1}) \left( \mathbf{x}_t^\star - \mathbf{x}_t \right)^\top \bm\theta^\star
    \\
    &\overset{}{\leq} \sum\limits_{t \in \mathcal{T}_1}{}  \dot{\mu}(\mathbf{x}_t^\top \bm\theta_{t+1}) \left\{ \lvert {\mathbf{x}_t^\star}^\top \bm\theta^\star - {\mathbf{x}_t^\star}^\top \bm\theta_t \rvert + \lvert \mathbf{x}_t^\top \bm\theta^\star - \mathbf{x}_t^\top \bm\theta_t \rvert + {\mathbf{x}_t^\star}^\top \bm\theta_t - \mathbf{x}_t^\top \bm\theta_t \right\}
    \\
    &\overset{}{\leq} \sum\limits_{t \in \mathcal{T}_1}{}  \dot{\mu}(\mathbf{x}_t^\top \bm\theta_{t+1}) \left\{ \sum_{i=1}^N   \sqrt{\eta_t(\delta)} \left(\lVert \mathbf{x}_t^{\star,i}\rVert_{(\mathbf{W}^i_t)^{-1}}  + \lVert \mathbf{x}^i_t \rVert_{(\mathbf{W}^i_t)^{-1}} \right) + \sum_{i=1}^N \left( \tilde{\mathbf{x}}^{\star,i}_t - \tilde{\mathbf{x}}^i_t \right)^\top \bm\theta^i_t \right\}.
    \\
\end{align*}

Here, the last inequality follows by applying the Cauchy-Schwarz inequality, and subsequently, applying Lemma \ref{lemma: conversion of norms} and Lemma \ref{lemma: ada_ofu_ecolog_confidence_bound}. Now, using the action-selection rule for each slot $i \in [N]$, we know that 

\[
    {\mathbf{x}^i_t}^\top \bm\theta_t^i + \sqrt{\eta_t(\delta)}\matnorm{\mathbf{x}^{i}_t}{(\mathbf{W}^i_t)\inv} \geq {\mathbf{x}^{\star,i}_t}^\top \bm\theta_t^i + \sqrt{\eta_t(\delta)}\matnorm{\mathbf{x}^{\star,i}_t}{(\mathbf{W}^i_t)\inv},
\]

and hence, summing over all $i \in [N]$ and rearranging gives us

\[
    \sqrt{\eta_t(\delta)} \sum_{i=1}^N \left( \matnorm{\mathbf{x}^{i}_t}{(\mathbf{W}^i_t)\inv} - \matnorm{\mathbf{x}^{\star,i}_t}{(\mathbf{W}^i_t)\inv} \right) \geq \sum_{i=1}^N \left({\mathbf{x}^{\star,i}_t} - {\mathbf{x}^i_t} \right)^\top \bm\theta_t^i
\]

Thus, we have

\begin{align*}
    R_1(T)_1 &\overset{}\leq C\sqrt{\eta_T(\delta)}\sum\limits_{t\in \mathcal{T}_1}\dot{\mu}(\mathbf{x}_t^\top\bm\theta_{t+1})\sum\limits_{i=1}^N \|\mathbf{x}_t^i\|_{(\mathbf{W}_t^i)^{-1}} 
     \\
     &\overset{}{\leq} C\sqrt{\eta_T(\delta)}  \sqrt{\sum\limits_{t \in \mathcal{T}_1}{}\sens{\mathbf{x}_t}{\bm\theta_{t+1}}}\sqrt{\sum\limits_{t \in \mathcal{T}_1}{}\left(\sum\limits_{i=1}^{N} \sqrt{\sens{\mathbf{x}_t}{\bm\theta_{t+1}}} \matnorm{\mathbf{x}^i_t}{(\mathbf{W}^i_t)\inv} \right)^2}\\
     &\overset{}{\leq}C\sqrt{\eta_T(\delta)}  \sqrt{\sum\limits_{t \in \mathcal{T}_1}{}\sens{\mathbf{x}_t}{\bm\theta_{t+1}}}\sqrt{ Nd\log(T/4N) + \underbrace{\sum\limits_{t \in \mathcal{T}_1}{}\sum\limits_{i=1}^{N}\sum\limits_{\substack{j=1 \\ j \neq i}}^{N}\sens{\mathbf{x}_t}{\bm\theta_{t+1}}\matnorm{\mathbf{x}^i_t}{(\mathbf{W}^i_t)\inv}\matnorm{\mathbf{x}^j_t}{(\mathbf{W}^j_t)\inv}}_{M(T)}}
     \\
     &\overset{}{\leq} CSN^{1/2}d^{1/2}\sqrt{ Nd\log(T/4N) + M(T)} \sqrt{\log(T/\delta)} \pbrak{\sqrt{R(T)} + \sqrt{\sum\limits_{t\notin\mathcal{T}}{}\sens{{\mathbf{x}_t^\star}}{\bm\theta^\star}}}.
\end{align*}

The second-to-last inequality is a direct application of Lemma \ref{lemma: elliptical potential lemma} to $\sqrt{\dsigmoid{\inner{\mathbf{x}_t}{\mathbf{\bm\theta_{t+1}}}}}\mathbf{x}^i_t$ using the fact that $\twonorm{\sqrt{\dsigmoid{\inner{\mathbf{x}_t}{\mathbf{\bm\theta_{t+1}}}}}\mathbf{x}^i_t} \leq \frac{1}{2\sqrt{N}}$, while the last inequality follows from the definition of $\mathcal{T}_1$, Lemma \ref{Lemma: Abielle result}, and the definition of $\eta_t(\delta)$.

We now bound $M(T)$ as follows:
\begin{align*}
    M(T) &= \sum\limits_{t \in \mathcal{T}_1}{}\sum\limits_{i=1}^{N}\sum\limits_{\substack{j=1 \\ j \neq i}}^{N}\sens{\mathbf{x}_t}{\bm\theta_{t+1}}\matnorm{\mathbf{x}^i_t}{(\mathbf{W}^i_t)\inv}\matnorm{\mathbf{x}^j_t}{(\mathbf{W}^j_t)\inv}
    \\
    &\overset{}{\leq}  \sum\limits_{t \in \mathcal{T}_1}{}\sum\limits_{i=1}^{N}\sum\limits_{\substack{j=1 \\ j \neq i}}^{N} \frac{1}{4N \sqrt{\eigmin {\mathbf{W}^i_t}\eigmin{\mathbf{W}^j_t}}}
    \\
    &\overset{}{\leq} \frac{N^2}{4N}\sum\limits_{t \in \mathcal{T}_1}{}\frac{1}{1 + 0.5 \rho |[t-1] \setminus \mathcal{T}|}
    \\
    &\overset{}{\leq} \frac{N}{2\rho}\log(T).
\end{align*}
Here, the first inequality follows from Rayleigh's Quotient and the fact that $\twonorm{
\mathbf{x}^i_t} \leq\frac{1}{4N}$, the second inequality follows from a direct application of Lemma \ref{lemma : norm of slot design matrix}, and the last inequality follows from the sum of Harmonic Series.

Substituting this back, we get the bound on $R_1(T)_1$ as:
\[R_1(T)_1 \leq CS N d^{1/2}\sqrt{ \left(d\log(T/4N) + \frac{1}{2\rho}\log T \right)\log(T/\delta)} \pbrak{\sqrt{R(T)} + \sqrt{\sum\limits_{t\notin\mathcal{T}}{}\sens{{\mathbf{x}_t^\star}}{\bm\theta^\star}}}.\]

The bound on $R_1(T)_2$ is as follows:
\begin{align*}
    R_1(T)_2 &= \sum\limits_{t \in \mathcal{T}_1}{} \ddot{\mu}(z_t)\pbrak{\inner{\mathbf{x}_t}{\bm\theta^\star} - \inner{\mathbf{x}_t}{\bm\theta_{t+1}}}\inner{\pbrak{{\mathbf{x}_t^\star} - \mathbf{x}_t}}{\bm\theta^\star} 
    \\
    &\overset{}{\leq} C\sqrt{\eta_t(\delta)}\sum\limits_{t \in \mathcal{T}_1}{}\modulus{\pbrak{\inner{\mathbf{x}_t}{\bm\theta^\star} - \inner{\mathbf{x}_t}{\bm\theta_{t+1}}}}\sum\limits_{i=1}^{N}\matnorm{\mathbf{x}^i_t}{(\mathbf{W}^i_t)\inv}.
\end{align*}

where the inequality follows from the fact that $\lvert \ddot{\mu}(.) \rvert \leq 1$, and steps similar to the one used in bounding $R_1(T)_1$. 

We continue bounding $R_1(T)_2$ as follows:

\begin{align*}
    R_1(T)_2 &\overset{}{\leq} C\sqrt{\eta_t(\delta)}\sum\limits_{t \in \mathcal{T}_1}{}\pbrak{\sum\limits_{i=1}^{N}\matnorm{\tilde{\mathbf{x}}^i_t}{\mathbf{W}_t\inv}\matnorm{\bm\theta^\star - \bm\theta_{t+1}}{\mathbf{W}_t}}\sum\limits_{i=1}^{N}\matnorm{\mathbf{x}^i_t}{(\mathbf{W}^i_t)\inv}
    \\
    &\overset{}{\leq} C \eta_t(\delta)\sum\limits_{t \in \mathcal{T}_1}{}\pbrak{\sum\limits_{i=1}^{N}\matnorm{\mathbf{x}^i_t}{(\mathbf{W}^i_t)\inv}}^2
    \\
    &\overset{}{\leq} C \eta_t(\delta)\kappa\sum\limits_{t \in \mathcal{T}_1}{}\pbrak{\sum\limits_{i=1}^{N}\matnorm{\sqrt{\sens{\mathbf{x}_t}{\bm\theta_{t+1}}}\mathbf{x}^i_t}{(\mathbf{W}^i_t)\inv}}^2
    \\
    &\overset{}{\leq} C  S^2 N d \kappa \log(T/\delta) \pbrak{Nd \log(T/4N) + \frac{N}{2\rho} \log T}.
\end{align*}

Here, the second inequality follows since $\mathbf{W}_t \mleq \mathbf{W}_{t+1}$, both $\bm\theta_{t+1} , \bm\theta^\star \in \C_t(\delta)$, and applying Lemma \ref{lemma: conversion of norms}, the third inequality follows from the definition of $\kappa$, and the last inequality follows from the bound obtained for $R_1(T)_1$ as well as the definition of $\eta_t(\delta)$. 

Now, summing over all indices in $\mathcal{T}_2$, we get:

\begin{align*}
    \sum\limits_{t \in \mathcal{T}_2}{} \sens{\mathbf{x}_t}{\bm\theta^\star}\inner{\pbrak{{\mathbf{x}_t^\star} - \mathbf{x}_t}}{\bm\theta^\star} &\overset{}{\leq}\sum\limits_{t \in \mathcal{T}_2}{} \sqrt{\sens{\mathbf{x}_t}{\bm\theta^\star}}\sqrt{\sens{\mathbf{x}_t}{\bm\theta_{t+1}}}\inner{\pbrak{{\mathbf{x}_t^\star} - \mathbf{x}_t}}{\bm\theta^\star}
    \\
    &\overset{}{\leq} CS N d^{1/2}\sqrt{ \left(d\log(T/4N) + \frac{1}{2\rho}\log T \right)\log(T/\delta)} \pbrak{\sqrt{R(T)} + \sqrt{\sum\limits_{t\notin\mathcal{T}}{}\sens{{\mathbf{x}_t^\star}}{\bm\theta^\star}}}.
\end{align*}

Here, the first inequality follows from the definition of $\mathcal{T}_2$, and the second inequality follows similarly to the steps used to bound $R_1(T)_1$.

Combining all the bounds on $R_1(T)$, we get,
\begin{align*}
    R_1(T) &\leq CS N d^{1/2}\sqrt{ \left(d\log(T/4N) + \frac{1}{2\rho}\log T \right)\log(T/\delta)} \pbrak{\sqrt{R(T)} + \sqrt{\sum\limits_{t\notin\mathcal{T}}{}\sens{{\mathbf{x}_t^\star}}{\bm\theta^\star}}}\\
    &+  C S^2 N^2 d \kappa \log(T/\delta) \pbrak{ d \log(T/4N) + \frac{1}{\rho}\log(T)}.
\end{align*}

\underline{\textbf{Bounding $R_2(T)$: }}

\begin{align*}
    R_2(T) &= \sum\limits_{t \in [t^\prime , T] \setminus \mathcal{T}}{}\int_{0}^{1} (1-v)\dsigmoid{v\inner{\mathbf{x}_t}{\bm\theta^\star} + (1-v)\inner{{\mathbf{x}_t^\star}}{\bm\theta^\star}}\diff v \; \pbrak{\inner{\pbrak{{\mathbf{x}_t^\star} - \mathbf{x}_t}}{\bm\theta^\star}}^2
    \\
    &\overset{}{\leq} 2 \eta_t(\delta)\sum\limits_{t \in [t^\prime , T] \setminus \mathcal{T}}{} \pbrak{\sum\limits_{i=1}^{N}\matnorm{\mathbf{x}^i_t}{(\mathbf{W}^i_t)\inv}}^2
    \\
    &\overset{}{\leq}  C  S^2 N^2 d \kappa \log(T/\delta) \pbrak{d \log(T/4N) + \frac{1}{\rho}\log T}
\end{align*}

Here, the first inequality follows by bounding $\lvert \dot\mu \rvert \leq 1$, and a simple application of the Cauchy-Schwarz inequality, while the second inequality follows from the steps used to bound $R_1(T)_2$ and the definition of $\eta_t(\delta)$.

Combining all the bounds for $R(T)$, we get
\begin{align*}
    R(T) &\leq CS N d^{1/2}\sqrt{ \left(d\log(T/4N) + \frac{1}{2\rho}\log T \right)\log(T/\delta)} \pbrak{\sqrt{R(T)} + \sqrt{\sum\limits_{t\notin\mathcal{T}}{}\sens{{\mathbf{x}_t^\star}}{\bm\theta^\star}}}\\
    &+   C  S^2 N^2 d \kappa \log(T/\delta) \pbrak{d \log(T/4N)
    + \frac{1}{\rho}\log T}.
\end{align*}
Applying Lemma \ref{lemma: quadratic inequality}, we get that
\begin{align*}
    R(T) &\leq CS N d^{1/2}\sqrt{ \left(d\log(T/4N) + \frac{1}{2\rho}\log T \right)\log(T/\delta)} \sqrt{\sum\limits_{t\notin\mathcal{T}}{}\sens{{\mathbf{x}_t^\star}}{\bm\theta^\star}} \\
    &+ C \kappa (1 + \kappa) S^2N^2d\log(T/\delta)\pbrak{ d \log(T/4N) + \frac{1}{\rho}\log(T)}.
\end{align*}

Substituting this bound back in the expression for $Regret(T)$, and using the bounds on $T_0$ and $|\mathcal{T}|$ (Lemma \ref{lemma: warmup}), gives us the desired expression.

\end{proof}

\newpage
\subsection{Supporting Lemmas for Theorem \ref{appendix: proof_regret_oful}}

\begin{lemma}
Let $\mathbf{W}_t, \mathbf{W}^i_t , \mathbf{W}^{i,j}_t$, and $\mathbf{U}_t$ be defined as in Section \ref{appendix:notation} for all slots $i , j \in [N]$. Then, we have that
$$\mathbf{U}_t^{-\frac{1}{2}}\mathbf{W}_t\mathbf{U}_t^{-\frac{1}{2}} = \mathbf{I}_{Nd} + \mathbf{A}_t$$
where 
\[\mathbf{A}_t = 
    \begin{bmatrix}
    \mathbf{0}_{d} & 
    ({\mathbf{W}^{1}_t})^{-\frac{1}{2}} \mathbf{W}^{1,2}_t ({\mathbf{W}^{2}_t})^{-\frac{1}{2}} & 
    \ldots & 
   ({\mathbf{W}^{1}_t})^{-\frac{1}{2}} \mathbf{W}^{1,N}_t ({\mathbf{W}^{N}_t})^{-\frac{1}{2}}\\
    \\
   ({\mathbf{W}^{2}_t})^{-\frac{1}{2}} \mathbf{W}^{2,1}_t ({\mathbf{W}^{1}_t})^{-\frac{1}{2}} &
   \mathbf{0}_{d} &
   \ldots &
  {(\mathbf{W}^{2}_t})^{-\frac{1}{2}} \mathbf{W}^{2,N}_t ({\mathbf{W}^{N}_t})^{-\frac{1}{2}}\\
    \\
    \vdots & \vdots & \ddots & \vdots\\
    \\
     ({\mathbf{W}^{N}_t})^{-\frac{1}{2}} \mathbf{W}^{N,1}_t ({\mathbf{W}^{1}_t})^{-\frac{1}{2}} &
     ({\mathbf{W}^{N}_t})^{-\frac{1}{2}} \mathbf{W}^{N,2}_t ({\mathbf{W}^{2}_t})^{-\frac{1}{2}} &
     \ldots& 
    \mathbf{0}_{d}
    \end{bmatrix}.
\]
    \label{lemma: design_matrix_decomp}
\end{lemma}

\begin{proof}  To prove the claim, it is enough to show $\mathbf{W}_t = \mathbf{U}_t + \mathbf{U}_t^{\frac{1}{2}}\mathbf{A}_t\mathbf{U}_t^{\frac{1}{2}}$. We can decompose $\mathbf{W}_t$ as follows:
\begin{align*}
    \mathbf{W}_t &= \mathbf{I}_{Nd} + \sum_{s \in [t-1] \setminus \mathcal{T}} \sens{\mathbf{x}_s}{\bm\theta_{s+1}} \mathbf{x}_s \mathbf{x}^\top_s
    \\
    &\overset{}{=}\mathbf{I}_{Nd} +   \sum_{s \in [t-1] \setminus \mathcal{T}} \sens{\mathbf{x}_s}{\bm\theta_{s+1}} \pbrak{\sum\limits_{i=1}^{N} \tilde{\mathbf{x}}^i_s} \pbrak{\sum\limits_{i=1}^{N} \tilde{\mathbf{x}}^{i ^\top}_s}
    \\
    & \overset{}{=} \mathbf{I}_{Nd} +   \sum_{s \in [t-1] \setminus \mathcal{T}} \sens{\mathbf{x}_s}{\bm\theta_{s+1}} \begin{bmatrix} 
    \mathbf{x}^1_s \mathbf{x}^{1^\top}_s &  \mathbf{x}^1_s \mathbf{x}^{2^\top}_s & \ldots & \mathbf{x}^1_s \mathbf{x}^{N^\top}_s \\
    \\
     \mathbf{x}^2_s \mathbf{x}^{1^\top}_s &  \mathbf{x}^2_s \mathbf{x}^{2^\top}_s & \ldots & \mathbf{x}^2_s \mathbf{x}^{N^\top}_s \\
     \\
     \vdots & \vdots & \ddots & \vdots\\
     \\
     \mathbf{x}^N_s \mathbf{x}^{1^\top}_s &  \mathbf{x}^N_s \mathbf{x}^{2^\top}_s & \ldots & \mathbf{x}^N_s \mathbf{x}^{N^\top}_s 
    \end{bmatrix}
        \\
        &= \mathbf{U}_t + \begin{bmatrix}
         \mathbf{0}_d & \mathbf{W}^{1,2}_t & \ldots & \mathbf{W}^{1,N}_t\\
         \\
         \mathbf{W}^{2,1}_t & \mathbf{0}_d & \ldots & \mathbf{W}^{2,N}_t\\
          \\
          \vdots & \vdots & \ddots & \vdots\\
          \\
           \mathbf{W}^{N,1}_t & \mathbf{W}^{N,2}_t & \ldots & \mathbf{0}_d
     \end{bmatrix}.
\end{align*}

Here, the first equality is a simple consequence of the fact that $\mathbf{x}_s = \sum\limits_{i=1}^N \tilde{\mathbf{x}}^i_s$, while the last equality follows from the definitions of $\mathbf{W}^i_t$, $\mathbf{W}^{i,j}_t$, and $\mathbf{U}_t$ (Section \ref{appendix:notation}).

Denote $\mathbf{B}_t = 
        \begin{bmatrix}
         \mathbf{0}_d & \mathbf{W}^{1,2}_t & \ldots & \mathbf{W}^{1,N}_t\\
         \\
         \mathbf{W}^{2,1}_t & \mathbf{0}_d & \ldots & \mathbf{W}^{2,N}_t\\
          \\
          \vdots & \vdots & \ddots & \vdots\\
          \\
           \mathbf{W}^{N,1}_t & \mathbf{W}^{N,2}_t & \ldots & \mathbf{0}_d\\
     \end{bmatrix}$.    
We finish the claim by showing that  $\mathbf{A}_t = \mathbf{U}_t^{-\frac{1}{2}}\mathbf{B}_t\mathbf{U}_t^{-\frac{1}{2}}$.

Note that since $\mathbf{U}_t$ is a diagonal block matrix, $\mathbf{U}_t^{-\frac{1}{2}} = \textrm{diag}\pbrak{(\mathbf{W}^1_t)^{-\frac{1}{2}} , \ldots , (\mathbf{W}^N_t)^{-\frac{1}{2}}}$. We can write the $(i,j)^{\text{th}}$ element (in this case, $d \times d$ block) of $\mathbf{U}_t^{-\frac{1}{2}}\mathbf{B}_t\mathbf{U}_t^{-\frac{1}{2}}$ as:
\begin{align*}
    \sbrak{\mathbf{U}_t^{-\frac{1}{2}} \mathbf{B}_t \mathbf{U}_t^{-\frac{1}{2}}}_{i,j} &= \sum\limits_{k=1}^{N}\sum\limits_{l=1}^{N} \sbrak{\mathbf{U}_t^{-\frac{1}{2}}}_{i,k} \sbrak{\mathbf{B}_t}_{k,l} \sbrak{\mathbf{U}_t^{-\frac{1}{2}}}_{l,j}\\
    &= \mathbbm{1}\{i=k\} \mathbbm{1}\{k \neq l\} \mathbbm{1}\{j=l\} \sbrak{\mathbf{U}_t^{-\frac{1}{2}}}_{i,k} \sbrak{\mathbf{B}_t}_{k,l} \sbrak{\mathbf{U}_t^{-\frac{1}{2}}}_{l,j}\\
    &=\begin{cases}
    (\mathbf{W}^j_t)^{-\frac{1}{2}} \mathbf{W}^{i,j}_t (\mathbf{W}^j_t)^{-\frac{1}{2}}& i\neq j\\
    \mathbf{0}_{d\times d} & i = j
    \end{cases}\\
    &= \sbrak{\mathbf{A}_t}_{i,j}.
\end{align*}
The second equality follows from the fact that the off-diagonal entries in $\mathbf{U}_t^{-\frac{1}{2}}$ are zero matrices and likewise, the diagonal entries in $\mathbf{B}_t$ are zero matrices. This completes the proof.

\end{proof}

\begin{proposition}
    Let $\Lambda\pbrak{\mathbf{A}}$ denote the set of eigenvalues of $\mathbf{A}$. Then,
    \begin{align*}
        \Lambda\pbrak{\mathbf{A}} = \Lambda\pbrak{\begin{bmatrix}
            \mathbf{0} & \mathbf{0}\\
            \mathbf{0} & \mathbf{A}
        \end{bmatrix}}.
    \end{align*}
    \label{prop: same_eig}
\end{proposition}

\begin{proposition}
    Let $\mathbf{A}$ and $\mathbf{B}$ be two symmetric p.s.d matrices. Then, we have that
    \begin{align*}
        \eigmax{\mathbf{A} + \mathbf{B}} \leq\eigmax{\mathbf{A}} + \eigmax{\mathbf{B}},
    \end{align*}
    \[\eigmin{\mathbf{A} + \mathbf{B}} \geq \eigmin{\mathbf{A}} + \eigmin{\mathbf{B}}.\]
    \label{prop: max_min_eig}
\end{proposition}

\begin{lemma}
    Define the matrix recurrence relation as follows:
    \begin{align*}
        \mathbf{A}^{(k)} = \begin{bmatrix}
           \mathbf{0} & \mathbf{Z}_k\\
            {\mathbf{Z}_k}^\top & \mathbf{A}^{(k-1)}
        \end{bmatrix} \text{ and } \mathbf{A}^{(1)} = \begin{bmatrix}
            \mathbf{0} & \mathbf{Z}_1\\
            {\mathbf{Z}_1}^\top & \mathbf{0}
        \end{bmatrix}
    \end{align*}
    where $\{\mathbf{Z}_i\}_i$ are matrices of appropriate dimensions.
    Then, we have that
    \[
    \eigmax{\mathbf{A}^{(k)}} \leq \sum\limits_{i=1}^{k}\singmax{\mathbf{Z}_i} \quad \text{ and } \quad \eigmin{\mathbf{A}^{(k)}} \geq -\sum\limits_{i=1}^{k}\singmax{\mathbf{Z}_i}.
    \]
    \label{lemma: recurrence}
\end{lemma}

\begin{proof}  The proof follows by induction. For $k=1$, we see that the statement indeed holds from Lemma \ref{lemma: max_min_eig_hermitian}.

Assume that the statement holds for $k = n$, i.e,  $\eigmax{\mathbf{A}^{(n)}} \leq \sum\limits_{i=1}^{n}\singmax{\mathbf{Z}_i}$ and $\eigmin{\mathbf{A}^{(n)}} \geq -\sum\limits_{i=1}^{n}\singmin{\mathbf{Z}_i}$.

Consider $\mathbf{A}^{(n+1)} = 
    \begin{bmatrix}
        \mathbf{0} & \mathbf{Z}_{n+1}\\
        {\mathbf{Z}_{n+1}}^\top & \mathbf{A}^{(n)}
    \end{bmatrix} = 
    \begin{bmatrix}
        \mathbf{0} & \mathbf{Z}_{n+1}\\
        {\mathbf{Z}_{n+1}}^\top & \mathbf{0}
    \end{bmatrix} + 
    \begin{bmatrix}
        \mathbf{0} & \mathbf{0}\\
        \mathbf{0} & \mathbf{A}^{(n)}
    \end{bmatrix}$.

We have that,
\begin{align*}
    \eigmax{\mathbf{A}^{(n+1)}} &\overset{}{\leq} \eigmax{\begin{bmatrix}
        \mathbf{0} & \mathbf{Z}_{n+1}\\
        {\mathbf{Z}_{n+1}}^\top & \mathbf{0}
    \end{bmatrix}} + \eigmax{\begin{bmatrix}
        \mathbf{0} & \mathbf{0}\\
        \mathbf{0} & \mathbf{A}^{(n)}
    \end{bmatrix}}
    \\
    &\overset{}{=} \singmax{\mathbf{Z}_{n+1}} + \eigmax{\mathbf{A}^{(n)}} \\
    &\overset{}{\leq} \singmax{\mathbf{Z}_{n+1}} + \sum\limits_{i=1}^{n}\singmax{\mathbf{Z_i}} 
    \\
    &= \sum\limits_{i=1}^{n+1}\singmax{\mathbf{Z_i}}.
\end{align*}
where the first inequality follows from Proposition \ref{prop: max_min_eig}, the second equality follows from Lemma \ref{lemma: max_min_eig_hermitian} and Lemma \ref{prop: same_eig}, and the last inequality follows from the induction hypothesis.

Similarly, we have that
\begin{align*}
    \eigmin{\mathbf{A}^{(n+1)}} &\overset{}{\geq} \eigmin{\begin{bmatrix}
        \mathbf{0} & \mathbf{Z}_{n+1}\\
        {\mathbf{Z}_{n+1}}^\top & \mathbf{0}
    \end{bmatrix}} + \eigmin{\begin{bmatrix}
        \mathbf{0} & \mathbf{0}\\
        \mathbf{0} & \mathbf{A}^{(n)}
    \end{bmatrix}}
    \\
    &\overset{}{=} -\singmax{\mathbf{Z}_{n+1}} + \eigmin{\mathbf{A}^{(n)}} \\
    &\overset{}{\geq} -\singmax{\mathbf{Z}_{n+1}} - \sum\limits_{i=1}^{n}\singmax{\mathbf{Z_i}} 
    \\
    &= -\sum\limits_{i=1}^{n+1}\singmax{\mathbf{Z_i}}.
\end{align*}
where the first inequality follows from Proposition \ref{prop: max_min_eig}, the second equality follows from Lemma \ref{lemma: max_min_eig_hermitian} and Lemma \ref{prop: same_eig}, and the last inequality follows from the induction hypothesis.
\end{proof}

\begin{lemma}
    Let $\mathbf{x}^i_t$ and $\mathbf{x}^j_t$ be the items chosen in two different slots $i$ and $j$ at round $t$. Then, we have that 
    $$\E\sbrak{\mathbf{x}^i_t {\mathbf{x}^j_t}^\top | \filteration{t}} = \mathbf{0}_d.$$
    \label{lemma: independence of cross terms}
\end{lemma}
\begin{proof}  It is easy to see that the item chosen in slot $i$ during round $t$ only depends on $\cbrak{\mathbf{x}_s}_{s=1}^{t-1}, \cbrak{\bm\theta_{s+1}}_{s=1}^{t-1}$, and $\cbrak{\mathbf{x}^i_s}_{s=1}^{t}$, all of which are $\filteration{t}$-measurable. Thus, conditioning on $\mathcal{F}_t$, we have that the items being chosen in two different slots are independent of one another. Hence, we have that
$$\E\sbrak{\mathbf{x}^i_t \mathbf{x}^{j ^\top}_t | \filteration{t}} = \E\sbrak{\mathbf{x}^i_t | \filteration{t}}\E\sbrak{\mathbf{x}^{j}_t | \filteration{t}} = \mathbf{0}_d.$$
where the last equality follows from Assumption \ref{assumption: diversity}.
\end{proof}

\begin{lemma}
    For all rounds $t \in [T]$, the diversity assumptions in Assumption \ref{assumption: diversity} can be extended to the set of vectors $\cbrak{\sqrt{\dsigmoid{\inner{\mathbf{x}_t}{\bm\theta_{t+1}}}}\mathbf{x}^i_t}_{i=1}^N$, i.e, we can show the following:
    \begin{enumerate}
    \item $\E\sbrak{\sqrt{\dsigmoid{\inner{\mathbf{x}_t}{\bm\theta_{t+1}}}}\mathbf{x}^i_t | \filteration{t}} = \mathbf{0}_d$,
    \item $\E\sbrak{\dsigmoid{\inner{\mathbf{x}_t}{\bm\theta_{t+1}}}\mathbf{x}^i_t{\mathbf{x}^j_t}^\top | \filteration{t}} = \mathbf{0}_d$ where $i \neq j$,
    \item $\E\sbrak{\dsigmoid{\inner{\mathbf{x}_t}{\bm\theta_{t+1}}}\mathbf{x}^i_t{\mathbf{x}^i_t}^\top | \filteration{t}} \mgeq \rho\mathbf{I}_d$ .
    \end{enumerate}
    \label{lemma: extension of diversity}
\end{lemma}
\begin{proof} We first attempt to bound $\sens{\mathbf{x}_t}{\bm\theta_{t+1}}$.  

Using the Cauchy-Schwarz inequality, it is easy to see that $-S \leq \inner{\mathbf{x}_t}{\bm\theta_{t+1}} \leq S$. Since $\dsigmoid{.}$ is an increasing function on $(-\infty ,0 ]$ and a decreasing function on  $[0,\infty)$, we have that 
$$\sens{\mathbf{x}_t}{\bm\theta_{t+1}} \in 
\begin{cases}
\sbrak{\dsigmoid{S} , \frac{1}{4}} & \text{if } \inner{\mathbf{x}_t}{\bm\theta_{t+1}} \in [0,S],\\
\sbrak{\dsigmoid{-S} , \frac{1}{4}} & \text{if } \inner{\mathbf{x}_t}{\bm\theta_{t+1}} \in [-S,0].
\end{cases}$$
Since $\dsigmoid{-S} = \dsigmoid{S}$, we have that $\sens{\mathbf{x}_t}{\bm\theta_{t+1}} \in \sbrak{\dsigmoid{S} , \frac{1}{4}}$.

Now, using the diversity conditions (Assumption \ref{assumption: diversity}), we have that

\[\sqrt{\dsigmoid{S}}\E\sbrak{\mathbf{x}^i_t | \filteration{t}}\leq\E\sbrak{\sqrt{\dsigmoid{\inner{\mathbf{x}_t}{\bm\theta_{t+1}}}}\mathbf{x}^i_t | \filteration{t}} \leq \sqrt{\frac{1}{4}}\E\sbrak{\mathbf{x}^i_t | \filteration{t}}.
\]

Since, $\E\sbrak{\mathbf{x}^i_t | \filteration{t}} = \mathbf{0}_d$, we obtain the first claim.

To prove the second claim, we have that
\[
\dsigmoid{S}\E\sbrak{\mathbf{x}^i_t {\mathbf{x}^{j}_t}^\top| \filteration{t}} \leq \E\sbrak{\dsigmoid{\inner{\mathbf{x}_t}{\bm\theta_{t+1}}}\mathbf{x}^i_t{\mathbf{x}^{j}_t}^\top | \filteration{t}} \leq \frac{1}{4}\E\sbrak{\mathbf{x}^i_t {\mathbf{x}^{j}_t}^\top | \filteration{t}}.
\]
From Lemma \ref{lemma: independence of cross terms}, we have that $\E\sbrak{\mathbf{x}^i_t{\mathbf{x}^{j}_t}^\top | \filteration{t}} = \mathbf{0}_d$, which proves the second claim.

Finally, since $\kappa = \max\limits_{\mathbf{x}} \max\limits_{\bm\theta}\frac{1}{\dsigmoid{\inner{\mathbf{x}}{\bm\theta}}}$, we have that $\kappa \geq \frac{1}{\sens{\mathbf{x}_t}{\bm\theta_{t+1}}}$. This, combined with Assumption \ref{assumption: diversity}, gives us
\begin{align*}
    \E\sbrak{\dsigmoid{\inner{\mathbf{x}_t}{\bm\theta_{t+1}}}\mathbf{x}^i_t{\mathbf{x}^i_t}^\top | \filteration{t}} \mgeq \frac{1}{\kappa} \E\sbrak{\mathbf{x}^i_t {\mathbf{x}^{i}_t}^\top | \filteration{t}} \mgeq \rho\mathbf{I}_d.
\end{align*}
This finishes the proof. 
\end{proof}

\begin{lemma}
    Let $\mathbf{W}^{i,j}_t$ be defined as in Section \ref{appendix:notation}. For all $i \in [N]$, $j \in [i+
    1,N]$, and $t$ such that $|[t-1]\setminus \mathcal{T}| \geq 0$, with probability at least $1-\delta$, we have that
    \[\norm{\mathbf{W}^{i,j}_t} \leq \sqrt{\frac{|[t-1]\setminus\mathcal{T}|}{2N^2}\log\pbrak{\frac{d N(N-1)}{\delta}}}.\]
    \label{lemma: norm_cross_terms}
\end{lemma}
\begin{proof}  From Lemma \ref{lemma: extension of diversity}, we know that \[\E\sbrak{\dsigmoid{\inner{\mathbf{x}_t}{\bm\theta_{t+1}}}\mathbf{x}^i_t{\mathbf{x}^j_t}^\top | \filteration{t}} = \mathbf{0}_d.\] Fix $i \in [N]$ and $j \in [i+1,N]$. Invoking Lemma \ref{lemma: generalized norm of cross terms} with $\mathbf{x}_s = \sqrt{\dsigmoid{\inner{\mathbf{x}_t}{\bm\theta_{t+1}}}}\mathbf{x}^i_t$, $\mathbf{z}_s = \sqrt{\dsigmoid{\inner{\mathbf{x}_t}{\bm\theta_{t+1}}}}\mathbf{x}^j_t$, $m_1 = m_2 = \sqrt{\frac{\sens{\mathbf{x}_t}{\bm\theta_{t+1}}}{N}} \leq \frac{1}{2\sqrt{N}}$, $d_1=d_2=d$, and $\delta = \frac{2\delta}{N(N-1)}$ results in:
$$\P\cbrak{\exists t \text{ such that } |[t-1] \setminus \mathcal{T}| \geq 1: \norm{\mathbf{W}^{i,j}_t} \geq \sqrt{\frac{|[t-1] \setminus \mathcal{T}|}{2N^2}\log\pbrak{\frac{2d N(N-1)}{2\delta}}}} \leq \frac{2\delta}{N(N-1)}.$$
A union bound over all $i \in [N]$ and $j \in [i+1,N]$ finishes the proof.
\end{proof}

\begin{lemma}
    Let $\mathbf{W}^i_t$ be defined as in Section \ref{appendix:notation}. Then, for all $t$ such that $|[t-1] \setminus \mathcal{T}| \geq T_0$ and  $i \in \sbrak{N}$, with probability $1 - \delta$, we have that
   \[\eigmin{ \mathbf{W}^i_t} \geq 1 + 0.5\rho |[t-1] \setminus \mathcal{T}|.\]
    \label{lemma : norm of slot design matrix}
\end{lemma}
\begin{proof} From Lemma \ref{lemma: extension of diversity}, we know that $\E\sbrak{\sqrt{\dsigmoid{\inner{\mathbf{x}_t}{\bm\theta_{t+1}}}}\mathbf{x}^i_t | \filteration{t}} = \mathbf{0}_d$ and $\E\sbrak{\dsigmoid{\inner{\mathbf{x}_t}{\bm\theta_{t+1}}}\mathbf{x}^i_t{\mathbf{x}^i_t}^\top | \filteration{t}} \mgeq \rho\mathbf{I}_d$. Fix some $i \in [N]$. Thus, invoking Lemma \ref{lemma: min_eig_design} with $\mathbf{x}_t = \sqrt{\dsigmoid{\inner{\mathbf{x}_t}{\bm\theta_{t+1}}}}\mathbf{x}^i_t$, $m = \frac{1}{2\sqrt{N}}$, $d = d$, $\gamma = 1$, $c = \frac{1}{2}$, and $\delta = \frac{\delta}{N}$, we get that with probability atleast $1 - \frac{\delta}{N}$, 
$$\eigmin{\mathbf{W}^i_t} \geq 1 + 0.5\rho |[t-1] \setminus \mathcal{T}|,  \forall  t  \text{ such that } |[t-1] \setminus \mathcal{T}| \geq \frac{3 + 2N\rho}{3\rho^2 N^2}\log\pbrak{\frac{2dNT}{\delta}}.$$
Define $T_0 = \frac{3 + 2\rho N}{3\rho^2}(N-1)^2\log\pbrak{\frac{2d NT}{\delta}}$. Using the fact that $(N-1)^2 \geq 1/N^2$ gives us:
$$\P\cbrak{\forall t \text{ such that } |[t-1] \setminus \mathcal{T}| \geq T_0: \eigmin{\mathbf{W}^i_t} \geq 1 + 0.5\rho |[t-1] \setminus \mathcal{T}|} \geq 1 - \frac{\delta}{N}.$$
A union bound over all $i \in [N]$ finishes the proof.
\end{proof}






\begin{lemma}
    For $i \in [N-1]$, define the matrix $\mathbf{Z}^{(i)}_t$ as follows:
    \[
        \mathbf{Z}^{(i)}_t
        = \begin{bmatrix}
            f(N-i , N-i+1) , f(N-i , N-i+2) , \ldots , f(N-i , N)
        \end{bmatrix}, 
        \quad f(a,b) = (\mathbf{W}^{a}_t)^{-\frac{1}{2}} \mathbf{W}^{a,b}_t (\mathbf{W}^{b}_t)^{-\frac{1}{2}}.
    \]
    Then, assuming the conditions in Lemma \ref{lemma : norm of slot design matrix} hold, and assuming $\rho \geq \frac{12}{N}$, with probability $1 - 2\delta$, we have that 
    $$\norm{\mathbf{Z}^{(i)}_t} \leq \frac{i}{2N(N-1)}.$$
    \label{lemma: bound on norm of Z}
\end{lemma}
\begin{proof} We know that $\norm{\mathbf{Z}} = \sup\limits_{\twonorm{\mathbf{b}} \leq 1}\twonorm{\mathbf{Z}\mathbf{b}}$. Thus,
\begin{align*}
    \norm{\mathbf{Z}^{(i)}_t} &= \sup\limits_{\twonorm{\mathbf{b}} \leq 1} \twonorm{\mathbf{Z}^{(i)}_t \mathbf{b}} 
    \\
    &=\sup\limits_{\sum\limits_{j=1}^{i}\twonorm{\mathbf{b}_j} \leq 1} \twonorm{\sum\limits_{j=1}^{i} f(N-i , N-i+j) \mathbf{b}_j} 
    \\ 
    &\overset{}{\leq} \sup\limits_{\sum\limits_{j=1}^{i} \twonorm{\mathbf{b}_j} \leq 1} \sum\limits_{j=1}^{i} \twonorm{f(N-i , N-i+j) \mathbf{b}_j}
    \\
    &\leq \sum\limits_{j=1}^{i} \sup\limits_{\twonorm{\mathbf{b}_j} \leq 1} \twonorm{f(N-i , N-i+j) \mathbf{b}_j}
    \\
    &\overset{}{\leq} \sum\limits_{j=1}^{i} \norm{(\mathbf{W}^{N-i}_t)^{-\frac{1}{2}}} \norm{\mathbf{W}^{N-i , N-i+j}_t} \norm{(\mathbf{W}^{N-i+j}_t)^{-\frac{1}{2}}}
    \\
    &\overset{}{\leq} \sum\limits_{j=1}^{i} \frac{\norm{\mathbf{W}^{N-i , N-i+j}_t}}{\sqrt{\eigmin{\mathbf{W}^{N-i}_t} \eigmin{\mathbf{W}^{N-i+j}_t}}}
    \end{align*}
    where the second-to-last inequality follows from the submultiplicativity of the norm, while the last inequality follows from Rayleigh's quotient. Using Lemma \ref{lemma: norm_cross_terms} and Lemma \ref{lemma : norm of slot design matrix}, we have
    
    \begin{align*}
    \norm{\mathbf{Z}^{(i)}_t} &\overset{}{\leq} \sum\limits_{j=1}^{i} \frac{\sqrt{\frac{|[t-1] \setminus \mathcal{T}|}{2N^2}\log\pbrak{\frac{d N(N-1)}{\delta}}}}{1 + 0.5\rho |[t-1] \setminus \mathcal{T}|}
    \\
    &\overset{}{\leq} \sum\limits_{j=1}^{i} \sqrt{\frac{\frac{1}{2N^2}\log\pbrak{\frac{d N(N-1)}{\delta}}}{\frac{3+2\rho N}{12}(N-1)^2\log\pbrak{\frac{2d NT}{\delta}}}} 
    \\
    &
    \overset{}{\leq} \frac{i}{2N(N-1)}.
\end{align*}

where the second-to-last inequality follows from the definition of $T_0$ in Lemma \ref{lemma : norm of slot design matrix}, while the last inequality uses the fact that $\rho N \geq 12$.

Finally, using Lemma \ref{lemma: norm_cross_terms} and Lemma \ref{lemma : norm of slot design matrix}, we have that the above event occurs with probability at least $1 - 2\delta$.
\end{proof}

\begin{lemma}
    Let $\mathbf{U}_t$ and $\mathbf{W}_t$ be defined as in Section \ref{appendix:notation}. Then, assuming the conditions in Lemma \ref{lemma : norm of slot design matrix} hold, with probability at least $1-2\delta$, we have, 
    $$ \frac{3}{4}\mathbf{U}_t \mleq \mathbf{W}_t \mleq \frac{5}{4}\mathbf{U}_t.$$
    \label{lemma: ineq on W}
\end{lemma}
\begin{proof}  For $i \in [N-1]$, define the matrix recurrence relation:
\[
     \quad \mathbf{A}^{(1)}_t = \begin{bmatrix}
        \mathbf{0}_{d \times d} & \mathbf{Z}^{(1)}_t\\
        {\mathbf{Z}^{(1)}_t}^\top & \mathbf{0}_{d \times d}
    \end{bmatrix},
    \quad \mathbf{A}^{(i)}_t = \begin{bmatrix}
        \mathbf{0}_{d \times d} & \mathbf{Z}^{(i)}_t\\
        {\mathbf{Z}^{(i)}_t}^\top & \mathbf{A}^{(i-1)}_t
    \end{bmatrix}
\]
where 
\[
        \mathbf{Z}^{(i)}_t
        = \begin{bmatrix}
            f(N-i , N-i+1) , f(N-i , N-i+2) , \ldots , f(N-i , N)
        \end{bmatrix}, 
        \quad f(a,b) = (\mathbf{W}^{a}_t)^{-\frac{1}{2}} \mathbf{W}^{a,b}_t (\mathbf{W}^{b}_t)^{-\frac{1}{2}}.
\]
Then, it is easy to see that $\mathbf{A}_t$ defined in Lemma \ref{lemma: design_matrix_decomp} is the same as $\mathbf{A}^{(N-1)}_t$.  From Lemma \ref{lemma: recurrence}, we have that $$\eigmax{\mathbf{A}_t} \leq \sum\limits_{i=1}^{N-1}\singmax{\mathbf{Z}^{(i)}_t} = \sum\limits_{i=1}^{N-1} \norm{\mathbf{Z}^{(i)}_t} \leq \sum\limits_{i=1}^{N-1} \frac{i}{2N(N-1)}= \frac{1}{4}.$$
    Similarly,
    $$\eigmin{\mathbf{A}_t} \geq -\sum\limits_{i=1}^{N-1}\singmax{\mathbf{Z}^{(i)}_t} = -\sum\limits_{i=1}^{N-1} \norm{\mathbf{Z}^{(i)}_t} \geq -\sum\limits_{i=1}^{N-1} \frac{i}{2N(N-1)} = -\frac{1}{4}.$$
    
Thus, we can write 

\begin{align*}
    -\frac{1}{4} \mathbf{I}_{d} \mleq \mathbf{A}_t \mleq \frac{1}{4}\mathbf{I}_{d}
    \implies -\frac{1}{4} \mathbf{I}_{d} \mleq  \mathbf{U}_t^{-\frac{1}{2}}\mathbf{W}_t\mathbf{U}_t^{-\frac{1}{2}} - \mathbf{I}_{d}\mleq \frac{1}{4}\mathbf{I}_{d}
    \implies \frac{3}{4}\mathbf{U}_t \mleq  \mathbf{W}_t \mleq \frac{5}{4}\mathbf{U}_t.
\end{align*}
\end{proof}

\begin{lemma}
    Let $\tilde{\mathbf{x}}^i$ be the lift of $\mathbf{x}^i$, and $\mathbf{W}^i_t$ and $\mathbf{W}_t$ be defined as in Section \ref{appendix:notation}. Then, with probability at least $1 - 2\delta$, assuming the conditions in Lemma \ref{lemma : norm of slot design matrix} hold, we have 
    $$\matnorm{\tilde{\mathbf{x}}^i}{\mathbf{W}_t\inv} \leq \frac{4}{3}\matnorm{\mathbf{x}^i}{(\mathbf{W}_t^{i})\inv}$$
    \label{lemma: conversion of norms}
\end{lemma}
\begin{proof} 
From Lemma \ref{lemma: ineq on W}, we have
\begin{align*}
    \matnorm{\tilde{\mathbf{x}}^i}{\mathbf{W}\inv} \leq \frac{4}{3} \matnorm{\tilde{\mathbf{x}}^i}{\mathbf{U}\inv} = \frac{4}{3} \matnorm{\mathbf{x}^i}{(\mathbf{W}^i)\inv}.
\end{align*}
where the last inequality follows from the definition of the lift of $\mathbf{x}$ and the block-diagonal structure of $\mathbf{U}$.

\end{proof}

\begin{lemma}
    Let $\mathbf{U}^\mathcal{H}_t$ and $\mathbf{W}^\mathcal{H}_t$ be defined as in Section \ref{appendix:notation}. With probability at least $1 - 2\delta$, for all $t$ such that $|\mathcal{H}_t| \geq T_0$ and $\rho \geq \frac{12}{N}$, we have
    $$\frac{3}{4}\mathbf{U}^\H_t \mleq  \mathbf{V}^\H_t \mleq \frac{5}{4}\mathbf{U}^\H_t.$$
    \label{lemma: bounds on V for failing data dependent condition}
\end{lemma}
\begin{proof}  

First, notice the similarity in structures between $\mathbf{V}^{\mathcal{H}}_t$ and $\mathbf{W}_t$, as well as between $\mathbf{U}^\mathcal{H}_t$ and $\mathbf{U}_t$. Thus, a decomposition similar to the one in Lemma \ref{lemma: design_matrix_decomp} holds, i.e, we can show that
\[
    (\mathbf{U}^{\mathcal{H}}_t)^{-\frac{1}{2}} \mathbf{V}^{\mathcal{H}}_t (\mathbf{U}^{\mathcal{H}}_t)^{-\frac{1}{2}}  = \mathbf{I}_{Nd} + \mathbf{A},
\]
where 
\[
    \mathbf{A} = \begin{bmatrix}
        \mathbf{0}_d & (\mathbf{V}^{\mathcal{H},1}_t)^{-\frac{1}{2}} \mathbf{V}^{\mathcal{H},1,2}_t (\mathbf{V}^{\mathcal{H},2}_t)^{-\frac{1}{2}} & \ldots & (\mathbf{V}^{\mathcal{H},1}_t)^{-\frac{1}{2}} \mathbf{V}^{\mathcal{H},1,N}_t (\mathbf{V}^{\mathcal{H},N}_t)^{-\frac{1}{2}}
        \\
        \\
        (\mathbf{V}^{\mathcal{H},2}_t)^{-\frac{1}{2}} \mathbf{V}^{\mathcal{H},2,1}_t (\mathbf{V}^{\mathcal{H},1}_t)^{-\frac{1}{2}} & \mathbf{0}_d & \ldots & (\mathbf{V}^{\mathcal{H},2}_t)^{-\frac{1}{2}} \mathbf{V}^{\mathcal{H},2,N}_t (\mathbf{V}^{\mathcal{H},N}_t)^{-\frac{1}{2}}
        \\
        \\
        \vdots & \vdots & \ddots & \vdots
        \\
        \\
        (\mathbf{V}^{\mathcal{H},N}_t)^{-\frac{1}{2}} \mathbf{V}^{\mathcal{H},N,1}_t (\mathbf{V}^{\mathcal{H},1}_t)^{-\frac{1}{2}} & (\mathbf{V}^{\mathcal{H},N}_t)^{-\frac{1}{2}} \mathbf{V}^{\mathcal{H},N,2}_t (\mathbf{V}^{\mathcal{H},2}_t)^{-\frac{1}{2}} & \ldots & \mathbf{0}_d
    \end{bmatrix}.
\]
Now, to prove this claim, it is sufficient to obtain a bound on the norm of the matrices $\mathbf{V}^{\mathcal{H},i,j}_t$ and $\mathbf{V}^{\mathcal{H},i}_t$ (defined in Section \ref{appendix:notation}).

We first show that the diversity assumptions also hold for the set of vectors $\cbrak{\frac{1}{\sqrt\kappa}\mathbf{x}^i_t}_{i=1}^N$. For this, we show that $\frac{1}{\sqrt{\kappa}}$ is bounded.

From the proof of Lemma \ref{lemma: extension of diversity}, we have shown that $\sens{\mathbf{x}}{\bm\theta} \in \sbrak{\dsigmoid{S} , \frac{1}{4}}$. Since, $\kappa = \max\limits_{\mathbf{x}}\max\limits_{\bm\theta}\frac{1}{\sens{\mathbf{x}}{\bm\theta}}, \kappa \in \sbrak{4 , \frac{1}{\dsigmoid{S}}}$. Hence, $\frac{1}{\kappa} \in \sbrak{\dsigmoid{S} , \frac{1}{4}}$.

Now, we have that
\[
\sqrt{\dsigmoid{S}}\E\sbrak{\mathbf{x}^i_s | \filteration{s}} \leq \E\sbrak{\frac{1}{\sqrt\kappa}\mathbf{x}^i_t|\filteration{s}} \leq \frac{1}{2}\E\sbrak{\mathbf{x}^i_s | \filteration{s}}.
\]

From Assumption \ref{assumption: diversity}, we have that $\E\sbrak{\mathbf{x}^i_s | \filteration{s}} = \mathbf{0}_d$, which gives us 
\[\E\sbrak{\frac{1}{\sqrt\kappa}\mathbf{x}^i_t|\filteration{s}} = \mathbf{0}_d.\]
Similarly, we have that
\[
\dsigmoid{S}\E\sbrak{\mathbf{x}^i_s{\mathbf{x}^{j}_s}^\top | \filteration{s}} \leq \E\sbrak{\frac{1}{\kappa} \mathbf{x}^i_s{\mathbf{x}^{j}_s}^\top|\filteration{s}}  \leq \frac{1}{4}\E\sbrak{\mathbf{x}^i_s{\mathbf{x}^{j}_s}^\top | \filteration{s}}.
\]
From Lemma \ref{lemma: independence of cross terms}, we have that $\E\sbrak{\mathbf{x}^i_s{\mathbf{x}^{j}_s}^\top | \filteration{s}} = \mathbf{0}_d$, which gives us
\[
\E\sbrak{\frac{1}{\kappa} \mathbf{x}^i_s{\mathbf{x}^{j}_s}^\top|\filteration{s}} = \mathbf{0}_d.
\]

Finally, from Assumption \ref{assumption: diversity}, we have that

\begin{align*}
    \E\sbrak{\frac{1}{\kappa}\mathbf{x}^i_s{\mathbf{x}^{i}_s}^\top | \filteration{s}} \succeq \frac{1}{\kappa} \rho \kappa \mathbf{I}_d = \rho \mathbf{I}_d.
\end{align*}

Thus, we have shown that the diversity conditions hold for the set of vectors $\cbrak{\frac{1}{\sqrt\kappa}\mathbf{x}^i_t}_{i=1}^N$.
 Using an idea similar to Lemma \ref{lemma: norm_cross_terms}, we can show that
\[
    \P\cbrak{\forall i \in [N] , \forall j \in [i+1,N], \forall t \text{ such that } |\mathcal{H}_t| \geq 1: \norm{\mathbf{V}^{\mathcal{H},i,j}_t} \leq \sqrt{\frac{8|\mathcal{H}_t|}{\kappa^2N^2}\log\pbrak{\frac{d N(N-1)}{\delta}}}} \geq 1 - \delta.
\]

Similarly, using an idea similar to Lemma \ref{lemma : norm of slot design matrix}, we can show that
\[
    \P\cbrak{\forall i \in [N], \forall t \text{ such that } |\mathcal{H}_t| \geq \frac{48 + 8\kappa N\rho}{3\rho^2\kappa^2}(N-1)^2\log\pbrak{\frac{2d NT}{\delta}}: \eigmin{\mathbf{V}^{\mathcal{H},i}_t} \geq \gamma_t(\delta) + 0.5\rho |\mathcal{H}_t|} \geq 1 - \delta.
\]

Since $\kappa \geq 4$, note that $\frac{48 + 8\kappa N\rho}{3\rho^2\kappa^2}(N-1)^2\log\pbrak{\frac{2d NT}{\delta}} \leq T_0$, where the exact expression for $T_0$ appears in Lemma \ref{lemma : norm of slot design matrix}.



Now, for $i \in [N-1]$, define
\[
        \mathbf{Z}^{(i)}_t
        = \begin{bmatrix}
            f(N-i , N-i+1) , f(N-i , N-i+2) , \ldots , f(N-i , N)
        \end{bmatrix}, 
        \quad f(a,b) = (\mathbf{V}^{\mathcal{H},a}_t)^{-\frac{1}{2}} \mathbf{V}^{\mathcal{H},a,b}_t (\mathbf{V}^{\mathcal{H},b}_t)^{-\frac{1}{2}}
\]
Using similar ideas to Lemma \ref{lemma: bound on norm of Z}, we have
    \[
        \norm{\mathbf{Z}^{(i)}_t} \leq \frac{i}{2N(N-1)}.
    \]
    Then, using ideas similar to the ones in Lemma \ref{lemma: ineq on W}, we obtain, with probability at least $1 - 2\delta$, and for all $t$ such that $|\mathcal{H}_t| \geq T_0$,  
$$\frac{3}{4}\mathbf{U}^\H_t \mleq  \mathbf{V}^\H_t \mleq \frac{5}{4}\mathbf{U}^\H_t.$$
\end{proof}

\begin{lemma}
    (\cite{Faury2022}, Proposition 7) Let $\delta \in \pbrak{0,1}$ and $\cbrak{\pbrak{\bm\theta_t , \mathbf{W}_t , \bm\theta_t}}_r$ be maintained by the ada-OFU-ECOLog algorithm. Then, 
    $$\P\cbrak{\forall t \geq 1: \bm\theta^\star \in \bm\theta_t \text{ and } \matnorm{\bm\theta^\star - \bm\theta_{t+1}}{\mathbf{W}_{t+1}} \leq CS^2d\log(t/\delta)} \geq 1 - 2\delta.$$
    \label{lemma: ada_ofu_ecolog_confidence_bound}
\end{lemma}



\begin{lemma}
    (\cite{Abeille2021}, Theorem 1) Define 
    \[
    R_T = \sum\limits_{t=1}^{T}\sigmoid{\inner{{\mathbf{x}_t^\star}}{\bm\theta^\star}} - \sigmoid{\inner{\mathbf{x}_t}{\bm\theta^\star}}.\]
    Then, we have that
    \[\sum\limits_{t=1}^{T}\sens{\mathbf{x}_t}{\bm\theta^\star} \leq R(T) + \sum\limits_{t=1}^{T}\sens{{\mathbf{x}_t^\star}}{\bm\theta^\star}.
    \]
    \label{Lemma: Abielle result}
\end{lemma}
\begin{proof}  The proof follows by expanding the function $\dot\mu(.)$ using an exact Taylor expansion (Section \ref{appendix:notation}) as follows:
\begin{align*}
    \sum\limits_{t=1}^{T}\sens{\mathbf{x}_t}{\bm\theta^\star} &= \sum\limits_{t=1}^{T}\sens{{\mathbf{x}_t^\star}}{\bm\theta^\star} +  \sum\limits_{t=1}^{T}\int_0^{1}  \ddot{\mu}\pbrak{\inner{\mathbf{x}_t}{\bm\theta^\star} + v\inner{({\mathbf{x}_t^\star} - \mathbf{x}_t)}{\bm\theta^\star}}\; \diff v \inner{\pbrak{\mathbf{x}_t  -{\mathbf{x}_t^\star}}}{\bm\theta^\star}
    \\
    &\overset{}{\leq} \sum\limits_{t=1}^{T}\sens{{\mathbf{x}_t^\star}}{\bm\theta^\star} +  \sum\limits_{t=1}^{T}\int_0^{1}  \dsigmoid{\inner{\mathbf{x}_t}{\bm\theta^\star} + v\inner{({\mathbf{x}_t^\star} - \mathbf{x}_t)}{\bm\theta^\star}}\; \diff v \inner{\pbrak{{\mathbf{x}_t^\star}- \mathbf{x}_t}}{\bm\theta^\star} \\
    &\overset{}{=} \sum\limits_{t=1}^{T}\sens{{\mathbf{x}_t^\star}}{\bm\theta^\star} + \sum\limits_{t=1} ^{T}\sigmoid{\inner{{\mathbf{x}_t^\star}}{\bm\theta^\star}} - \sigmoid{\inner{\mathbf{x}_t}{\bm\theta^\star}}\\
    &= \sum\limits_{t=1}^{T}\sens{{\mathbf{x}_t^\star}}{\bm\theta^\star} + R(T).
\end{align*}

Here, the second inequality follows from the fact that ${\mathbf{x}^\star_t}^\top\bm\theta^\star \geq \mathbf{x}_t^\top \bm\theta^\star$ and the self-concordance property (Claim \ref{claim:self-concordance}) of the logistic function, while the second-to-last equality follows from the mean-value theorem.
\end{proof}

\begin{lemma}
    Let $\mathcal{T}$ represent the set of all time instances where the data-dependent condition fails, i.e $\forall t \in \mathcal{T}$, $\sens{\mathbf{x}_t}{\bar{\bm\theta}_t} \geq 2\sens{\mathbf{x}_t}{\bm\theta^u_t}$ for all $u \in \cbrak{0,1}$. Then,
    \begin{align*}
        \modulus{\mathcal{T}} = \tilde{O}(\kappa d^2 N^2 S^6).
    \end{align*}
    \label{lemma: number of times condition  fails}
\end{lemma}
\begin{proof}
By the self-concordance property of the logistic function, we know that
$$\sens{\mathbf{x}_t}{\bar{\bm\theta}_t} \leq \sens{\mathbf{x}_t}{\bm\theta^u_{t}}\exp\pbrak{\modulus{\inner{\mathbf{x}_t}{(\bar{\bm\theta}_t - \bm\theta^u_t)}}}.$$

Thus, if $t \in \mathcal{T}$, we have that $\modulus{\inner{\mathbf{x}_t}{(\bar{\bm\theta}_t - \bm\theta^u_t)}} \geq \log 2$.

Summing this over all indices in $\mathcal{T}$, we get that
\begin{align*}
\sum\limits_{t \in \mathcal{T}}{} \log^2 2  = \modulus{\mathcal{T}} \log^2 2 &\leq \sum\limits_{t \in \mathcal{T}}{}\modulus{\inner{\mathbf{x}_t}{(\bar{\bm\theta}_t - \bm\theta^u_t)}}^2
\\
&\overset{}{\leq} \sum\limits_{t \in \mathcal{T}}{} \matnorm{\mathbf{x}_t}{(\mathbf{V}^{\H}_{t})\inv}^2\matnorm{\bar{\bm\theta}_t - \bm\theta^u_t}{\mathbf{V}^\H_{t}}^2
\\
&\overset{}{\leq} CS^6Nd\log(T/\delta)\sum\limits_{t \in \mathcal{T}}{}\matnorm{\mathbf{x}_t}{(\mathbf{V}^{\H}_{t})\inv}^2
\end{align*}
where the last inequality follows from the fact that $\bm\theta^u_t$ and $\bar{\bm\theta}_t$ belong to $\Theta_t$, and the definition of $\beta_T(\delta)$. We continue bounding the terms as follows:
\begin{align*}
\lvert \mathcal{T} \rvert \log^2 2 &\overset{}{\leq} CS^6Nd\log(T/\delta)\sum\limits_{t \in \mathcal{T}}{}\matnorm{\mathbf{x}_t}{(\mathbf{V}^{\H}_{t})\inv}^2
\\
&\overset{}{\leq} C  S^6 N^2 d^2 \kappa \log(T/\delta)\log(T/\kappa)
\end{align*}

where the last inequality is a direct application of the Elliptical Potential Lemma (Lemma \ref{lemma: elliptical potential lemma}) on $\frac{1}{\sqrt\kappa} \mathbf{x}_t \in \R^{Nd}$ and the fact that $\left\lVert \frac{1}{\sqrt\kappa} \mathbf{x}_t \right\rVert \leq \frac{1}{\sqrt\kappa}$.
This finishes the proof.
\end{proof}

\clearpage

\section{\texttt{Slate-GLM-TS} and \texttt{Slate-GLM-TS-Fixed}}
\label{appendix:ts-algos}

\subsection{Algorithm in a fixed-arm setting}
In Algorithm \ref{algo:TS-Fixed}, we first present a Thompson Sampling-based algorithm \slateglincbtsfixed\ in the non-contextual (fixed-arm) setting. Following this, we analyze the regret of this algorithm in Theorem \ref{theorem:TS}. Since the arm-sets are fixed in this setting, we directly use the minimum eigenvalue bound in Assumption \ref{assumption_TS}. (See Remarks on Assumption \ref{assumption: diversity} in Section \ref{section:preliminaries}). 

\begin{algorithm}[H] 
\caption{\texttt{Slate-GLM-TS-Fixed}}
\label{algo:TS-Fixed} 
\begin{algorithmic}[1]
\STATE \textbf{Inputs:} Number of rounds $T$, Failure probability $\delta$ , Distribution $\mathcal{D}^{TS}$, and warm-up length $\tau$.
\STATE Initialize $\mathbf{V}^{\mathcal{H}}_0 = \lambda \mathbf{I}_{Nd}$ and $\mathbf{V}^{\mathcal{H},i}_0 = \lambda \mathbf{I}_d \; \forall i \in [N]$.
\STATE Obtain the set of items $\mathcal{X}^i, \forall i \in [N]$.
\FOR{each round $t$ in $[1,\tau]$}
    \STATE Choose $\mathbf{x}^i_t = \argmax_{\mathbf{x}\in\mathcal{X}^i}\matnorm{\mathbf{x}}{(\mathbf{V}^{\mathcal{H},i}_t)\inv} \;\forall i\in [N]$, play the slate $\mathbf{x}_t = (\mathbf{x}^1_t, \ldots, \mathbf{x}^N_t)$, and obtain reward $y_t$.
    \STATE Update $\mathbf{V}^{\mathcal{H}}_t \gets \mathbf{V}^{\mathcal{H}}_{t-1} + \frac{1}{\kappa} \mathbf{x}_t\mathbf{x}_t^\top$ and $\mathbf{V}^{\mathcal{H},i}_t \gets \mathbf{V}^{\mathcal{H},i}_{t-1} + \frac{1}{\kappa}\mathbf{x}^i_t{\mathbf{x}^{i}_t}^\top$, $\forall i\in [N]$.
\ENDFOR
\STATE Compute $\widehat{\bm\theta}_{\tau} = \argmin \summation{s=1}{\tau}l_{s+1}(\bm\theta) + \frac{\lambda}{2}\twonorm{\bm\theta}^2$ and set $\Theta = \cbrak{\matnorm{\bm\theta - \widehat{\bm\theta}_{\tau}}{\mathbf{V}^{\mathcal{H}}_{\tau}} \leq \beta_\tau(\delta)}$.
\STATE Initialize $\mathbf{W}_\tau = \mathbf{I}_{dN}, \mathbf{W}^{i}_\tau = \mathbf{I}_{d}, \forall i \in [N]$ and $\bm\theta_{\tau+1} \in \Theta$.
\FOR{each round $t \in [\tau+1 , T]$}
    \STATE Set reject = True.
    \WHILE{reject}
        \STATE For each slot $i \in [N]$, sample $\bm\eta^{i}\overset{\mathrm{iid}}{\sim} \mathcal{D}^{TS}$, and set  $\tilde{\bm\theta}^{i}_t = \bm\theta^{i}_t + \eta_t(\delta)(\mathbf{W}_t^i)^{-1/2}\bm\eta^{i}$.
    \STATE If $\tilde{\bm\theta}_t = (\tilde{\bm\theta}^{1}_t , \ldots , \tilde{\bm\theta}^{N}_t) \in \Theta_t$, set reject = False.
    \ENDWHILE
    \STATE Choose $\mathbf{x}^i_t = \argmax_{\mathbf{x} \in \X^i} \inner{\mathbf{x}}{\tilde{\bm\theta}^{i}_t} \;\forall i \in [N]$, play the slate $\mathbf{x}_t = (\mathbf{x}^1_t, \ldots, \mathbf{x}^N_t)$, and obtain reward $y_t$.

    \STATE Let $\bm\theta_{t+1}$ be solution of  \ref{equation:optimization} up to precision $1/t$.
    
    \STATE Update $\mathbf{W}_{t+1} = \mathbf{W}_t + \dot{\mu}(\mathbf{x}_t^T\bm\theta_{t+1})\mathbf{x}_{t}\mathbf{x}_{t}^T$, and $\mathbf{W}^i_{t+1} = \mathbf{W}^i_{t} + \dot{\mu}({\mathbf{x}_t}^\top \bm\theta_{t+1})\mathbf{x}^i_t{\mathbf{x}^i_t}^\top$, $\forall i\in [N]$.
\ENDFOR
\end{algorithmic}
\end{algorithm}

\begin{assumption}
    The minimum eigenvalue of the design matrices grows linearly, i.e,
\[
    \forall i \in [N], \forall t \in [\tau], \eigmin{\mathbf{V}^{\mathcal{H},i}_t} \geq \rho(\mathbf{V}^{\mathcal{H},i}) t.
\]
\[
    \forall i \in [N], \forall t \in [\tau+1,T], \eigmin{\mathbf{V}^{i}_t} \geq \rho(\mathbf{V}^{i}) (t-\tau) \quad \text{ and } \quad \eigmin{\mathbf{H}^{i}_t} \geq \rho(\mathbf{H}^{i}) (t-\tau).
\]
For simplicity, we shall denote $\rho = \max\limits_{i \in [N]} \max \{\rho(\mathbf{V}^i) , \rho(\mathbf{H}^i), \rho(\mathbf{V}^{\mathcal{H},i})\}$.
\label{assumption_TS}
\end{assumption}

Define $T_0 = O(N^2 \rho^{-2})$, where the exact expression is given in Lemma \ref{lemma: TS_bound_W}. We now state the regret guarantees for Algorithm \ref{algo:TS-Fixed}, and subsequently, provide a proof for the same.

\begin{theorem}
\label{theorem:TS}
    (Regret of \texttt{Slate-GLM-TS-Fixed}) Let $T \geq \tilde{O}(T_0 + \kappa d^2 N^2 S^6)$, then, the regret of \texttt{Slate-GLM-TS-Fixed} after $T$ rounds is bounded by
    \[
        Regret(T) \leq \tilde{O}(N^2\rho^{-2} + \kappa d^2 N^2 S^6) + CSN^{3/2}d^{3/2} \log \left(\frac{T}{\delta} \right) \sqrt{T\dsigmoid{\inner{\mathbf{x}_\star}{\thetastar}}} 
       + CN^3d^3S^2 \log^2 \left(\frac{T}{\delta} \right).
    \]
    \label{appendix: regret_proof_TS}
\end{theorem}
\begin{proof} 

Define the optimal slate $\mathbf{x}_\star = \argmax_{\mathbf{x} \in \mathcal{X}} \mathbf{x}^\top \bm\theta_\star$. First, using a trivial regret bound of $1$ for all rounds $t \in [\tau]$ gives us
\[
    Regret(T) \leq \tau + \sum_{t = \tau + 1}^T \mu(\mathbf{x}_\star^\top \bm\theta_\star) - \mu(\mathbf{x}_t^\top \bm\theta_\star).
\]

Now, we can only invoke Lemma \ref{lemma: TS_bound_W} for rounds $t$ such that $t - \tau \geq T_0$. Let $t^\prime = \tau + T_0$ such that for all rounds $s \geq t^\prime$, we have $s - \tau \geq T_0$. Thus, we use a trivial regret bound of $1$ for all rounds $t \in [\tau , t^\prime]$, and bound the regret for rounds $t \in [t^\prime ,T]$. This gives us
\[
    Regret (T) \leq T_0 + \tau + \underbrace{\sum_{t \in [t^\prime , T]} \mu(\mathbf{x}_\star^\top \bm\theta_\star) - \mu(\mathbf{x}_t^\top \bm\theta_\star)}_{R(T)}.
\]

We decompose $R(T)$ as follows:
\[R(T) = \underbrace{\summation{t\in[t^\prime,T]}{} \sigmoid{\inner{\mathbf{x}_\star}{\thetastar}} - \sigmoid{\inner{\mathbf{x}_t}{\tilde{\bm\theta}_t}}}_{R^{TS}(T)} + \underbrace{\summation{t\in[t^\prime,T]}{}\sigmoid{\inner{\mathbf{x}_t}{\tilde{\bm\theta}_t}} - \sigmoid{\inner{\mathbf{x}_t}{\thetastar}}}_{R^{PRED}(T)}.\]

\underline{\textbf{Bounding $R^{PRED}(T)$}}:
\begin{align*}
    R^{PRED}(T) &= \summation{t\in[t^\prime,T]}{}\sigmoid{\inner{\mathbf{x}_t}{\tilde{\bm\theta}_t}} - \sigmoid{\inner{\mathbf{x}_t}{\thetastar}}
    \\
    &\leq \summation{t\in[t^\prime,T]}{}\dsigmoid{\inner{\mathbf{x}_t}{\thetastar}} \modulus{\inner{\mathbf{x}_t}{\pbrak{\tilde{\bm\theta}_t - \thetastar}}}\\
    &\overset{}{\leq} C\sqrt{e}\sqrt{Nd \; \sigma_t(\delta)}\summation{t \in [t^\prime,T]}{}\sqrt{ \dsigmoid{\inner{\mathbf{x}_t}{\thetastar}}} \sqrt{\dsigmoid{\inner{\mathbf{x}_t}{\bm\theta_{t+1}}}} \matnorm{\mathbf{x}_t}{\mathbf{W}_t\inv}
    \\
    &\overset{}{\leq} C\sqrt{e}\sqrt{Nd \; \sigma_t(\delta)}\sqrt{\summation{t\in[t^\prime,T]}{}\dsigmoid{\inner{\mathbf{x}_t}{\thetastar}}} \sqrt{\summation{t \in [t^\prime,T]}{}\dsigmoid{\inner{\mathbf{x}_t}{\bm\theta_{t+1}}}\matnorm{\mathbf{x}_t}{\mathbf{W}_t\inv}^2} 
    \\
     &\overset{}{\leq} CS N^{3/2} d^{3/2} \sqrt{\log(T/\delta)\log(T/2)}\pbrak{\sqrt{R(T)} + \sqrt{T\dsigmoid{\inner{\mathbf{x}_\star}{\thetastar}}}}.
\end{align*}

where the second inequality follows from the self-concordance result (Claim \ref{claim:self-concordance}) in tandem with Lemma \ref{lemma: TS_concentration} and the fact that $\modulus{\mathbf{x}_t^\top\pbrak{\thetastar - \bm\theta_{t+1}}}\leq\diam{\mathcal{X}}{\Theta} \leq 1$ (Lemma \ref{lemma: warmup}). The last inequality is a result of applying Lemma \ref{lemma: elliptical potential lemma} on $\sqrt{\dsigmoid{\inner{\mathbf{x}_t}{\bm\theta_{t+1}}}}\mathbf{x}_t$, followed by Lemma \ref{Lemma: Abielle result}, and substituting the definition of $\sigma_t(\delta)$  from Section \ref{appendix:notation}.

\underline{\textbf{Bounding $R^{TS}(T)$}}:

 Define $J(\bm\theta) = \max\limits_{\mathbf{x}\in\mathcal{X}} \inner{\mathbf{x}}{\bm\theta}$. Then, using the definition of $\mathbf{x}_\star = \argmax\limits_{\mathbf{x} \in \mathcal{X}} \mathbf{x}^\top \bm\theta_\star$, it is easy to see that $J(\thetastar) = \mathbf{x}_\star^\top\thetastar$. Also, note that, at each time round, the items are chosen independently for each slot, resulting in
$$J(\tilde{\bm\theta}_t) = \max\limits_{\mathbf{x}\in\mathcal{X}}\summation{i=1}{N}{\mathbf{x}^i}^\top \tilde{\bm\theta}^i_{t} = \summation{i=1}{N} \max\limits_{\mathbf{x}\in\mathcal{X}^i}\mathbf{x}^\top\tilde{\bm\theta}^i_{t} = \summation{i=1}{N} \mathbf{x}_t^\top\tilde{\bm\theta}^{i}_{t} = \mathbf{x}_t^\top\tilde{\bm\theta}_{t}.$$
where we use the selection rule $\mathbf{x}^i_t = \argmax\limits_{\mathbf{x} \in \mathcal{X}^i} \mathbf{x}^\top \tilde{\bm\theta}^i_t$ (\emph{Step 16}, Algorithm \ref{algo:TS-Fixed}).

Hence, we can write
\begin{align*}
    R^{TS}(T) &=  \alpha(J(\thetastar) , J(\tilde{\bm\theta}_t)) \pbrak{J(\thetastar) -J(\tilde{\bm\theta}_{t})} .
\end{align*}

We now follow the same line of thought as in Section D.2 of the Appendix in \cite{Faury2022} and Section C of \cite{Abeille2017}. The convexity of $J$ gives us:
\begin{align*}
    \modulus{J(\thetastar) - J(\tilde{\bm\theta}_{t+1})} &\leq \max\cbrak{\modulus{\inner{\nabla J(\thetastar)}{\pbrak{\thetastar - \tilde{\bm\theta}_{t}}}} , \modulus{\inner{\nabla J(\tilde{\bm\theta}_{t+1})} {\pbrak{\thetastar - \tilde{\bm\theta}_{t}}}}}
    \\
    &\overset{}{\leq} \max\cbrak{\modulus{\inner{\mathbf{x}_\star}{\pbrak{\thetastar - \tilde{\bm\theta}_{t}}}} , \modulus{\inner{\mathbf{x}_t} {\pbrak{\thetastar - \tilde{\bm\theta}_{t}}}}}
    \\
    &\overset{}{\leq}1
\end{align*}
where the first inequality follows from the fact that $\nabla J(\bm\theta) = \argmax\limits_{\mathbf{x}\in\mathcal{X}}\mathbf{x}^\top\bm\theta$ (\cite{Abeille2017}), while the last inequality is a consequence of Lemma \ref{lemma: warmup}. 

Using this fact, we have that 
\begin{align*}
    \alpha(J(\thetastar) , J(\tilde{\bm\theta}_t))  &= \int\limits_{0}^1 \dsigmoid{J(\thetastar) + v\pbrak{J(\thetastar) - J(\tilde{\bm\theta}_t)}} \diff v\
    \\
    &\leq \dsigmoid{J(\thetastar)} \int\limits_{0}^1 \exp\pbrak{v\modulus{J(\thetastar) - J(\tilde{\bm\theta}_t)}} \diff v
    \\
    &\leq
    2\dsigmoid{\inner{\mathbf{x}_\star}{\thetastar}}
\end{align*}
where the first inequality follows from self-concordance (Claim \ref{claim:self-concordance}), while the last inequality follows from the definition of $J(\bm\theta)$.

Following the same steps as the proof in \cite{Abeille2017} and Section D.2 in \cite{Faury2022}, we get that 
$$\summation{t\in[t^\prime,T]}{}J(\thetastar) - J(\tilde{\bm\theta}_t) \lesssim C\sqrt{Nd \; \sigma_t(\delta)}\summation{t \in [t^\prime,T]}{}\matnorm{\mathbf{x}_t}{\mathbf{W}_t\inv} + \sqrt{T}.$$

Thus, the bound on $R^{TS}(T)$ is given by:
\begin{align*}
    R^{TS}(T) &\leq C\sqrt{Nd\;\sigma_t(\delta)}\summation{t \in [t^\prime,T]}{}\dsigmoid{\inner{\mathbf{x}_\star}{\thetastar}}  \matnorm{\mathbf{x}_t}{\mathbf{W}_t\inv} + 2\dsigmoid{\inner{\mathbf{x}_\star}{\thetastar}}\sqrt{T}.
\end{align*}

A bound on the first term can be given as follows:
\begin{align*}
    C\sqrt{Nd\sigma_t(\delta)}\summation{t \in [t^\prime,T]}{}\dsigmoid{\inner{\mathbf{x}_\star}{\thetastar}} \matnorm{\mathbf{x}_t}{\mathbf{W}_t\inv} 
    &\overset{}{\leq} C\sqrt{Nd\sigma_t(\delta)}\summation{t \in [t^\prime,T]}{}\sqrt{\dsigmoid{\inner{\mathbf{x}_\star}{\thetastar}}} \sqrt{\dsigmoid{\inner{\mathbf{x}_t}{\bm\theta_{t+1}}}}\matnorm{\mathbf{x}_t}{\mathbf{W}_t\inv} 
    \\
    &\overset{}{\leq} C\sqrt{Nd\sigma_t(\delta)}\sqrt{\summation{t \in [t^\prime,T]}{}\dsigmoid{\inner{\mathbf{x}_\star}{\thetastar}}} \sqrt{\summation{t \in [t^\prime,T]}{}\dsigmoid{\inner{\mathbf{x}_t}{\bm\theta_{t+1}}}\matnorm{\mathbf{x}_t}{\mathbf{W}_t\inv}^2}
    \\
     &\overset{}{\leq} CN^{3/2}d^{3/2}S\sqrt{\log(T/2)\log(T/\delta)}\sqrt{T\dsigmoid{\inner{\mathbf{x}_\star}{\thetastar}}} 
\end{align*}
where the first inequality follows from the self-concordance bound (Claim \ref{claim:self-concordance}), and uses the fact that $\lvert \mathbf{x}_t^\top \bm\theta_{t+1} - \mathbf{x}_\star^\top \bm\theta_\star \rvert \leq 2\diam{\mathcal{X}}{\Theta} \leq 2 $, while the last inequality uses Lemma \ref{lemma: elliptical potential lemma} on $\sqrt{\dot\mu(\mathbf{x}_t^\top \bm\theta_{t+1})} \mathbf{x}_t$, as well as Lemma \ref{Lemma: Abielle result} and the definition of $\sigma_T(\delta)$ (Section \ref{appendix:notation}).

Combining the bounds on $R(T)$, we get
$$R(T) \leq C SN^{3/2}d^{3/2} \sqrt{\log(T/\delta)\log(T/2)}\pbrak{\sqrt{R(T)} + \sqrt{T\dsigmoid{\inner{\mathbf{x}_\star}{\thetastar}}}}.$$

Using Lemma \ref{lemma: quadratic inequality}, we get

$$R(T) \leq  C SN^{3/2}d^{3/2} \sqrt{\log(T/\delta)\log(T/2)}\sqrt{T\dsigmoid{\inner{\mathbf{x}_\star}{\thetastar}}} + C N^3d^3S^2\log(T/\delta)\log(T/2).$$

Thus, substituting this bound back, and using the bounds on $\tau$ (from Lemma \ref{lemma: TS_bounds_on_warmup}) and $T_0$ gives us the desired result.

\end{proof}

\clearpage
\subsection{Supporting Lemmas for Theorem \ref{appendix: regret_proof_TS}}
\label{appendix: general_lemmas_ts}

\begin{lemma}
\label{lemma:multiplicative-equivalence}
    Let $\mathbf{U}_t$ and $\mathbf{W}_t$ be defined as in Section \ref{appendix:notation}. Let $ t - \tau \geq \frac{(N-1)^2}{2\rho^2}\log\frac{dN(N-1)}{\delta}$, then we have
    $$\frac{1}{2}\mathbf{U}_t\mleq \mathbf{W}_t \mleq \frac{3}{2}\mathbf{U}_t.$$
    \label{lemma: TS_bound_W}
\end{lemma}
\begin{proof} Using the same ideas as in Lemma \ref{lemma: independence of cross terms}, Lemma \ref{lemma: extension of diversity}, and Lemma \ref{lemma: norm_cross_terms}, we have that
\[
    \P \cbrak{ \forall i \in [N], \;\forall j \in [i+1 , N], \forall t \geq \tau+1,\norm{\mathbf{W}^{i,j}_t} \leq \sqrt{\frac{t - \tau}{2N^2}\log\frac{dN(N-1)}{\delta}}} \geq 1 - \delta.
\]

Next, for $i \in [N-1]$, define
\[
        \mathbf{Z}^{(i)}_t
        = \begin{bmatrix}
            f(N-i , N-i+1) , f(N-i , N-i+2) , \ldots , f(N-i , N)
        \end{bmatrix}, 
        \quad f(a,b) = (\mathbf{W}^{a}_t)^{-\frac{1}{2}} \mathbf{W}^{a,b}_t (\mathbf{W}^{b}_t)^{-\frac{1}{2}}.
\]
 Assumption \ref{assumption_TS} along with ideas similar to the ones in Lemma \ref{lemma: bound on norm of Z} results in:
\begin{align*}
    \norm{\mathbf{Z}^{(i)}_t} &\leq \summation{j=1}{i} \frac{\norm{\mathbf{W}^{N-i , N-i+j}_t}}{\sqrt{\eigmin{\mathbf{W}^{N-i}_t} \eigmin{\mathbf{W}^{N-i+j}_t} }}
    \leq \summation{j=1}{i} \frac{\sqrt{\frac{t - \tau}{2N^2}\log\frac{dN(N-1)}{\delta}}}{\rho (t-\tau)}
    \leq \frac{i}{N(N-1)},
\end{align*}
where the last inequality follows from the fact that $t -\tau \geq \frac{(N-1)^2}{2\rho^2}\log\frac{dN(N-1)}{2\delta}$.

Finally, using the same line of thought as Lemma \ref{lemma: ineq on W}, we get
$$\frac{1}{2}\mathbf{U}_t\mleq \mathbf{W}_t \mleq \frac{3}{2}\mathbf{U}_t.$$

\end{proof}

\begin{lemma}
    Let $\mathbf{V}^{\mathcal{H}}_t$ and $\mathbf{U}^{\mathcal{H}}_t$ be defined as in Section \ref{appendix:notation}. Also, let $\tau \geq \frac{8(N-1)^2}{\kappa^2\rho^2}\log\frac{dN(N-1)}{\delta}$. Then, we have,
    $$\frac{1}{2}\mathbf{U}^{\mathcal{H}}_\tau \mleq \mathbf{V}^\mathcal{H}_\tau \mleq \frac{3}{2}\mathbf{U}^\mathcal{H}_\tau.$$
    \label{lemma: TS_bounds_on_warmup}
\end{lemma}
\begin{proof} 
Using the same ideas as in Lemma \ref{lemma: bounds on V for failing data dependent condition}, we have that
\[
    \P\cbrak{\forall i \in [N], \forall j \in [i+1,N], \norm{\mathbf{V}^{\mathcal{H},i,j}_\tau} \leq \sqrt{\frac{8\tau}{\kappa^2 N^2} \log \left( \frac{dN(N-1)}{\delta} \right)}} \geq 1 - \delta.
\]

Now, for $i \in [N-1]$, define
\[
        \mathbf{Z}^{(i)}_\tau
        = \begin{bmatrix}
            f(N-i , N-i+1) , f(N-i , N-i+2) , \ldots , f(N-i , N)
        \end{bmatrix}, 
        \quad f(a,b) = (\mathbf{V}^{\mathcal{H} , a}_\tau)^{-\frac{1}{2}} \mathbf{V}^{\mathcal{H},a,b}_\tau (\mathbf{V}^{\mathcal{H},b}_\tau)^{-\frac{1}{2}}.
\]

Following the same line of thought as Lemma \ref{lemma: bounds on V for failing data dependent condition} and making use of Assumption \ref{assumption_TS}, we get

\[\norm{\mathbf{Z}^{(i)}_\tau} \leq \summation{j=1}{i} \frac{\sqrt{\frac{8 \tau}{\kappa^2N^2}\log\pbrak{\frac{d N(N-1)}{\delta}}}}{\rho \tau} \leq \frac{i}{N(N-1)}.\]
\end{proof}
where the last inequality follows from the fact that $\tau \geq \frac{8(N-1)^2}{\kappa^2\rho^2}\log\frac{dN(N-1)}{\delta}$.

Finally, using ideas from Lemma \ref{lemma: ineq on W}, we can show that
$$\frac{1}{2}\mathbf{U}^{\mathcal{H}}_t \mleq \mathbf{V}^\mathcal{H}_t \mleq \frac{3}{2}\mathbf{U}^\mathcal{H}_t.$$

\begin{lemma}
    Let $\delta \in (0,1)$, then, setting $\tau  =O(S^6N^2d^2\kappa \log(T/\delta)^2)$ ensures that $\Theta$ returned after the warm-up phase satisfies the following:
    \begin{enumerate}
        \item $\P\cbrak{\thetastar \in \Theta} \geq 1 - \delta$
        \item $\diam{\mathcal{X}}{\Theta} \leq 1$
    \end{enumerate}
    \label{lemma: warmup}
\end{lemma}

\begin{proof} The first part of the proof is the same as Proposition 5 in \cite{Faury2022}, since the proof is independent of the manner in which the arms are selected.

For the second part, let $\tau = O(N^2 \rho^{-2}\kappa^{-2})$. Then, notice that:
\begin{align*}
\diam{\mathcal{X}}{\Theta} &= \max\limits_{\mathbf{x}\in\mathcal{X}}\max\limits_{\bm\theta_1,\bm\theta_2\in\Theta} \modulus{\inner{\mathbf{x}}{\pbrak{\bm\theta_1 - \bm\theta_2}}}
\\
&\overset{}{\leq}\max\limits_{\mathbf{x}\in\mathcal{X}}\matnorm{\mathbf{x}}{(\mathbf{V}^{\mathcal{H}}_\tau)\inv}\max\limits_{\bm\theta_1,\bm\theta_2\in\Theta}\matnorm{\bm\theta_1-\bm\theta_2}{\mathbf{V}^{\mathcal{H}}_\tau}
\\
&\overset{}{\leq}\sqrt{\beta_t(\delta)}\max\limits_{\mathbf{x}\in\mathcal{X}}\matnorm{\mathbf{x}}{(\mathbf{V}^{\mathcal{H}}_\tau)\inv}\\
&\leq\sqrt{\beta_t(\delta)}\frac{1}{\sqrt{\tau}}\sqrt{2\summation{t=1}
{\tau}\summation{i=1}{N} \max_{\mathbf{x} \in \mathcal{X}^i} \matnorm{ \mathbf{x}}{(\mathbf{V}^{\mathcal{H},i}_t)\inv}^2}
\\
&\leq\sqrt{\beta_t(\delta)}\frac{\sqrt{\kappa}}{\sqrt{\tau}}\sqrt{2\summation{t=1}
{\tau}\summation{i=1}{N}\matnorm{\frac{1}{\sqrt\kappa}\mathbf{x}^i_t}{(\mathbf{V}^{\mathcal{H},i}_t)\inv}^2}
\\
&\leq C\sqrt{\frac{Nd\beta_t(\delta)\kappa\log(T/\kappa N)}{\tau}}.
\end{align*}
Here, the second inequality follows from the definition of $\Theta$. The third inequality uses Lemma \ref{lemma: TS_bounds_on_warmup} and subsequently, for $t \leq \tau$, $\mathbf{V}^{\mathcal{H},i}_t \preceq \mathbf{V}^{\mathcal{H},i}_\tau$. The second-to-last inequality uses the action-selection rule (\emph{Step 5}, Algorithm \ref{algo:TS-Fixed}), while the final inequality follows from Lemma \ref{lemma: elliptical potential lemma}.

Thus, setting $\tau = O(\max\{Nd\beta_t(\delta)\kappa\log(T/\kappa N) , N^2 \rho^{-2}\kappa^{-2}\})$ ensures $\diam{\mathcal{X}}{\Theta} \leq 1$. The result follows from the definition of $\beta_t(\delta)$ (Section \ref{appendix:notation}). 

\end{proof}

\begin{lemma}
    Define the distribution $\mathcal{D} = \bigtimes\limits_{i=1}^N \mathcal{D}^{TS}$ where $\mathcal{D}^{TS}$ is a multivariate distribution that satisfies the properties given in Definition \ref{def: D_TS}. Then, $\mathcal{D}$ also satisfies the properties given in Definition \ref{def: D_TS}, making it a suitable distribution for Thompson Sampling.
    \label{Lemma: TS_distribution_suitability}
\end{lemma}
\begin{proof} Define $\bm\eta = \pbrak{\bm\eta^1 , \ldots , \bm\eta^N} \in \R^{Nd}$ where $\bm\eta^i \sim \mathcal{D}^{TS}$. Then, sampling $\bm\eta$ from $\mathcal{D}$ is the same as sampling $\bm\eta^i \overset{\textrm{iid}}{\sim} \mathcal{D}^{TS}$ for all slots $i \in [N]$.

We begin by showing the Concentration property, i.e $\exists C , C^\prime$ such that 
$$\P_{\bm\eta \sim \mathcal{D}} \cbrak{\twonorm{\bm\eta} \leq \sqrt{C(Nd)\log\frac{C^\prime (Nd)}{\delta^\prime}}} \geq 1 - \delta^\prime.$$

Since $\mathcal{D}^{TS}$ satisfies the concentration property, we know that $\twonorm{\bm\eta^i} \geq \sqrt{cd\log\frac{c^\prime d}{\delta}}$ with probability at most $\delta$. Hence, it is easy to see that 
$$\twonorm{\bm\eta} =\sqrt{\summation{i=1}{N} \twonorm{\bm\eta_i}^2} \geq \sqrt{cNd\log\frac{c^\prime d}{\delta}}$$
with probability at most $\delta^N$. Setting $C = \frac{c}{N} , C^\prime = \frac{(c^\prime)^N d^{N-1}}{N}$ and $\delta^\prime = \delta^N$, we get that
$$\twonorm{\bm\eta} \leq \sqrt{CN^2d\log\pbrak{\frac{C^\prime Nd}{\delta^\prime}}^{1/N}} = \sqrt{C(Nd)\log\frac{C^\prime (Nd)}{\delta^\prime}}$$
with probability at least $1 - \delta^\prime$. This proves that $\mathcal{D}$ satisfies the concentration property.

We now show that $\mathcal{D}$ satisfies the Anti-Concentration property, i.e $\exists P \in (0,1)$ such that $\forall \mathbf{u} \in \R^{Nd}$:
$$\P_{\bm\eta \in \mathcal{D}}\cbrak{\mathbf{u}^\top{\bm\eta} \geq \twonorm{\mathbf{u}}} \geq P.$$

Assume $\mathbf{u} = \pbrak{\mathbf{u}^1 , \ldots , \mathbf{u}^N}$ such that $\twonorm{\mathbf{u}} = 1$. This implies that $\summation{i=1}{N}\twonorm{\mathbf{u}^i}^2 = 1$ which in turn implies that $\twonorm{\mathbf{u}^i} \leq 1$.

Since, $\twonorm{\mathbf{u}^i} \leq 1$, we have that $\twonorm{\mathbf{u}^i}^2 \leq \twonorm{\mathbf{u}^i}$, and since $\bm\eta^i \sim \mathcal{D}^{TS}$, we have that
$$\P\cbrak{{\mathbf{u}^i}^\top\bm\eta^i \leq \twonorm{\mathbf{u}^i}^2} \leq \P\cbrak{{\mathbf{u}^i}^\top\bm\eta^i \leq \twonorm{\mathbf{u}^i}} \leq 1- p.$$

Hence, we have that
\begin{align*}
    \P\cbrak{\mathbf{u}^\top\bm\eta \leq \twonorm{\mathbf{u}}} &= \P\cbrak{\mathbf{u}^\top\bm\eta \leq \twonorm{\mathbf{u}}^2}
    \\
    &=\P\cbrak{\summation{i=1}{N}{\mathbf{u}^i}^\top\bm\eta^i \leq \summation{i=1}{N} \twonorm{\mathbf{u}^i}^2}
    \\
    &\leq\prod\limits_{i=1}^{N}\P\cbrak{{\mathbf{u}^i}^\top\bm\eta^i \leq \twonorm{\mathbf{u}^i}^2}
    \\
    &\leq (1-p)^N.
\end{align*}

Thus, we have that $ \P\cbrak{\mathbf{u}^\top\bm\eta \geq \twonorm{\mathbf{u}}} \geq 1 - (1-p)^N$, and setting $P = 1 - (1-p)^N$ finishes the claim.
\end{proof}

\begin{lemma}
    At round $t \geq T_0$, let $\tilde{\bm\theta}^i = \bm\theta^i_t + \sqrt{\sigma_t(\delta)}(\mathbf{W}^i_t)^{-\frac{1}{2}} \bm\eta^i$ for all $i \in [N]$, where $\bm\eta^i \sim \mathcal{D}^{TS}$, as given in \emph{Steps 7-8} of Algorithm \ref{algo:TS-Fixed}. Define $\tilde{\bm\theta}_t = \pbrak{\tilde{\bm\theta}^1_t , \ldots , \tilde{\bm\theta}^N_t}$. Then, assuming the events in Lemma \ref{lemma: TS_bound_W} hold,  we have that,
    $$\matnorm{\tilde{\bm\theta} - \bm\theta_t}{\mathbf{W}_t} \leq C\sqrt{Nd \; \sigma_t(\delta)}. $$
\label{lemma: TS_concentration}
\end{lemma}
\begin{proof} We can write $\tilde{\bm\theta}_t = \pbrak{\tilde{\bm\theta}^1_t , \ldots , \tilde{\bm\theta}^N_t}$ as the following:
$$\tilde{\bm\theta}_t = \bm\theta_t + \sqrt{\sigma_t(\delta)}
\begin{bmatrix}
    (\mathbf{W}^1_t)^{-\frac{1}{2}} \bm\eta^1\\
    \\
    \vdots\\
    \\
    (\mathbf{W}^N_t)^{-\frac{1}{2}} \bm\eta^N
\end{bmatrix} = \bm\theta_t + \sqrt{\sigma_t(\delta)}\;\textrm{diag}((\mathbf{W}^1_t)^{-\frac{1}{2}} ,  \ldots , (\mathbf{W}^N_t)^{-\frac{1}{2}}) \bm\eta = \bm\theta_t +  \sqrt{\sigma_t(\delta)} \mathbf{U}_t^{-\frac{1}{2}}\bm\eta$$
where $\bm\eta = \pbrak{\bm\eta^1 , \ldots , \bm\eta^N}$.

Thus, we get
\begin{align*}
    \matnorm{\tilde{\bm\theta} - \bm\theta_t}{\mathbf{W}_t} &= \sqrt{\sigma_t(\delta)}\matnorm{\mathbf{U}_t^{-\frac{1}{2}}\bm\eta}{\mathbf{W}_t}
    \overset{}{\leq}\frac{3}{2}\sqrt{\sigma_t(\delta)}  \matnorm{\mathbf{U}_t^{-\frac{1}{2}}\bm\eta}{\mathbf{U}_t}
    =  \frac{3}{2}\sqrt{\sigma_t(\delta)} \twonorm{\bm\eta}
    \overset{}{\leq} C\sqrt{Nd \; \sigma_t(\delta)}
\end{align*}
where the first inequality follows from Lemma \ref{lemma: TS_bound_W} and the second inequality follows from the concentration property shown in Lemma \ref{Lemma: TS_distribution_suitability}.

\end{proof}

\clearpage
\section{Concentration results for Random Matrices and Vectors}
\label{appendix:general}

\begin{lemma}
    (\cite{Chatterji2020}, Generalization of Lemma 7) Let $\cbrak{\mathbf{x}_s}_{s=1}^\top$ be a stochastic process in $\R^d$ such that for filtration $\filteration{t}$ that captures all relevant history, we have that $\E\sbrak{\mathbf{x}_s|\filteration{s}} = \mathbf{0}_{d}$ and $\E\sbrak{\mathbf{x}_s\mathbf{x}_s^\top | \filteration{s}} \mgeq \rho\mathbf{I}_d$. Further, let $\twonorm{\mathbf{x}_s} \leq m$ for all $s \geq 1$. Also, define the matrix 
    $$\mathbf{Q}_t = \gamma\mathbf{I}_d + \summation{s=1}{t}\mathbf{x}_s\mathbf{x}_s^\top.$$
    Then, with probability at least $1 - \delta$, we have that
    $$\eigmin{\mathbf{Q}_t} \geq \gamma + c\rho t.$$
    for $0 \leq c \leq 1$ and for all t such that $\frac{12m^4 + 4m^2\rho(1-c)}{3(1-c)^2\rho^2}\log\pbrak{\frac{2dT}{\delta}} \leq t \leq T.$
    \label{lemma: min_eig_design}
\end{lemma}
\begin{proof} 

Assume $\E\sbrak{\mathbf{x}_s\mathbf{x}_s^\top | \filteration{s}} = \mathbf{\Sigma}_c \mgeq \rho\mathbf{I}_d$. Define the matrix martingale $\mathbf{Z}_t = \summation{s=1}{t}\sbrak{\mathbf{x}_s\mathbf{x}_s^\top - \mathbf{\Sigma}_c}$ with $\mathbf{Z}_0 = 0$ and the corresponding martingale difference sequence $\mathbf{X}_s = \mathbf{Z}_s - \mathbf{Z}_{s-1}$ for all $s \geq 1$. 

Since $\lVert \mathbf{x}_s \rVert_2 \leq m$, we have that $\lVert \mathbf{\Sigma}_c \rVert \leq m^2$ and hence, $\lVert \mathbf{X}_s \rVert \leq \lVert \mathbf{x}_s\mathbf{x}_s^\top \rVert + \lVert \mathbf{\Sigma}_c \rVert \leq 2m^2$.

Also, we have that 
\begin{align*}
\summation{s=1}{t}\norm{\E\sbrak{\mathbf{X}_s\mathbf{X}_s^\top | \filteration{s}}} &=  \summation{s=1}{t}\norm{\E\sbrak{\mathbf{x}_s\mathbf{x}_s^\top \mathbf{x}_s\mathbf{x}_s^\top - \mathbf{x}_s\mathbf{x}_s^\top\mathbf{\Sigma}_c^\top -\mathbf{\Sigma}_c \mathbf{x}_s\mathbf{x}_s^\top + \mathbf{\Sigma}_c\mathbf{\Sigma}_c^\top|\filteration{s-1}}}\\
&\leq  \summation{s=1}{t}\norm{\E\sbrak{\pbrak{\mathbf{x}_s^\top \mathbf{x}_s}\mathbf{x}_s \mathbf{x}_s^\top + \mathbf{\Sigma}_c\mathbf{\Sigma}_c^\top|\filteration{s-1}}}\\
&\leq 2m^4t.
\end{align*}

In a similar way, we can calculate 

\[
\summation{s=1}{t}\norm{\E\sbrak{\mathbf{X}_s^\top \mathbf{X}_s | \filteration{s}}} \leq 2m^4t.
\]

Thus, applying the Matrix Freedman Inequality (Lemma \ref{lemma: freedman inequality}) with $R = 2m^2 , \omega^2 = 2m^4t, d_1 = d_2 = d$ and $u = (1-c)\rho t$, we get
\begin{align*}
    \P\cbrak{\norm{\summation{s=1}{t}\sbrak{\mathbf{x}_s \mathbf{x}_s^\top - \mathbf{\Sigma}_c}} \geq (1-c)\rho t} \leq 2d \exp\pbrak{-\frac{(1-c)^2\rho^2t^2/2}{2m^4t + 2m^2(1-c)\rho t/3}}.
\end{align*}

Choosing $t \geq \frac{12m^4 + 4m^2\rho(1-c)}{3(1-c)^2\rho^2}\log\pbrak{\frac{2dT}{\delta}}$, with probability at least $1 - \frac{\delta}{T}$, 
\begin{align*}
    (1-c)\rho t \geq \norm{\summation{s=1}{t}\sbrak{\mathbf{x}_s\mathbf{x}_s^\top - \mathbf{\Sigma}_c}} .
\end{align*}

Using the triangle inequality, along with the fact that $\eigmax{-\mathbf{A}} = -\eigmin{\mathbf{A}}$, with probability at least $1 - \frac{\delta}{T}$,

\[(1-c)\rho t \geq \norm{\summation{s=1}{t}\sbrak{\mathbf{x}_s\mathbf{x}_s^\top - \mathbf{\Sigma}_c}} 
\geq \left\lvert \eigmax{\summation{s=1}{t}\sbrak{\mathbf{x}_s\mathbf{x}_s^\top}} - t\eigmax{\mathbf{\Sigma}_c} \right\rvert 
= \left\lvert \eigmin{\summation{s=1}{t}\sbrak{\mathbf{x}_s\mathbf{x}_s^\top}} - t\eigmin{\mathbf{\Sigma}_c} \right\rvert.\]

Finally, using the fact that $\mathbf{\Sigma}_c \succeq \rho \mathbf{I}_d$, with probability at least $1 - \frac{\delta}{T}$, we have

\[\eigmin{\summation{s=1}{t}\mathbf{x}_s\mathbf{x}_s^\top} \geq c\rho t. \]

A union bound over all $t \in [T]$ finishes the proof.

\end{proof}

\begin{lemma}
    (\cite{Das_2024}, Lemma 17) Let $\delta \in (0,1)$, $\mathbf{x}_s \in \R^{d_1}$ and $\mathbf{z}_s \in \R^{d_2}$ such that $\E\sbrak{\mathbf{x}_s\mathbf{z}_s^\top|\filteration{s-1}} = \mathbf{0}_{d_1\times d_2}$. Define $\mathbf{M}_t = \summation{s=1}{t}\mathbf{x}_s\mathbf{z}_s^\top$. Further, assume that $\twonorm{\mathbf{x}_s} \leq m_1$ and $\twonorm{\mathbf{z}_s} \leq m_2$. Then, with probability at least $1 - \delta$
    $$\norm{\mathbf{M}_t} \leq 2(m_1\wedge m_2)^2\sqrt{2t\log\pbrak{\frac{d_1+d_2}{\delta}}}.$$
    \label{lemma: generalized norm of cross terms}
\end{lemma}
\begin{proof} Denote $\mathbf{X}_s = \mathbf{x}_s\mathbf{z}_s^\top$. Since $\E\sbrak{\mathbf{X}_s|\filteration{s-1}} = \mathbf{0}_{d_1\times d_2}$, $\mathbf{X}_s$ is a Martingale Difference sequence. Further, $\mathbf{M}_t = \summation{s=1}{t}\mathbf{X}_s$ is the sum of Martingale Difference Sequences.

Consider the square of the Hermitian Dilation (see Definition \ref{def:HermitianDilation}) of $\mathbf{X}_s$
\begin{align*}
    \H(\mathbf{X}_s)^2 = \begin{bmatrix}
        \mathbf{0}_{d_1\times d_1} & \mathbf{X}_s\\
        \mathbf{X}_s^\top & \mathbf{0}_{d_2\times d_2}
    \end{bmatrix}^2 &= \begin{bmatrix}
        \mathbf{X}_s\mathbf{X}_s^\top & \mathbf{0}_{d_1\times d_2}\\
        \mathbf{0}_{d_2\times d_1} & \mathbf{X}_s^\top\mathbf{X}_s
    \end{bmatrix} \\
    &= \begin{bmatrix}
        \twonorm{\mathbf{z}_s}^2\mathbf{x}_s\mathbf{x}_s^\top & \mathbf{0}_{d_1\times d_2}\\
        \mathbf{0}_{d_2\times d_1} & \twonorm{\mathbf{x}_s}^2\mathbf{z}_s\mathbf{z}_s^\top
    \end{bmatrix} \\
    &\mleq (m_1 \wedge m_2)^2\begin{bmatrix}
        \mathbf{x}_s\mathbf{x}_s^\top & \mathbf{0}_{d_1\times d_2}\\
        \mathbf{0}_{d_2\times d_1} & \mathbf{z}_s\mathbf{z}_s^\top
        \end{bmatrix}\\
    & \mleq (m_1 \wedge m_2)^4\mathbf{I}_{d_1+d_2} .
\end{align*}

Applying the Matrix Azuma inequality (Lemma \ref{lemma: Matrix Azuma}) with $\mathbf{A}_s = (m_1 \wedge m_2)^2\mathbf{I}_{d_1+d_2}$, we have that $\sigma_t^2 = (m_1\wedge m_2)^4t$ and thus,
$$\P\cbrak{\exists t \geq 1: \singmax{\mathbf{M}_t} \geq \epsilon} \leq (d_1+d_2)\exp\pbrak{-\frac{\epsilon^2}{8 (m_1 \wedge m_2)^4t}}. $$

Choosing $\epsilon = \sqrt{8 (m_1 \wedge m_2)^4t\log\pbrak{\frac{d_1+d_2}{\delta}}}$ finishes the proof.

\end{proof}

\clearpage

\section{Other Useful Results and Definitions}
\begin{definition}
    (Hermitian Dilation) The Hermitian matrix for a matrix $\mathbf{A}$ is defined as 
    \begin{align*}
        \H(\mathbf{A}) = \begin{bmatrix}
            \mathbf{0} & \mathbf{A}\\
            \mathbf{A}^\top & \mathbf{0}
        \end{bmatrix}.
    \end{align*}
    \label{def:HermitianDilation}
\end{definition}

\begin{lemma}
    (\cite{Das_2024}, Lemma 16) Let $\mathbf{\H(\mathbf{Z}}) = \begin{bmatrix}
        \mathbf{0} & \mathbf{Z}\\
        \mathbf{Z}^\top & \mathbf{0}
    \end{bmatrix}$ where $\mathbf{Z}$ has positive singular values.
    Then, it holds almost surely, \[
    \eigmax{\H(\mathbf{Z})} = -\eigmin{\H(\mathbf{Z})} = \singmax{\mathbf{Z}}.
    \]
    \label{lemma: max_min_eig_hermitian}
\end{lemma}

\begin{lemma}
    (Matrix Freedman Inequality \cite{Tropp2011b}, Corollary 1.3) Define a matrix martingale $\mathbf{Z}_s \in \R^{d_1\times d_2}$ with respect to the filtration  $\filteration{s}$ that captures all relevant history and a martingale difference sequence $\mathbf{X}_s = \mathbf{Z}_s - \mathbf{Z}_{s-1}$. Assume that the difference sequence is almost surely uniformly bounded, i.e $\norm{\mathbf{X}_s} \leq R$. Define the quantities
    $$\mathbf{W}_{row,t} = \summation{s=1}{t}\E\sbrak{\mathbf{X}_s\mathbf{X}_s^\top | \filteration{s}},$$
    $$\mathbf{W}_{col,t} = \summation{s=1}{t}\E\sbrak{\mathbf{X}_s^\top\mathbf{X}_s | \filteration{s}}.$$
Then, for all $u \geq 0$ and $\omega^2 > 0$, we have

$$\P\cbrak{\exists t \geq 0: \norm{\mathbf{Z}_t} \geq u \text{ and } \max\cbrak{\norm{\mathbf{W}_{row,t}} , \norm{\mathbf{W}_{row,t}}}\leq \omega^2} \leq (d_1 + d_2) \exp\pbrak{-\frac{u^2/2}{\omega^2 + Ru/3}}.$$
\label{lemma: freedman inequality}
\end{lemma}

\begin{lemma}
    (Matrix Azuma Inequality, \cite{Tropp_2011a}, Theorem 7.1) Let $\cbrak{\mathbf{X}_s}_{s=1}^\infty$ be a matrix martingale difference sequence in $\R^{d_1\times d_2}$ and let $\H(\mathbf{X}_s)$ represent the Hermitian Dilation (see Definition \ref{def:HermitianDilation}) of $\mathbf{X}_s$. Let $\cbrak{\mathbf{A}_s}_{s=1}^\infty$ be a sequence of matrices in $\R^{(d_1+d_2) \times(d_1 + d_2)}$ such that $\E\sbrak{\mathbf{X}_s|\filteration{s}} = \mathbf{0}$ and $\H(\mathbf{X}_s)^2 \mleq \mathbf{A}_s^2$. Let $\sigma_t^2 = \lambda_{max}\left({\summation{s=1}{t}\mathbf{A}_k^2}\right)$ for $t\geq 1$. Then, for all $\epsilon \geq 0$:
    $$\P\cbrak{\exists t\geq1: \singmax{\summation{s=1}{t}\mathbf{X}_s} \geq \epsilon} \leq (d_1+d_2)\exp\pbrak{-\frac{\epsilon^2}{8\sigma_t^2}}.$$
    \label{lemma: Matrix Azuma}
\end{lemma}

\begin{lemma}
    (Elliptical Potential Lemma, \cite{Yadkori2011}, Lemma 11) Let $\cbrak{\mathbf{x}_s}_{s=1}^t$ represent a set of vectors in $\R^d$ and let $\twonorm{\mathbf{x}_s} \leq L$. Let $\mathbf{V}_s = \lambda\mathbf{I}_d + \summation{m=1}{s-1}\mathbf{x}_m\mathbf{x}_m^\top$. Then, for $\lambda \geq 1$,
    $$\summation{s=1}{t}\matnorm{\mathbf{x}_s}{\mathbf{V}_s\inv}^2 \leq 2d\log\pbrak{1 + \frac{tL^2}{\lambda d}} \leq 4d\log({tL^2}).$$
    \label{lemma: elliptical potential lemma}
\end{lemma}

\begin{lemma}
    (\cite{Abeille2021}, Proposition 7) Let $b , c \geq 0$ and $x^2 - bx - c \leq 0$. Then, $x^2 \leq 2b^2 + 2c$.
    \label{lemma: quadratic inequality}
\end{lemma}
\begin{proof} Since the coefficient of the quadratic term is 1, the quadratic expression can attain non-positive values only if it has two distinct or equal real roots. We denote the roots by $\alpha_1$ and $\alpha_2$. Without loss of generality, assume $\alpha_1 = \frac{b - \sqrt{b^2 + 4c}}{2}$ and $\alpha_2 = \frac{b + \sqrt{b^2 + 4c}}{2}$. Then, if $x \in [\alpha_1 , \alpha_2]$, we have that $x^2-bx-c \leq 0$. Hence, we can also say
$$x \leq 
\alpha_2 = \frac{b + \sqrt{b^2 + 4c}}{2} \leq b + \sqrt{c}$$
using the fact that $\sqrt{a+b}\leq \sqrt{a} + \sqrt{b}$ for $a,b \geq 0$.
Finally, 
$$x^2 \leq b^2 + c + 2b\sqrt{c} \leq 2b^2 + 2c$$
using the fact that $(b - \sqrt{c})^2 \geq 0 \implies 2b\sqrt{c} \leq b^2 + c$.

\end{proof}

\begin{definition}
    (Multivariate distribution for Thompson Sampling, \cite{Abeille2017}, Definition 1) $\mathcal{D}^{TS}$ is a suitable multivariate distribution on $\R^d$ for Thompson Sampling if it is absolutely continuous with respect to the Lebesgue measure and satisfies the following properties:
    \begin{enumerate}
        \item Concentration: There exist constants $c$ and $c^{'}$ such that $\forall \delta \in (0,1)$,
        $$\P_{\bm{\eta} \sim\mathcal{D}^{TS}}\cbrak{\twonorm{\bm{\eta}} \leq \sqrt{cd\log \frac{c^{'}d}{\delta}}} \geq 1 -\delta.$$

    \item Anti-Concentration: There exists a strictly positive probability $p$ such that for any $\mathbf{u} \in \R^d$,
    $$\P_{\bm{\eta} \sim \mathcal{D}^{TS}} \cbrak{\mathbf{u}^\top \bm{\eta} \geq \twonorm{\mathbf{u}}} \geq p.$$
    \end{enumerate}
\label{def: D_TS}
\end{definition}

\clearpage
\section{Additional Experiments and Experimental Details}
\label{appendix:experiments}
In this section, we provide additional plots to back the experiments shown in Section \ref{section:experiments}. Also, we provide additional details about our experimental setup. 

\begin{figure*}[h]
	\centering
	\begin{subfigure}[b]{0.33\columnwidth}  
		\centering 
		\includegraphics[width=56mm]{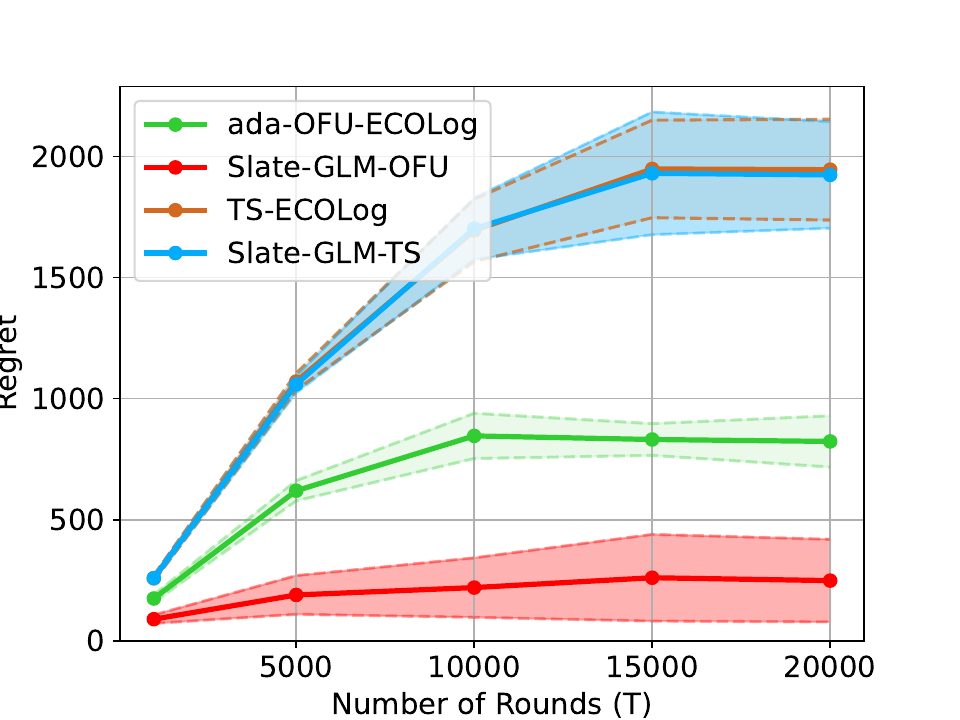}
		\caption{{\small Regret vs.\ $T$: Finite Context Setting}}   
		\label{fig:finite_error}
	\end{subfigure}
	\hfill
	\begin{subfigure}[b]{0.33\columnwidth}   
		\centering 
	\includegraphics[width=56mm]{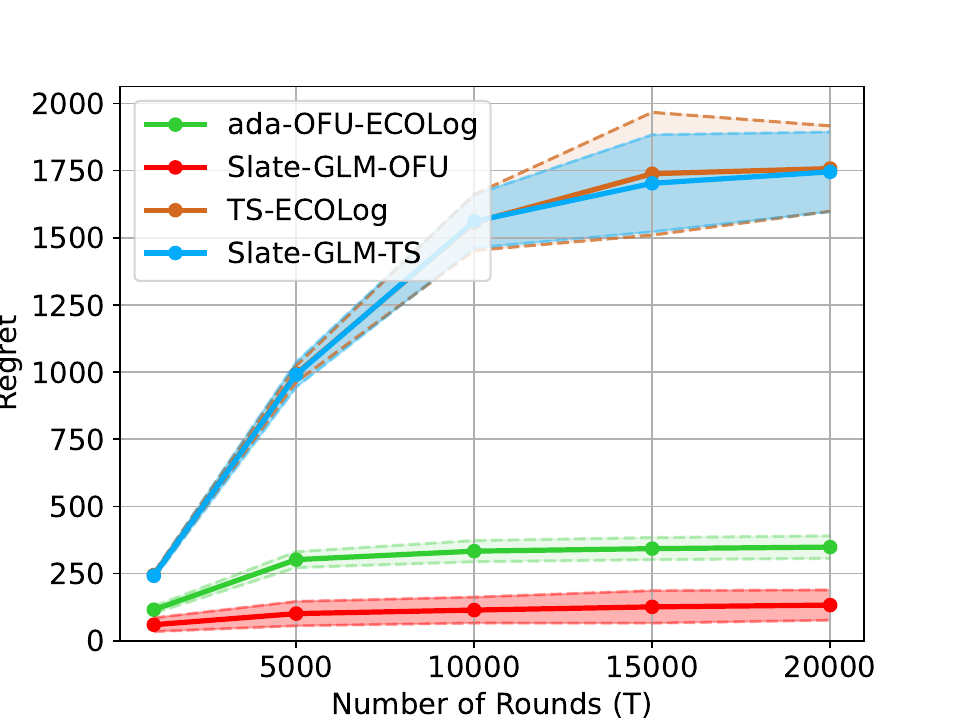}
		\caption{{\small Regret vs.\ $T$: Infinite Context Setting}}   
		\label{fig:infinite_error}
	\end{subfigure}
	\hfill
	\begin{subfigure}[b]{0.33\columnwidth}
		\centering
		\includegraphics[width=56mm]{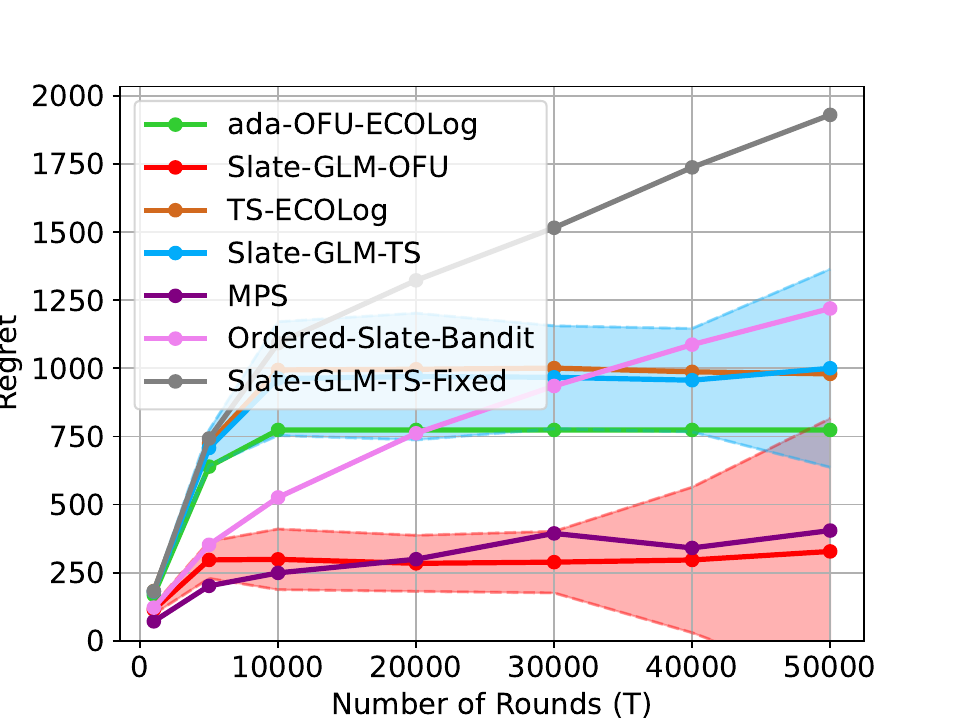}
		\caption{{\small Regret vs.\ $T$: Fixed-Arm Setting}}     
		\label{fig:non_contextual_our_error}
	\end{subfigure}
	\vskip\baselineskip
	\begin{subfigure}[b]{0.33\columnwidth}  
		\centering 
		\includegraphics[width=56mm]{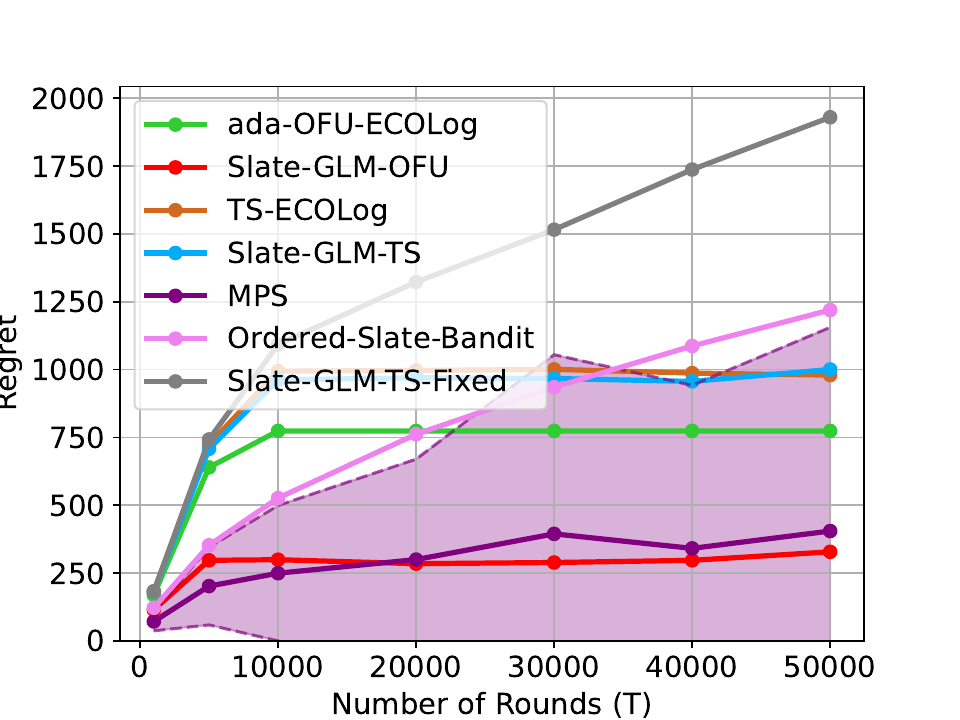}
		\caption{{\small Regret vs.\ $T$: Fixed-Arm Setting}} 
		\label{fig:non_contextual_mps_error}
	\end{subfigure}
	\hfill
	\begin{subfigure}[b]{0.33\columnwidth}   
		\centering 
		\includegraphics[width=56mm]{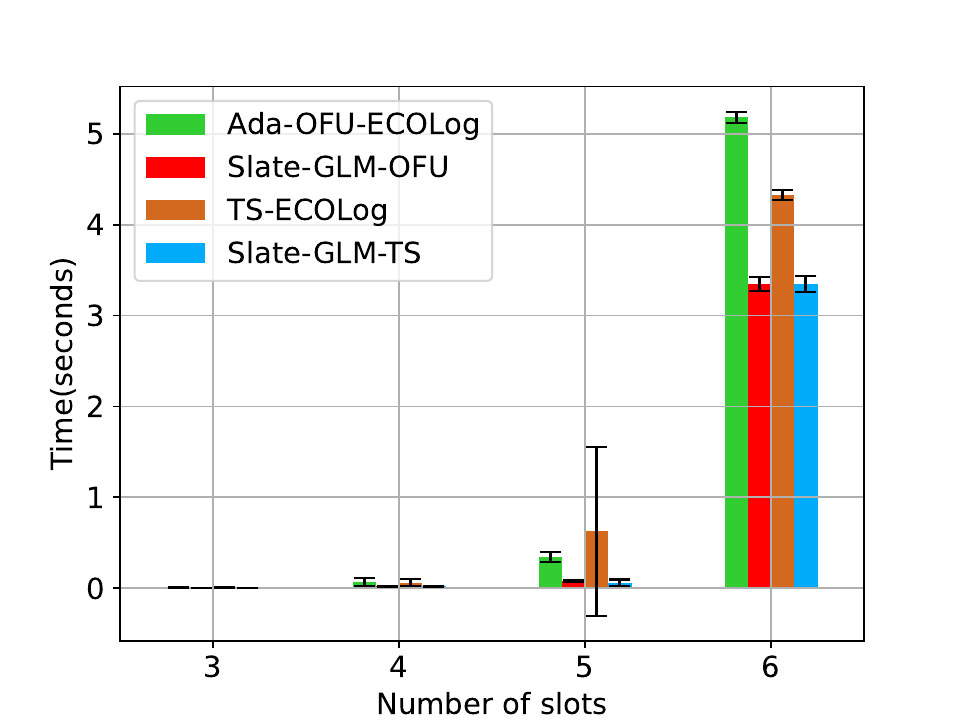}
				\caption{{\small Average running time (per-round)}}   
		\label{fig:average_time_error}
	\end{subfigure}
	\hfill
	\begin{subfigure}[b]{0.33\columnwidth}
		\centering
		\includegraphics[width=56mm]{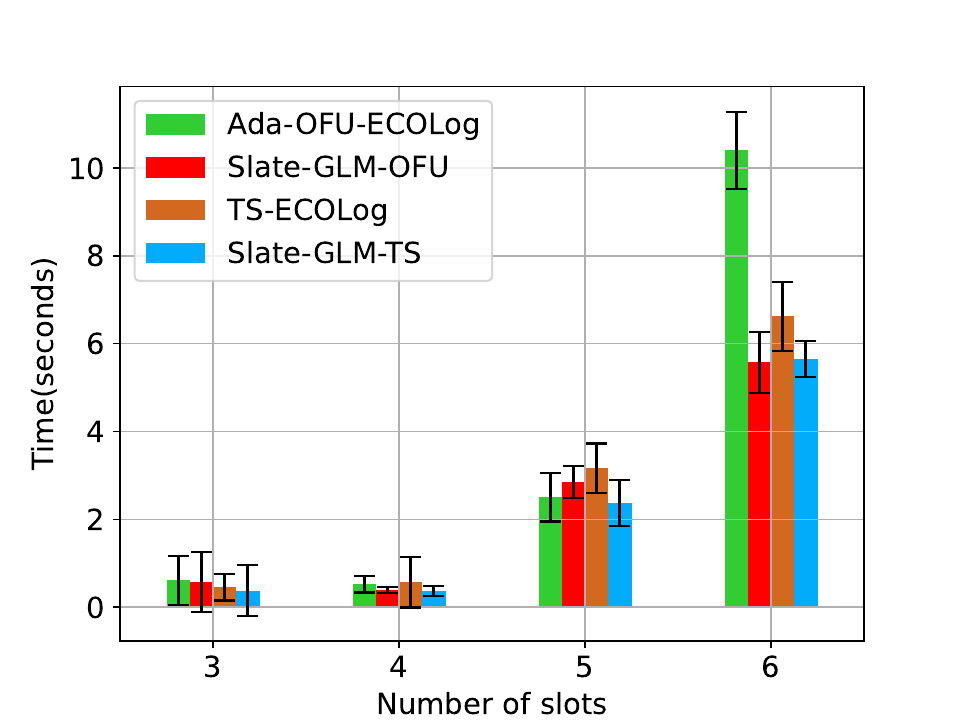}
		\caption[observationalAlgo]%
        {{\small Maximum running time (per-round)}}   
		\label{fig:max_time_error}
	\end{subfigure}
    \vskip\baselineskip
	\begin{subfigure}[b]{0.33\columnwidth}  
		\centering 
		\includegraphics[width=56mm]{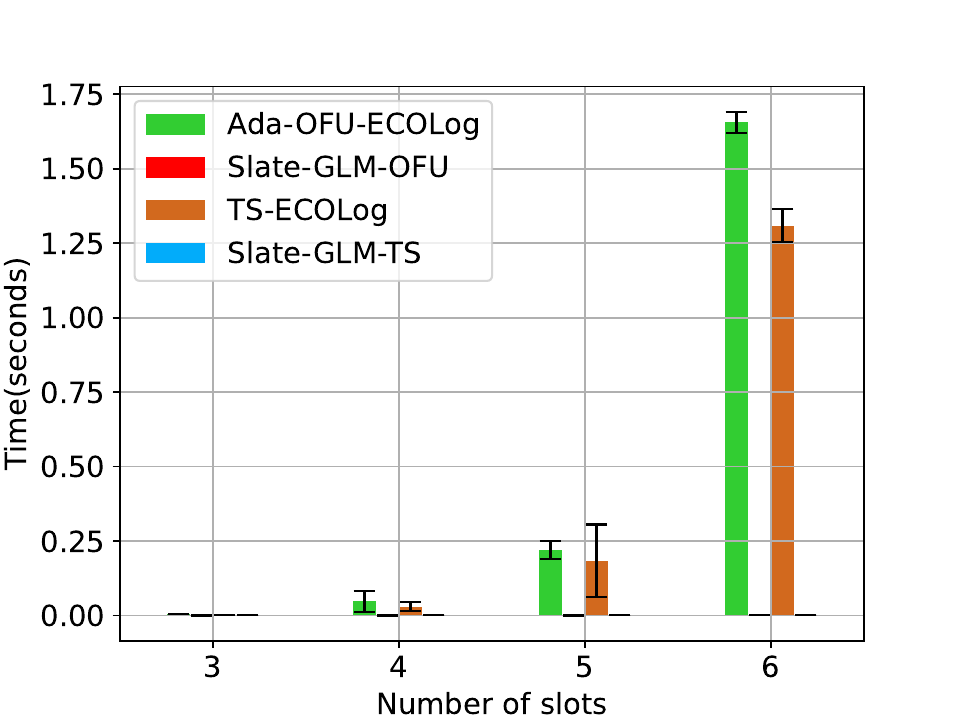}
		\caption{{\small Average time taken to pull an arm}} 
		\label{fig:pull_time_avg_error}
	\end{subfigure}
	\hfill
	\begin{subfigure}[b]{0.33\columnwidth}   
		\centering 
		\includegraphics[width=56mm]{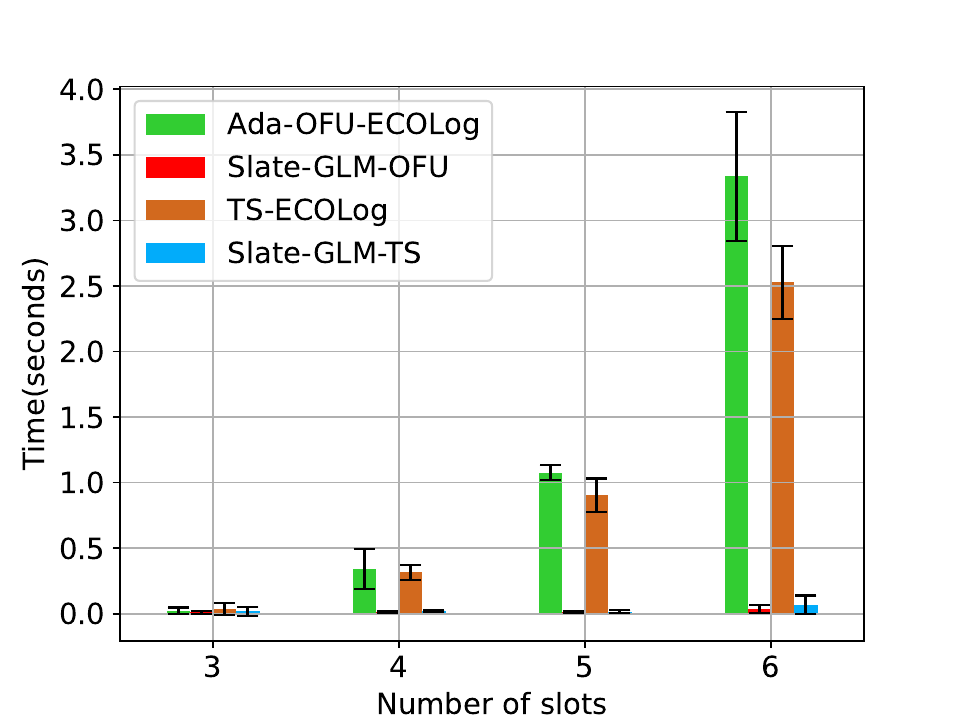}
				\caption{{\small Maximum time taken to pull an arm}}   
		\label{fig:pull_time_max_error}
	\end{subfigure}
	\hfill
	\begin{subfigure}[b]{0.33\columnwidth}
		\centering
		\includegraphics[width=56mm]{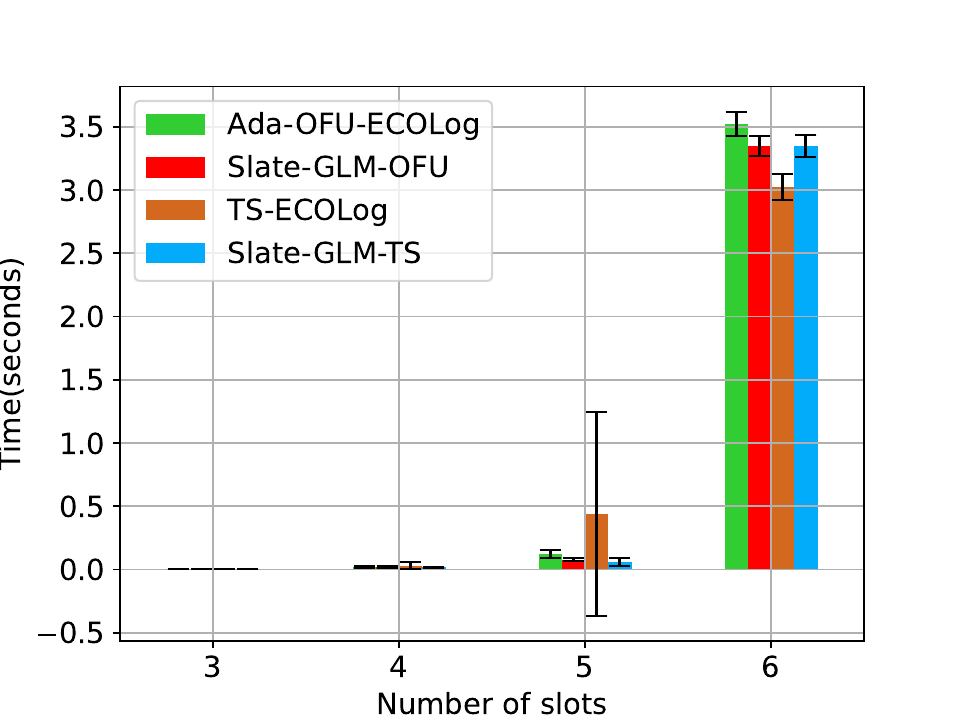}
		\caption[observationalAlgo]%
        {{\small Average time taken to update parameters}}   
		\label{fig:update_time_average_error}
	\end{subfigure}
	\caption{}
    \label{fig:Plots_appendix}
\end{figure*}

In all of the figures, the shaded regions represent two standard deviations. Figures \ref{fig:finite_error} and \ref{fig:infinite_error} depict the graphs from \textbf{Experiment 1} (Section \ref{section:experiments}) wherein we compare our algorithms \texttt{Slate-GLM-OFU} and \texttt{Slate-GLM-TS} to their counterparts \texttt{ada-OFU-ECOLog} and \texttt{TS-ECOLog} in the finite and infinite context settings.

Figures \ref{fig:non_contextual_our_error} and \ref{fig:non_contextual_mps_error} depict the graphs from \textbf{Experiment 3} (Section \ref{section:experiments}), wherein we compare our algorithms \texttt{Slate-GLM-OFU}, \texttt{Slate-GLM-TS}, and \texttt{Slate-GLM-TS-Fixed} to several state-of-the-art non-contextual logistic bandit algorithms. In Figure \ref{fig:non_contextual_our_error}, we only show the uncertainity involved in \texttt{Slate-GLM-OFU} and \texttt{Slate-GLM-TS}. We see that \texttt{Slate-GLM-OFU} has the best performance, with the only algorithm having comparable performance being \texttt{MPS}. On the other hand, \texttt{Slate-GLM-TS} performs worse than \texttt{ada-OFU-ECOLog} and \texttt{MPS}, while being on par with \texttt{TS-ECOLog}. However, in Figure \ref{fig:non_contextual_mps_error}, we showcase that the variance of \texttt{MPS} is very high, hence, making the algorithm less reliable in practice.

Figures \ref{fig:average_time_error} and \ref{fig:max_time_error} showcase two standard deviations in the average and maximum (per-round) running time of the algorithms. We see that both \texttt{ada-OFU-ECOLog} and \texttt{TS-ECOLog} show an exponential increase in their running times. Further, the significant gap between the average and maximum (per-round) running times of \texttt{Slate-GLM-OFU} and \texttt{Slate-GLM-TS} (as highlighted in the table below) indicates that the true per-round time is much lower than the maximum. As we have mentioned in the main paper, we calculate the per-round running time for an algorithm as the sum of the per-round pull and update times. Figures \ref{fig:pull_time_avg_error} and \ref{fig:pull_time_max_error} show the average and maximum pull times (per round), while Figure \ref{fig:update_time_average_error} display the average per-round update times. We see that the pull time for \texttt{ada-OFU-ECOLog} and \texttt{TS-ECOLog} increases exponentially with the number of slots, whereas the update times remain similar for all algorithms. Hence, the differences in per-round running times can be majorly attributed to the pulling times for each algorithm, which is in line with our theoretical claims. We also tabulate the average and maximum per-round pulling times for each algorithm in Table \ref{tab:running_times} for more clarity.

\begin{table}[H]
    \centering
\resizebox{\columnwidth}{!}{
    \begin{tabular}{ccccccccc}
    \hline
      \multirow{2}{*}{\textbf{Slots}}
      &
      \multicolumn{2}{c}
      {\texttt{ada-OFU-ECOLog}}
      &
      \multicolumn{2}{c}
      {\texttt{Slate-GLM-OFU}}
      &
      \multicolumn{2}{c}
      {\texttt{TS-ECOLog}}
      &
      \multicolumn{2}{c}
      {\texttt{Slate-GLM-TS}}
      \\
      \cline{2-9}
      &
      \multicolumn{1}{c}{\textbf{Average (ms)}}
      &
      \multicolumn{1}{c}{\textbf{Maximum (ms)}}
      &
      \multicolumn{1}{c}{\textbf{Average (ms)}}
      &
      \multicolumn{1}{c}{\textbf{Maximum (ms)}}&
      \multicolumn{1}{c}{\textbf{Average (ms)}}
      &
      \multicolumn{1}{c}{\textbf{Maximum (ms)}}
      &
      \multicolumn{1}{c}{\textbf{Average (ms)}}
      &
      \multicolumn{1}{c}{\textbf{Maximum (ms)}}
      \\
      \hline
      3 
      & 
      \multicolumn{1}{c}{$4.3 \pm 0.2$}
      & 
      \multicolumn{1}{c}{$23.0 \pm 24.5$}
      & 
      \multicolumn{1}{c}{$0.3 \pm 0.0$}
      & 
      \multicolumn{1}{c}{$9.5 \pm 12.5$}
      & 
      \multicolumn{1}{c}{$3.1 \pm 0.1$}
      & 
      \multicolumn{1}{c}{$36.6 \pm 47.7$}
      & 
      \multicolumn{1}{c}{$0.6 \pm 0.1$}
      & 
      \multicolumn{1}{c}{$19.2 \pm 33.4$}
      \\
      \hline
      4 
      & 
      \multicolumn{1}{c}{$47.5 \pm 36.4$}
      & 
      \multicolumn{1}{c}{$341.8 \pm 154.8$}
      & 
      \multicolumn{1}{c}{$0.8 \pm 0.9$}
      & 
      \multicolumn{1}{c}{$10.5 \pm 7.3$}
      & 
      \multicolumn{1}{c}{$30.3 \pm 15.6$}
      & 
      \multicolumn{1}{c}{$316.7 \pm 57.1$}
      & 
      \multicolumn{1}{c}{$2.2 \pm 1.1$}
      & 
      \multicolumn{1}{c}{$22.8 \pm 5.4$}
      \\
      \hline
      5 
      & 
      \multicolumn{1}{c}{$221.4 \pm 30.1$}
      & 
      \multicolumn{1}{c}{$1075.7 \pm 57.8$}
      & 
      \multicolumn{1}{c}{$0.6 \pm 0.2$}
      & 
      \multicolumn{1}{c}{$12.0 \pm 9.7$}
      & 
      \multicolumn{1}{c}{$184.1 \pm 121.8$}
      & 
      \multicolumn{1}{c}{$905.3 \pm 126.5$}
      & 
      \multicolumn{1}{c}{$1.2 \pm 0.5$}
      & 
      \multicolumn{1}{c}{$13.8 \pm 11.5$}
      \\
      \hline
      6 
      & 
      \multicolumn{1}{c}{$1655.6 \pm 36.3$}
      & 
      \multicolumn{1}{c}{$3335.5 \pm 494.6$}
      & 
      \multicolumn{1}{c}{$0.9 \pm 0.2$}
      & 
      \multicolumn{1}{c}{$35.8 \pm 28.9$}
      & 
      \multicolumn{1}{c}{$1309.4 \pm 55.3$}
      & 
      \multicolumn{1}{c}{$2528.3 \pm 278.2$}
      & 
      \multicolumn{1}{c}{$1.9 \pm 0.2$}
      & 
      \multicolumn{1}{c}{$68.3 \pm 71.1$}
      \\
      \hline
    \end{tabular}
    }
    \caption{Average and Maximum per-round running times (in milliseconds), averaged over 10 different seeds for sampling rewards, with 2 standard deviations}
    \label{tab:running_times}
\end{table}

Now, we provide additional details about our experimental setup. In \textbf{Experiment 3}, we implement \texttt{Ordered Slate Bandit} and \texttt{ETC-Slate} from \cite{Kale2010} and \cite{Rhuggenaath2020} respectively. Since these algorithms are designed for semi-bandit feedback, we modify these algorithms to implement them in our setting. These modifications are detailed below:

\textbf{\texttt{Ordered Slate Bandit}}: The original algorithm in \cite{Kale2010} assumes that there exists a base set $\mathcal{X}$ such that $\modulus{\mathcal{X}} = K$ and the learner picks a slate of $N$ items from $\mathcal{X}$. Hence, their algorithm assumes that each base item is equally likely to be placed in any slot. Thus, they start with the initial distribution $P^\prime$ such that $P^\prime_{i,j} = 1 \;\forall i\in[N] \;\forall j\in[K]$. On the other hand, we cannot make the same assumption since we get a different set of items $\mathcal{X}^i_t$ for each slot $i\in[N]$. Thus, we change the initial distribution to $P$ such that $P_{i,j} = 1$ if and only if $j \in [K(i-1)+1 , K(i)]$. This modification restricts the items that can be selected for a particular slot. A similar modification is made for the exploratory distribution in each round. There is a significant difference in the manner in which the loss matrix is constructed. Since the algorithm is designed for semi-bandit feedback, the algorithm propagates the loss for the item chosen in each slot at each round. We make use of the fact that the loss is the additive inverse of the reward, and hence, we have two choices for the loss we wish to propagate. Since we operate in the logistic setting, the obvious choice is to propagate the non-linear losses to the algorithm. However, since the total loss for a slate is assumed to be the sum of the loss obtained for each slot, the linear loss seems more suitable (the logistic function is not sub-additive). We experiment with both these choices, and find that the algorithm with non-linear losses incurs very high regret. Hence, we only compare our algorithms to the Ordered Slate Bandit algorithm with linear losses, referred to as \texttt{Ordered Slate Bandit}.

\textbf{\texttt{ETC-Slate}: } The original algorithm in \cite{Rhuggenaath2020} is also designed for semi-bandit feedback, wherein, it is assumed that the reward for each slot is sampled from a distribution such as the uniform distribution (see Example 1 in \cite{Rhuggenaath2020}). However, in our case, we do not have a notion of a reward distribution at the slot level. Hence, to create a reward distribution at the slot level, we assume that the reward for slot $i$ is sampled from $\mathcal{N}({\mathbf{x}_s^i}^\top\thetastar^i , 0.0001)$. This ensures that, in expectation, the reward attributed to a particular slot is the linear reward for the item played. We set the slate-level reward function $f$ to simply be the sigmoid function applied to the sum of the rewards obtained at the slot levels and then proceed with the algorithm. We find that \texttt{ETC-Slate} incurs very high regret, and hence, do not include the algorithm in our comparisions.

\clearpage 

\section{Empirical Validation of the Diversity Assumption (Assumption \ref{assumption: diversity})}
\label{appendix:empirical-validation}

\begin{figure}[h]
    \centering
    \includegraphics[width = \columnwidth]{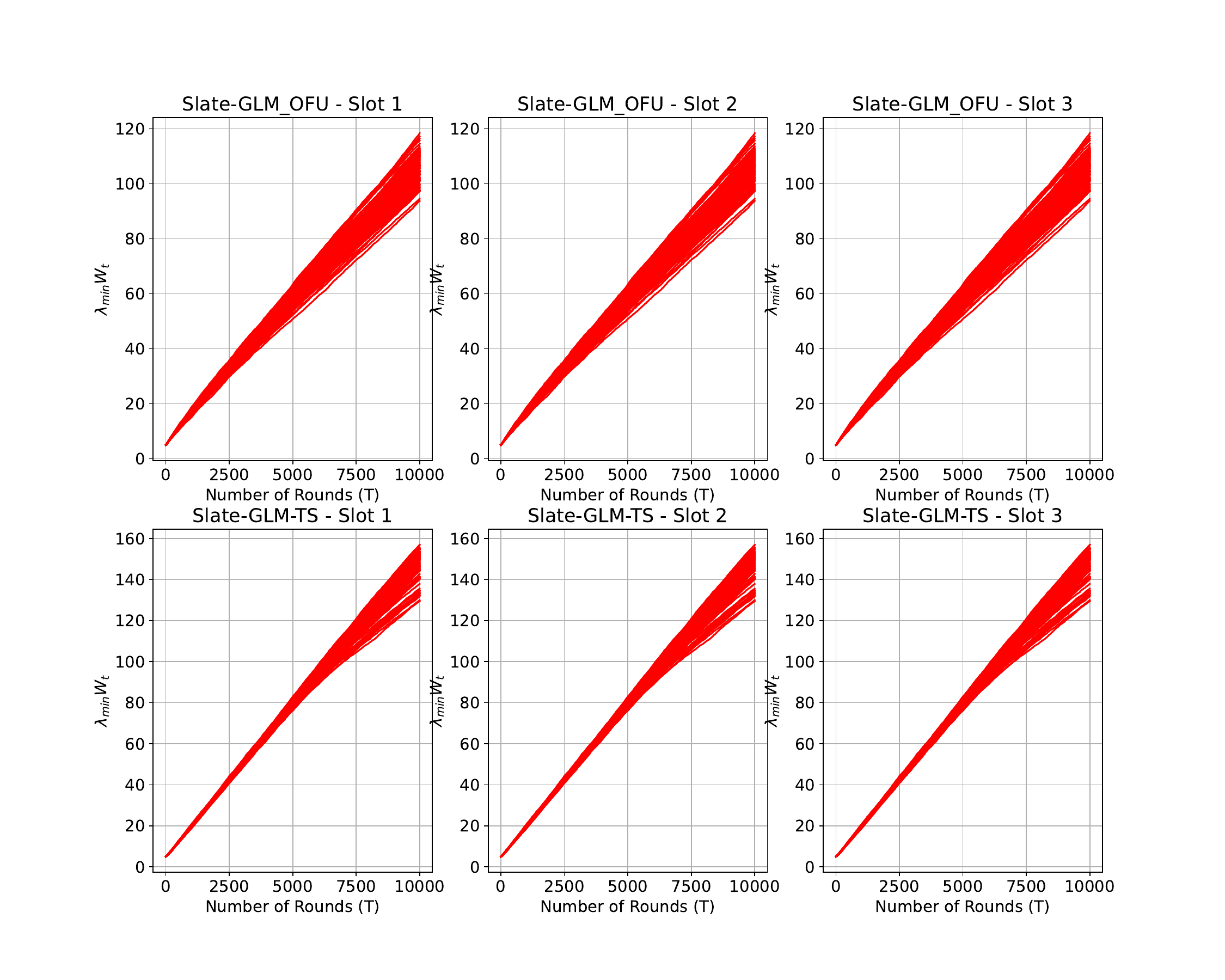}
    \caption{Demonstration of the algorithm-dependent assumption for \texttt{Slate-GLM-OFU} and \texttt{Slate-GLM-TS} wherein we plot the minimum eigenvalues of $\mathbf{W}^i_t$ as a function of the time round for 100 independent runs}
    \label{appendix_fig: eigenvalues}
\end{figure}

In this section, we show that our (instance and algorithm-dependent) diversity assumption we make indeed holds for several randomly chosen instances. We choose the number of slots $N$ to be $3$ and the number of items in each slot $\modulus{\mathcal{X}^i_t}$ is fixed to $5$. The dimension of items for each slot is fixed to $5$, resulting in the slate having a dimension $Nd = 15$. The items for each slot are randomly sampled from $[-1,1]^5$ and normalized to have norm $1/\sqrt{3}$, while $\thetastar$ is randomly sampled from $[-1,1]^{15}$. We operate in the Infinite context setting, wherein the items in each slot change every time round (check \textbf{Experiment 1} in Section \ref{section:experiments} for more details). We run both \texttt{Slate-GLM-OFU} and \texttt{Slate-GLM-TS} 100 times with different seeds for $T = 10000$ rounds. For each run of the algorithm, we plot the minimum eigenvalue of $\mathbf{W}^i_t$ ($i \in [3]$) as a function of the time round $t$ and show our results in Figure \ref{appendix_fig: eigenvalues}. The figures clearly depict (near) linear growth in the eigenvalues of the matrices $\mathbf{W}^i_t$ for all the slots $i \in [3]$ and all rounds $t \in [T]$.

\end{document}